\documentclass[10pt,journal,compsoc]{IEEEtran}
\ifCLASSOPTIONcompsoc
  \usepackage[nocompress]{cite}
\else
  \usepackage{cite}
\fi
\ifCLASSINFOpdf
   \usepackage[pdftex]{graphicx}
  \graphicspath{{../pdf/}{../jpeg/}}
  \DeclareGraphicsExtensions{.pdf,.jpeg,.png}
\else
  \usepackage[dvips]{graphicx}
  \graphicspath{{../eps/}}
  \DeclareGraphicsExtensions{.eps}
\fi
\usepackage{amsmath}
\usepackage{cases}

\usepackage{algorithm}
\usepackage{algorithmic}

\newcommand{\BEGIN}[1]{\STATE \textbf{Begin} \begin{ALC@g}} 
\newcommand{\END}{\end{ALC@g} \STATE \textbf{End}}
\newcommand{\SWITCH}[1]{\STATE \textbf{switch} (#1)}
\newcommand{\ENDSWITCH}{\STATE \textbf{end switch}}
\newcommand{\CASE}[1]{\STATE \textbf{case} #1\textbf{:} \begin{ALC@g}}
\newcommand{\ENDCASE}{\end{ALC@g}}

\newcommand{\DEFAULT}{\STATE \textbf{default:} \begin{ALC@g}}
\newcommand{\ENDDEFAULT}{\end{ALC@g}}
\newcommand{\DEFAULTLINE}[1]{\STATE \textbf{default:} }
\newcommand{\ON}[1]{\STATE \textbf{On} #1\textbf{:} \begin{ALC@g}} 
\newcommand{\ENDON}{\end{ALC@g}}

\ifCLASSOPTIONcompsoc
  \usepackage[caption=false,font=footnotesize,labelfont=sf,textfont=sf]{subfig}
\else
  \usepackage[caption=false,font=footnotesize]{subfig}
\fi
\usepackage{url}

\begin{document}
%
\title{VDHLA: Variable Depth Hybrid Learning Automaton and Its Application to Defense Against the Selfish Mining Attack in Bitcoin}
%
%
%
%

\author{Ali~Nikhalat-Jahromi,
        Ali~Mohammad~Saghiri,
        and~Mohammad~Reza~Meybodi
\IEEEcompsocitemizethanks{\IEEEcompsocthanksitem Ali Nikhalat Jahromi, Ali Mohammad Saghiri, and Mohammad Reza Meybodi are with the Department
of Computer Engineering, Amirkabir University of Technology, Tehran,
Iran.\protect\\
E-mail: \{ali.nikhalat,a\_m\_saghiri,mmeybodi\}@aut.ac.ir}
}

\IEEEtitleabstractindextext{%
\begin{abstract}
Learning Automaton (LA) is an adaptive self-organized model that improves its action-selection through interaction with an unknown environment. LA with finite action set can be classified into two main categories: fixed and variable structure. Furthermore, variable action- set learning automaton (VASLA) is one of the main subsets of variable structure learning automaton. In this paper, we propose VDHLA, a novel hybrid learning automaton model, which is a combination of fixed structure and variable action set learning automaton. In the proposed model, variable action set learning automaton can increase, decrease, or leave unchanged the depth of fixed structure learning automaton during the action switching phase. In addition, the depth of the proposed model can change in a symmetric (SVDHLA) or asymmetric (AVDHLA) manner. To the best of our knowledge, it is the first hybrid model that intelligently changes the depth of fixed structure learning automaton. Several computer simulations are conducted to study the performance of the proposed model with respect to the total number of rewards and action switching in stationary and non-stationary environments. The proposed model is compared with FSLA and VSLA. In order to determine the performance of the proposed model in a practical application, the selfish mining attack which threatens the incentive-compatibility of a proof-of-work based blockchain environment is considered. The proposed model is applied to defend against the selfish mining attack in Bitcoin and compared with the tie-breaking mechanism, which is a well-known defense. Simulation results in all environments have shown the superiority of the proposed model. 
\end{abstract}

}

\maketitle

\IEEEdisplaynontitleabstractindextext

%
\IEEEpeerreviewmaketitle

\IEEEraisesectionheading{\section{Introduction}\label{sec:introduction}}

%
%
%
%
\IEEEPARstart{L}{earning} from a psychological perspective is the ability to choose the best solution based on past observations and experiences. This concept makes it possible to design a new sort of intelligent system that can improve its performance step by step on some set of tasks. Each task consists of two parts: learning agent and environment \cite{RSutton2018Reinforcement}. The agent's learning process can be categorized into three major categories; supervised learning, unsupervised learning, and reinforcement learning \cite{RSutton2018Reinforcement}.

\begin{figure}[!t]
    \centering
    \includegraphics[width=0.5\textwidth]{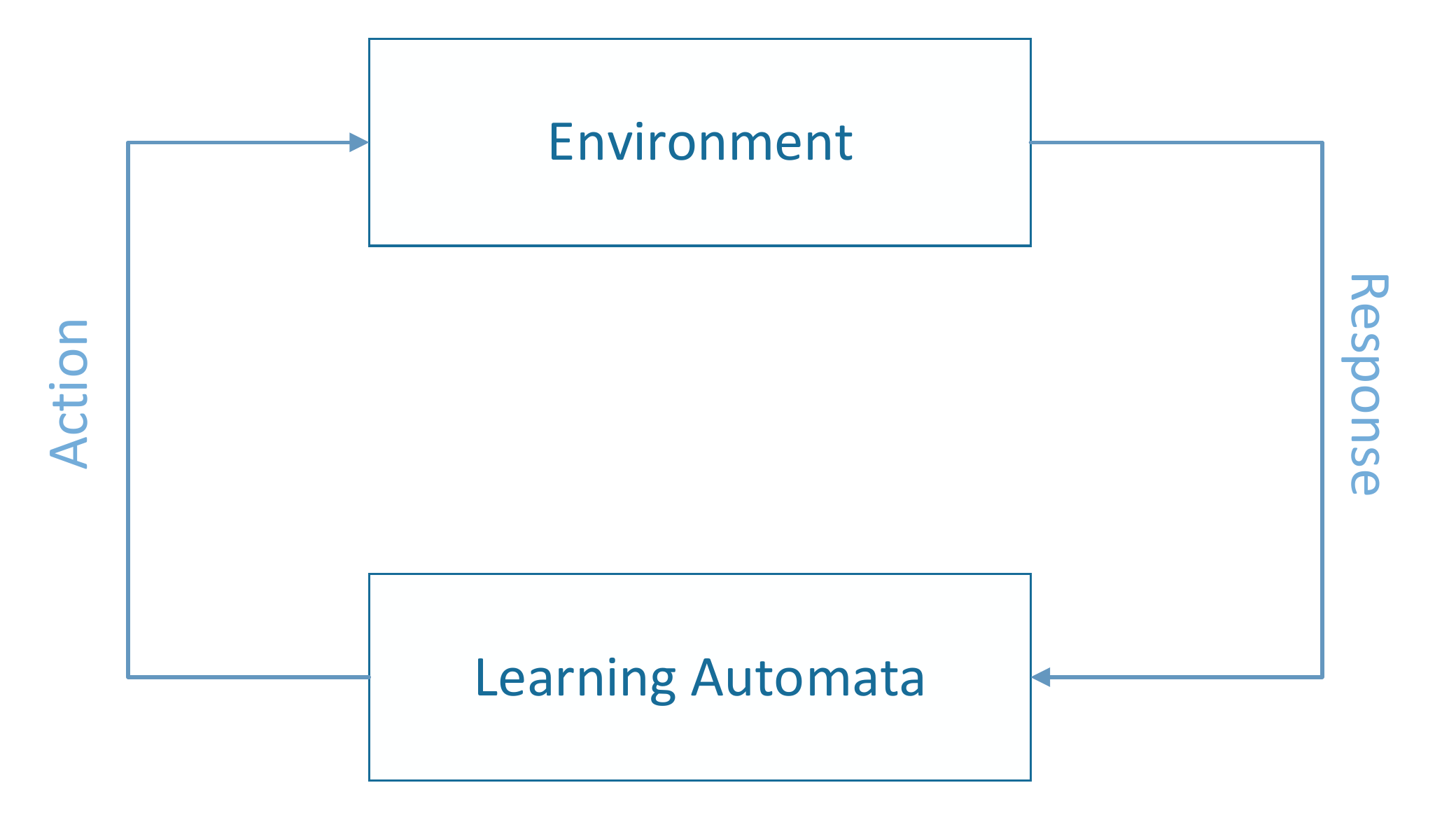}
    \caption{The interaction between learning automaton and the environment}
    \label{fig_automata_environment}
\end{figure}

A learning automaton \cite{RLAIntroduction, RLANetwork} is an adaptive self-organized reinforcement learning model that repeatedly interacts with a random environment to find the optimal action out of a set of offered actions. As shown in Fig \ref{fig_automata_environment}, the learning automaton performs an action according to the probability vector, which contains the probability of choosing each action in the random environment. Interactively, the environment will respond to the learning automaton's chosen action by generating a suitable output. The output, which can be either favorable or unfavorable, belongs to the allowable set of outputs. The learning automaton will update the conducted probability vector by receiving the output \cite{RThathachar2002Varieties}.

The history of learning automaton began in the work of Tsetlin \cite{RLAIntroduction}, which was referred to as a deterministic and stochastic automaton operating in a random environment. Tsetlin's automaton is known as a fixed structure learning automaton (FSLA). Then, the subsequent work was introduced, which was called variable structure stochastic automaton (VSLA) \cite{RLAIntroduction, RLANetwork, RThathachar1985New}. The first time learning automaton (LA) term was announced in the paper by Narendra and Thathchar \cite{RLAIntroduction}. Afterward, learning automaton's field rose to fame. A wide range of learning automaton's application has been reported since the last decade including: 1- Pattern Recognition: Convergence of Tsetlin machine for XOR operator \cite{RJiao2022Convergence}, identity and not operator \cite{RZhang2021Convergence} 2- Neural Networks: Convolutional Regression \cite{RAbeyrathna2021Convolutional}, The similarity between perceptrons and Tsetlin \cite{RSharma2022Equivalence}, Deep neural network \cite{RGuo2019Learning} 3- NLP: Solving pattern recognition tasks using propositional logic \cite{RBhattarai2022Convtexttm}, Semantic representation of words \cite{RBhattarai2023Tsetlin} 4- Optimization problem: Particle swarm \cite{RHashemi2011Note}, multilevel optimization \cite{RNarendra1991Learning} 5- Graph theory: Partitioning problem \cite{RYazidi2021Solving, ROommen2023Learning} 6- Computer Networks: Cognitive radio \cite{RThomas2006Cognitive}, Load balancing \cite{RYazidi2020Achieving} and Wireless \cite{RNicopolitidis2011Adaptive}.

These two major categories of learning automation form the integral part of our study \cite{RLAIntroduction, RLANetwork, RPapadimitriou2004New} Fixed Structure LA (or FSLA) and Variable Structure LA (or VSLA). The first category is a kind of state machine in which the probability of the transition from one state to another state or the action probability of any action in any state is fixed. On the other hand, in the second category, a learning automaton selects an action among a finite set of available actions. It updates its strategy with respect to the assigned probability vector. In our paper, we used a special kind of VSLA, in which the number of available actions varies at each instant. Only a subset of total actions is available in such a learning automaton. This type of LA is a variable action-set learning automaton (VASLA) \cite{RThathachar1987Learning}. This study uses these two categories to create a new kind of LA.

Two significant factors are associated with the performance of the learning automaton in the stochastic environment: speed and accuracy. The first factor relates to the speed rate of convergence to an optimal action, and the second factor is the number of rewards that the learning automaton will get by choosing an action.  

The most asked question about FSLA is how to choose the depth of it. Maximum accuracy is the most important concern in one situation, while convergence speed is vital in the other. This leads us to a trade-off between accuracy and speed because these two attitudes conflict. Sometimes it's inevitable to prefer one to the other, and if each of them is ignored, the disadvantages of FSLA will be apparent. This implies that by adequately large memory depth, its performance can be made as close to optimal desired, and FSLA can get higher rewards. However, the choice of a large depth remarkably decreases its convergence speed \cite{RLAIntroduction, RGholami2021Hla}.  

As we can see, the critical issue about FSLA is the choice of well-suited depth. To the best of our knowledge, there is no previous solution for resolving this acute problem. This is where our new model comes in to solve it. In this study, we propose VDHLA, a novel hybrid learning model, a combination of fixed structure and variable action set learning automaton, to face the dilemma of choosing between accuracy and speed.

In this paper for the first time in the literature we proposed a novel class of FSLA with the ability of learning its internal depth. In other word, the main contribution of our work is how to choose the appropriate depth for FSLA in a fully self-adaptive manner. The process of setting a depth depends on VASLA, which is conducted to learn from the performance (number of rewards and penalties) of FSLA when it reaches the depth of the chosen action. VASLA can set the depth of FSLA by choosing to increase, decrease or leave it unchanged. In this model, two strategies for choosing the appropriate depth are described as follows:

\begin{itemize}
	\item \textbf{Symmetric Variable Depth Hybrid Learning Automaton}: If a unique VASLA is conducted to control the depth of FSLA, it will change the depth of all actions symmetrically, which is called symmetric VDHLA (SVDHLA).
	\item \textbf{Asymmetric Variable Depth Hybrid Learning Automaton}: If each action has its own VASLA to control the related depth, VASLA will control the assigned action's depth asymmetrically, which is called asymmetric VDHLA (AVDHLA).  
\end{itemize}

Several computer simulations are conducted to evaluate the performance of the proposed learning model in two distinct parts. In the first part, the proposed learning automaton is examined purely to prove its performance in stationary and non-stationary environments (Markovian, State-Dependent) concerning the total number of rewards and penalties.
The results are compared with FSLA and VSLA, and the numerical results show the efficiency and the improvements.

The second part of the evaluation examines the effectiveness of the learning model in a practical application. Bitcoin \cite{RNakamoto2008Bitcoin} , a decentralized cryptocurrency, is considered to apply the proposed model. The main drawback of the Bitcoin consensus mechanism is that a kind of attack named selfish mining might threaten the decentralization and fairness capabilities of it. This study uses VDHLA to defend against the selfish mining attack \cite{REyal2018Majority, RJahromi2023Nik}. Then, we examine our defense mechanism with the well-known defense called tie-breaking. Also, the results compare with the defense mechanism which uses FSLA. Analysis of the results shows the superiority of the proposed model in practical applications.

The rest of this paper is organized as follows. The preliminaries related to the essential concepts of learning automaton are given in section 2. In section 3, related work is discussed. In section 4, we explain the proposed learning automaton in detail. Section 5 reports the result of the experiments, and section 6, discusses the paper's limitations and problems. Finally, section 7 concludes the paper.

\section{Preliminaries}
In this section, the required information about the proposed algorithm is given. We summarize information about the learning automaton and its practical subsets used in the following sections. In the rest of this section, at first, an overview of the learning automaton is given. Then, FSLA, VSLA, and VASLA are explained, respectively \cite{RLAIntroduction, RLANetwork}.

A general learning automaton can be presented by a quintuple $<\Phi, \alpha, \beta, F, H>$ where:

\begin{enumerate}
\item $\boldsymbol{\Phi}$ denotes the set of internal states. At any instant $i$, denoted by $\phi_i$, is an element of the finite set:

\begin{equation}
\label{equ_phi}
\Phi = \{\phi_1, \phi_2,..., \phi_n\}
\end{equation}

\normalsize
\item $\boldsymbol{\alpha}$ denotes the set of actions. The output or action of an automaton at the instant $i$, denoted by $\alpha_i$, is an element of the finite set:

\begin{equation}
\label{equ_alpha}
\alpha = \{\alpha_1, \alpha_2,..., \alpha_n\}
\end{equation}

\item $\boldsymbol{\beta}$ denotes a set of values that can be taken by reinforcement signal. The input of an automaton at the instant $i$, denoted by $\beta_i$, is an element of the set, which can be seen in \ref{equ_beta}. $\beta_i$ can be either a finite or infinite set, such as an interval on the real line.

\begin{equation}
\label{equ_beta}
\beta = \{\beta_1, \beta_2,..., \beta_n\}
\end{equation}

\item $\boldsymbol{F}$ denotes the transition function which determines the next state at the instant $i+1$ in terms of the state and input at the instant i(shows in \ref{equ_F}). As a matter of fact, $F$ is a mapping from $\Phi \times \beta \rightarrow \Phi$ and can be either deterministic or stochastic.

\begin{equation}
\label{equ_F}
\phi(i + 1) = F[\phi(i), \beta(i)]
\end{equation}

\item $\boldsymbol{H}$ denotes the output function, which determines the output of the automaton at any instant i in terms of the state at that instant(shows in \ref{equ_H}). Actually, $G$ is a mapping $\Phi \rightarrow \alpha$ and can be either deterministic or stochastic. 

\begin{equation}
\label{equ_H}
\alpha(i) = F[\phi(i)]
\end{equation}

\end{enumerate}
 
\subsection{Fixed-Structure Learning Automaton (FSLA)\label{sec:FSLA}}
Fixed-structure learning automaton \cite{RLAIntroduction, RLANetwork} has wide varieties such as Tsetlin, G$_{2N,2}$, Ponomarev, Krylov, and so on. Since the Tsetlin learning automaton is the most famous form of fixed-structure family, we concentrate on it in our study. The first part of this subsection is dedicated to $L_{2N,2}$, which is two-state Tsetlin. Afterward, $L_{2N,2}$ is expanded to obtain $L_{KN,K}$ as a general form of this family. 

\begin{figure}[!t]
    \centering
    \includegraphics[width=0.5\textwidth]{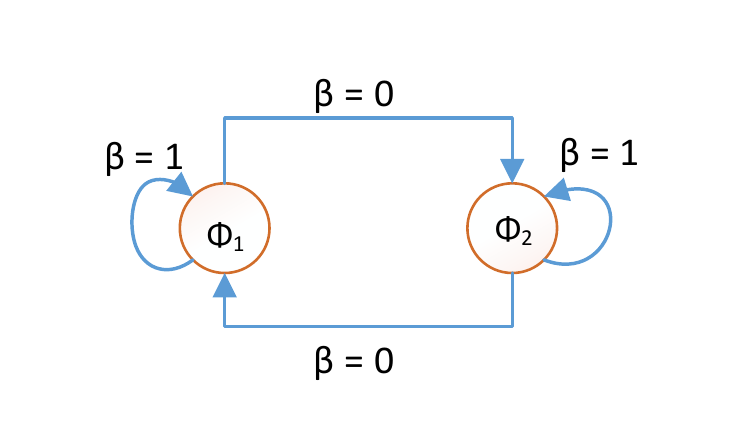}
    \caption{The $L_{2,2}$ learning automaton}
    \label{fig_L_2_2}
\end{figure}

$L_{2,2}$ or two-state learning automaton is the simplest form of fixed-structure learning automaton family with $\phi_1$, $\phi_2$ states, and $\alpha_1$, $\alpha_2$ actions. The reinforcement signal of the environment comes from $\{0, 1\}$ set. The learning automaton will stay in the same state if the corresponding response is favorable ($\beta=1$). On the other hand, if the corresponding response is not favorable ($\beta=0$), the state of the learning automaton will switch. Fig \ref{fig_L_2_2} depicts the structure of $L_{2,2}$.

\begin{figure}[!t]
    \centering
    \includegraphics[width=0.5\textwidth]{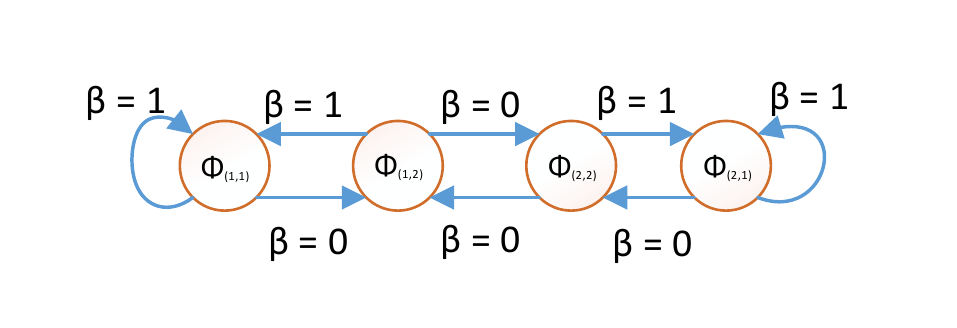}
    \caption{The $L_{2N,2}$ learning automaton with $N=2$}
    \label{fig_L_2N_2}
\end{figure}

Since the depth of $L_{2,2}$ equals 1, there is an action switching upon receiving an unfavorable response. To tackle this problem, the number of states increases from 1 to $N$ for each action. These changes in states will convert $L_{2,2}$ to $L_{2N,2}$. $L_{2N,2}$ has $2 \times N$ states and 2 actions. Such an automaton can incorporate the system's past behavior in its decision rule for choosing an appropriate action. In contrast with $L_{2,2}$, which switches from one action to the other action on receiving an unfavorable response, $L_{2N,2}$ preserves the number of successes and failures received for each action. When the number of failures goes beyond the number of successes by one, the automaton will switch to the other action. $L_{2N,2}$ learning automaton with $N=2$ is shown in Fig \ref{fig_L_2N_2}.

Till now, what has been described is an automaton with just two actions. This idea may generalizes to cases where the automaton can perform $K$ actions. The major difference between an automaton with $K$ actions and an automaton with two actions relates to switching from one action to the next. This learning automaton is called $L_{KN,K}$. In the following paragraph, this learning automaton is explained comprehensively. 

\begin{figure}[!b]
    \centering
    \includegraphics[width=0.5\textwidth]{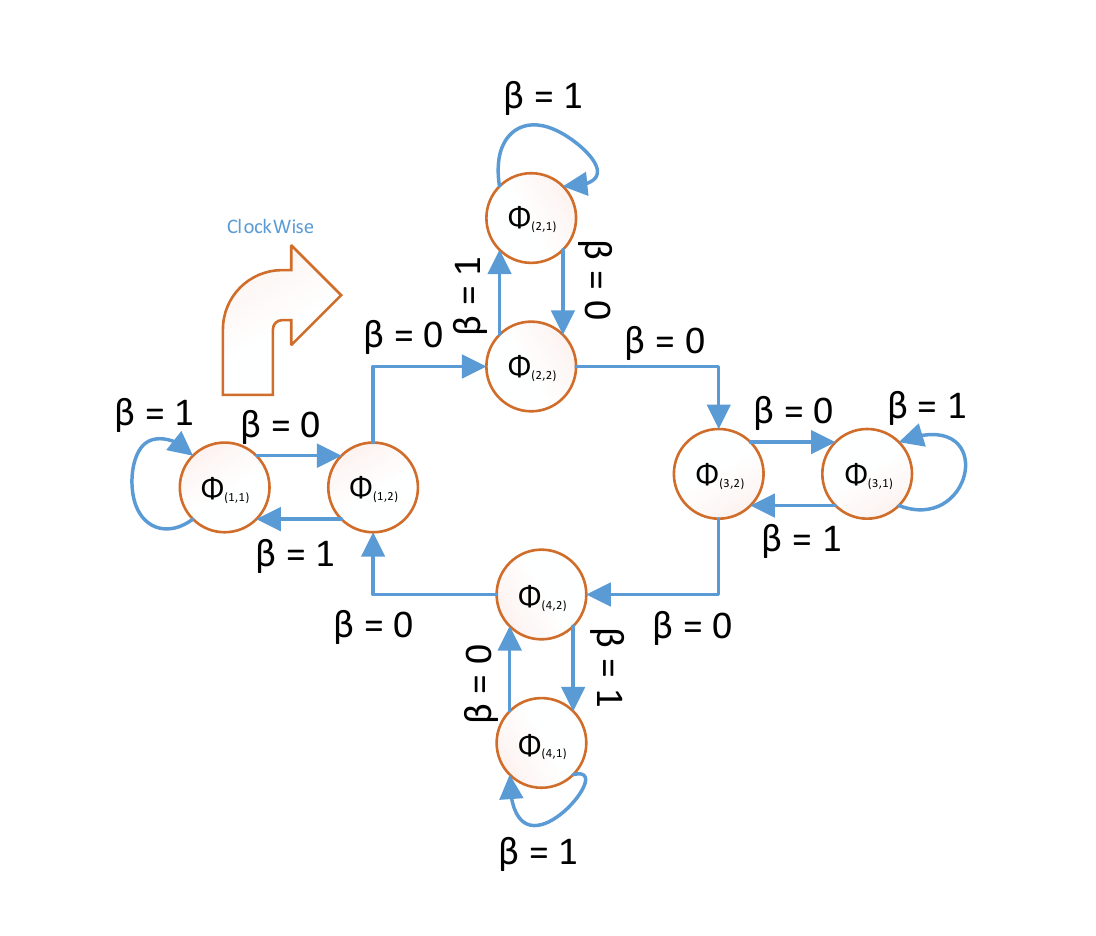}
    \caption{The $L_{KN,K}$ learning automaton with $K=4$ and $N=2$}
    \label{fig_L_KN_K}
\end{figure}

The $L_{KN,K}$ has $K$ actions $\alpha_1, \alpha_2, ..., \alpha_K$ and, $KN$ states $\phi_{(1,1)}, \phi_{(1,2)}, ..., \phi_{(1,N)}, ..., \phi_{(K,N)}$. Each states consists of an ordered pair $(i, j)$ ($ 1\le i \le K $, $1 \le j \le N$). $i$ indicates the action number, and $j$ shows the state number. If the automaton is in a state $\phi_{(i, j)}$, it performs the action $\alpha_i$. By receiving an unfavorable response, the state changes as follows:

\begin{equation}
\label{equ_unfavorable}
   \begin{cases}
      \phi_{(i, j)} \rightarrow \phi_{(i, j + 1)} & (1 \le j \le N - 1) \\
      \phi_{(i, j)} \rightarrow \phi_{(i + 1, j)} & (j = N) \\
   \end{cases}
\end{equation}
furthermore, if the automaton receives a favorable response, the state will change as follows:

\begin{equation}
\label{equ_favorable}
   \begin{cases}
      \phi_{(i, j)} \rightarrow \phi_{(i, j - 1)} & (2 \le j \le N) \\
      \phi_{(i, 1)} \rightarrow \phi_{(i, 1)} & O.W \\
   \end{cases}
\end{equation}
choosing the next action in this automaton is in a clock-wise manner($\phi_{(i, j)} \rightarrow \phi_{(i + 1, j)}$). The state transition graph of the Tsetlin learning automaton with $K=4$ and $N=2$ is shown in Fig \ref{fig_L_KN_K}. 

For example in Fig \ref{fig_L_KN_K}, the automaton is in state $\phi_{(1, 2)}$ as an initial state.  Upon receiving a favorable response ($\beta=1$), automaton moves toward depth, so the state $\phi_{(1, 2)}$ passes to $\phi_{(1, 1)}$. By receiving an unfavorable response ($\beta=0$), the automaton should switch its action ($\phi_{(2, 2)}$ state). Choosing the next action is clock-wise so the automaton will choose action $\alpha_2$.

\subsection{Variable-Structure Learning Automaton (VSLA)}
More advanced models can offer additional flexibility, considering stochastic systems in which the state transitions or action probabilities are updated at every instance using a reinforcement scheme. Such a learning automaton is called a variable-structure \cite{RLAIntroduction, RLANetwork}.

Variable-structure can be defined mathematically by a quadruple $<\alpha, \beta, P, T>$, where $\alpha = \{\alpha_1, \alpha_2, ..., \alpha_r\}$ denotes the finite action set from which the automaton can select the intended action, $\beta = \{\beta_1, \beta_2, ..., \beta_k\}$ denotes the set of inputs to the automaton, $P = \{p_1, p_2, ..., p_r\}$ denotes the action probability vector, such that $p_i$ is the probability of choosing the $\alpha_i$ action, and T denotes the learning algorithm that is used to update the action probability vector in terms of the environment's response at time $t$, i.e $p(t+1) = T[\alpha(t), \beta(t), p(t)]$.

The automaton performs its chosen action on the environment at time $t$. The probability vector conducted within the automaton will be updated by receiving the response.
If the chosen action is rewarded by the environment ($\beta = 1$):

\begin{equation}
\label{equ_probability_vector_rewarded}
	p_j(n+1) = 
   \begin{cases}
      p_j(n) + \lambda_1(1-p_j(n)) & \text{if}\:j = i \\
      (1 - \lambda_1)p_j(n) & \forall\:j \neq i \\
   \end{cases}
\end{equation}
vice versa, if the chosen action is punished by the environment ($\beta = 0$):

\begin{equation}
\label{equ_probability_vector_punished}
	p_j(n+1) = 
   \begin{cases}
      (1 - \lambda_2)p_j(n) & \text{if}\:j = i \\
      \frac{\lambda_2}{r-1} + (1 - \lambda_2)p_j(n) & \forall\:j \neq i \\
   \end{cases}
\end{equation}

In the above equations (\ref{equ_probability_vector_rewarded}, \ref{equ_probability_vector_punished}), $r$ is the number of actions that can be chosen by the automaton. $\lambda_1$ and $\lambda_2$ indicate the reward and penalty parameters that determine the amount of increases and decreases of the action probabilities.

$\lambda_1$ and $\lambda_2$ can have different values. Based on these values, updating schema can be categorized as the following:

\begin{itemize}[\IEEEsetlabelwidth{Z}]

\item $\boldsymbol{L_{R\_P}}$: This updating scheme which is called "linear reward-penalty" comes from the equality of reward and penalty parameter ($\lambda_1 = \lambda_2$). When both are the same, the probability vector of the learning automaton increases or decreases with the monotonic rate. 

\item $\boldsymbol{L_{R\_{\epsilon P}}}$: This updating scheme which is called "linear reward-$\epsilon$ penalty" leads to much greater value of reward parameter in relates to penalty parameter ($\lambda_1 >> \lambda_2$).

\item $\boldsymbol{L_{R\_I}}$: When there is no penalty in an updating scheme ($0 < \lambda_1 < 1, \lambda_2 = 0$), this updating scheme is called "linear reward-Inaction". The probability vector of the learning automaton will not change upon receiving an unfavorable response from the environment.

\item $\boldsymbol{L_{P\_I}}$: If the conducted probability vector in the learning automaton doesn't change by receiving the favorable action, this updating scheme is called "linear penalty-Inaction" ($\lambda_1 = 0, 0 < \lambda_2 < 1, $).

\item \textbf{Pure Chance}: An updating scheme in which there is no penalty and reward parameter($\lambda_1 = \lambda_2 = 0$) is called "Pure Chance". In this updating scheme, the probability vector of the automaton will not change in any conditions.

\end{itemize}

Table \ref{table_updating_scheme} is designed to summarize all of the updating scheme. In this table, $c$ is a constant small value (e.g 0.1 or 0.01). The lower the value c has, the more updating steps exists for the automaton.

\begin{table}[!t]
\renewcommand{\arraystretch}{1.3}
\caption{Updating Scheme for VSLA}
\label{table_updating_scheme}
\centering
\resizebox{\columnwidth}{!}{
\begin{tabular}{cccccc}
\hline
\bfseries Updating Scheme & \bfseries Pure Chance & \bfseries $\boldsymbol{L_{R\_I}}$ 
& \bfseries $\boldsymbol{L_{P\_I}}$ & \bfseries $\boldsymbol{L_{R\_P}}$ & $\boldsymbol{L_{R\_{\epsilon P}}}$\\
\hline\hline
$\lambda_1$ & 0 & $c$ & 0 & $c$ & $c_1$\\
$\lambda_2$ & 0 & 0 & $c$ & $c$ & $c_2(c_1 > c_2)$\\
\hline
\end{tabular}}
\end{table}

\subsection{Variable Action Set Learning Automaton (VASLA)\label{sec:VASLA}}
Under some circumstances, the number of available actions of the learning automaton varies at each instant. To overcome this constraint, a subset of variable-action learning automaton called variable action set learning automaton \cite{RThathachar1987Learning} is defined. Like variable-action, this automaton can be formulated by a quadruple $<\alpha, \beta, P, T>$, where $\alpha$ is a set of outputs, $\beta$ indicates a set of inputs actions, P denotes the state probability vector governing the choice of the state at each stage t, and T is the learning algorithm. The learning algorithm is a recurrence relation used to modify the state probability vector. 

Such a learning automaton has a finite set of r actions denoting $\{\alpha_1,\alpha_2,...,\alpha_r\}$. At each stage t, the action subset $\hat{\alpha}\subseteq{\alpha}$ is available for the learning automaton to choose from. Choosing the elements of $\hat{\alpha}$ is made randomly by an external agency. Both action selection and updating the action probability vector in this learning automaton are described below.

Let $K(t)=\sum_{\alpha_i\in\hat{\alpha}(t)}P_i(t)$ presents the sum of probabilities of the available actions in subset $\hat{\alpha}$. Before choosing an action, the available actions probability vector is scaled as the following equation \ref{scaled_probability}.
    
\begin{equation}
\label{scaled_probability}
\hat{P}_{i}(t) = \frac{p_i(t)}{K(t)} \quad \forall \alpha_i
\end{equation}

The crucial factor affecting the performance of the variable action set learning automaton is the learning algorithm for updating the action probabilities. Let $\alpha_i$ be the action chosen at instant $t$ as a sample realization from distribution $p(t)$. Let $\lambda_1$ and $\lambda_2$ be the reward and penalty parameters, and $r$ denotes the number of available actions. If the learning automaton chooses its intended action ($i=j$), the probability vector will update by the following equation:

\begin{equation}
\label{VASLA_equ_i=j}
{p}_{j}(n+1) = {p}_{j}(n)+\lambda_1\beta(1-{p}_{j}(n))-\lambda_2(1-\beta){p}_{j}(n)
\end{equation}

Conversely, the probability vector for the other actions ($i \neq j$) that are not chosen will update due to the next equation:

\begin{equation}
\label{VASLA_equ_i!=j}
{p}_{j}(n+1) = {p}_{j}(n)-\lambda_1\beta{p}_{j}(n)+\lambda_2(1-\beta)[\frac{1}{r-1}-{p}_{j}(n)]
\end{equation}

Other properties of the VASLA, including updating schemes of reward and penalty parameters summarized in Table \ref{table_updating_scheme}, are the same in relation to the VSLA.

\section{Proposed Algorithm}
In this section, a novel hybrid learning automaton will be introduced comprehensively. This new model of the automaton is called "Variable Depth Hybrid Learning Automaton" and abbreviated to "VDHLA". VDHLA is consisted of $L_{KN,K}$ model of the Tsetlin fixed structure learning automaton and some variable action set learning automata depending on the assortment which the desired automaton belongs to.   

Since the proposed model performs to update the depth of fixed structure learning automaton, from the manner of updating the depth, it can be categorized into two main types:

\begin{enumerate}

\item Symmetric Variable Depth Hybrid Learning Automaton (SVDHLA)
\item Asymmetric Variable Depth Hybrid Learning Automaton (AVDHLA)

\end{enumerate}

These categories lead this section to divide into two major parts. At first, SVDHLA with its related algorithms and examples is presented. Subsequently, it will expand to achieve AVDHLA which is the more advanced model. 

\subsection{Symmetric Variable Depth Hybrid Learning Automaton (SVDHLA)}
SVDHLA or Symmetric Variable Depth Hybrid Learning Automaton is the synthesis of two primary learning automaton model: 1- The fixed structure learning automaton which is used as the base learning model to interact with the outer environment 2- Just one Variable action set learning automaton to handle the depth of the fixed structure.

The fixed-structure learning automaton will perform exactly the same as what is described in section \ref{sec:FSLA}. Changes of states by receiving the reinforcement signal from the outside environment in this automaton follow \ref{equ_unfavorable} (for unfavorable response) and 7 (for favorable response).

To be more specific about the environment, we considered the P-model for modeling the outer environment of the proposed automaton in which the automaton may receives the reinforcement signal from the finite set of 0 or 1.

As stated before, there is only one variable action set learning automaton is conducted to handle the depth of the fixed-structure. When the fixed-structure reaches the outer state (state with the highest number) and intended to leave the chosen action upon receiving the unfavorable response from the environment, the variable action set will active to control the depth gradually.

Choosing the new depth relays on the action-selection of the variable action set among the predefined options:
 
\begin{enumerate}

\item \textbf{Grow}: If the VASLA chooses this action, the depth of all actions will increase by one symmetrically.

\item \textbf{Shrink}: If the VASLA chooses this action, the depth of all actions will decrease by one symmetrically.

\item \textbf{Stop}: If the VASLA chooses this action, the depth of all actions will remain the same as before. 

\end{enumerate}

Another point about the conducted VASLA is the initial probability vector. Since there is no previous learning and there exists three actions, so each item in the probability vector is $\frac{1}{3}$ equally. In this situation, the VASLA chooses its action totally random. The next instances help the automaton to improve itself to set the proper depth.

In the following subsections, input parameters, major units, capabilities , and an illustrative example of the proposed algorithm are presented in depth. Before that, an architecture of the new automaton is depicted in Fig \ref{fig_SVDHLA_architecture}.

\begin{figure}[!t]
    \centering
    \includegraphics[width=0.5\textwidth]{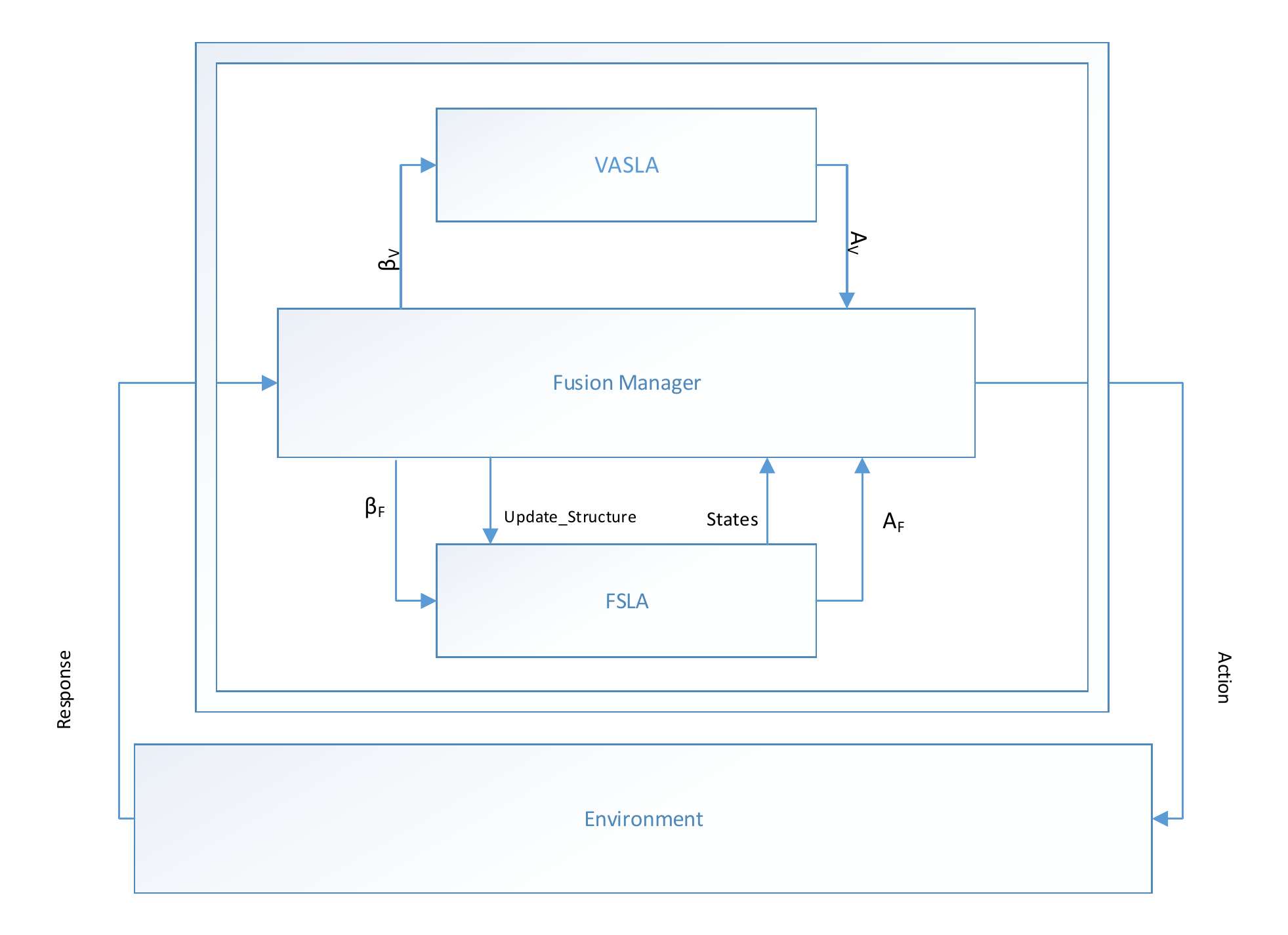}
    \caption{The architecture of SVDHLA}
    \label{fig_SVDHLA_architecture}
\end{figure}

\subsubsection{Input Parameters}
The proposed model can be formulated as $SVDHLA (K, N, \lambda_1, \lambda_2)$. This formalization transforms it to a black-box in which four parameters will receive from the input constructor. These four parameters include $K$: the number of allowed actions, $N$ : the initial memory depth, $\lambda_1$: the reward parameter, and $\lambda_2$: the penalty parameter.

\subsubsection{Units Architecture}  
As we can see in the architecture shown in the Fig \ref{fig_SVDHLA_architecture}, three major units construct the proposed model:

\begin{itemize}

\item \textbf{FSLA}: This unit implements the fixed-structure learning automaton based on $L_{KN,K}$. There exists two input parameters: $K$ and $N$. In addition to two common predefined functions of FSLA, namely $action\_selection$ (for selecting the next action) and $update(\beta)$ (for updating the reinforcement signal), three new functions are defined in this study:

\begin{enumerate}

\item \textbf{$\pmb{is\_depth\_transition}$}:  This function examines the state transition to the depth of the automaton. If the automaton gets reward and the next transition is the depth transition, this function returns true, otherwise it will return false.

\item \textbf{$\pmb{is\_action\_switching}$}: This function investigates the action switching. If the automaton penalizes from the outer environment and wants to change its action, this function returns true, otherwise it will return false.

\item \textbf{$\pmb{update\_depth(new\_depth\_length)}$}: This function can update the depth of all actions in FSLA.

\end{enumerate}

\item \textbf{VASLA}: This unit implements a variable action set learning model with the linear learning algorithm in the S-model environment. In the S-model environment, the reinforcement signal can get value over a span of 0 to 1. VASLA has three input parameters ($K, \lambda_1, \lambda_2$), but $K$ which denotes the number of actions is known and equals to 3 (Grow, Shrink, Stop). Furthermore, this unit has two functions $action\_selection$ and $update(\beta)$ for making a decision and updating the probability vector base on the number of depth transitions respectively.

\item \textbf{Fusion Manager}: This unit is responsible for joining FSLA and VSLA. Actually, it will involve in action-selection and update by implementing $action\_selection$ and $update(\beta)$ functions of SVDHLA. These two functions will be described in detail in the following paragraphs.

\end{itemize}

\paragraph{Action-Selection Process}
Two distinct learning automaton are used together, so there should be a mechanism to synchronize action-selection process. Algorithm \ref{algorithm_SVDHLA_action_selection} shows this mechanism. In this algorithm, the new variable N is defined to demonstrate the depth of the automaton at any time.

\begin{algorithm}[!t]
\begin{algorithmic}[1]
 \renewcommand{\algorithmicrequire}{\textbf{Notation:}}
 \renewcommand{\algorithmicensure}{\textbf{Output:}}
 \REQUIRE {$FSLA$} denotes a fixed structure learning automaton with $L_{KN,K}$,\\
{$VASLA$} denotes a variable action set learning automaton with 3 actions ($'Grow'$, $'Shrink'$, $'Stop'$),\\
{$N$} denotes the depth of FSLA,\\ 
{$L$} denotes the last action made by FSLA,\\
{$V$} denotes the last action made by VASLA \\
\BEGIN {1}
  \IF{FSLA.is\_action\_switching() = True:}
    \IF{N=1:}
    	\STATE V = VASLA.action\_selection([\text{'}Grow\text{'}, \text{'}Stop\text{'}])
    \ELSE
    	\STATE V = VASLA.action\_selection([\text{'}Grow\text{'},\text{'}Stop\text{'},\text{'}Shrink\text{'}])
    \ENDIF
	\SWITCH {$V$}
		\CASE {\text{'}Grow\text{'}}
  			\STATE N = N + 1
  			\STATE FSLA.update$\_$structure(N)
		\ENDCASE
		\CASE {\text{'}Shrink\text{'}}
  			\STATE N = N - 1
  			\STATE FSLA.update$\_$structure(N)
		\ENDCASE
		\CASE {\text{'}Stop\text{'}}
  			\STATE /*Do nothing about structure*/
		\ENDCASE
	\ENDSWITCH
	\STATE L = FSLA.action$\_$selection()
	\RETURN L 
	\ELSE 
		\STATE L = FSLA.action$\_$selection()
		\RETURN L 
  \ENDIF
\END
\end{algorithmic}
\caption{action$\_$selection()}  
\label{algorithm_SVDHLA_action_selection}
\end{algorithm}

In action\_selection function, one of these two situations can be occurred either:

\begin{enumerate}

\item If the automaton stays on the outer state and receive the penalty, action\_switching function will return True value. In this condition, the proposed automaton should decide about the next depth from one of these options: 'Grow', 'Shrink' and, 'Stop'. Now, the automaton is responsible for its action-selection, so the chosen depth should remain the same until the next change in action.

\item Otherwise if the automaton is not on the outer state, VASLA is dormant. In this occurrence, FSLA is responsible for choosing the next action.

\end{enumerate}

We should remark this point about the proposed model that the chosen action from FALSA should return to outer environment. VASLA decisions are just made to find the best depth for FSLA and doesn't relate to the environment.

\paragraph{Updating Process}
After selecting an action and receiving the response from the environment, both constituent LAs should understand result of their chosen action. Algorithm \ref{algorithm_SVDHLA_updating} is designed to do the updating process.

\begin{algorithm}[!t]
\begin{algorithmic}[1]
 \renewcommand{\algorithmicrequire}{\textbf{Notation:}}
 \renewcommand{\algorithmicensure}{\textbf{Output:}}
 \REQUIRE {$FSLA$} denotes a fixed structure learning automaton with $L_{KN,K}$,\\
{$VASLA$} denotes a variable action set learning automaton with 3 actions ($'Grow'$, $'Shrink'$, $'Stop'$),\\ 
{$L$} denotes the last action made by FSLA,\\
{$depth\_counter$} denotes counter for counting depth transition, \\
{$transition\_counter$} denotes counter for counting transitions, \\
{$\beta_v$ denotes the reinforcement signal of VASLA}
\BEGIN {1}
  \IF{beta = 1:}
    \IF{FSLA.is$\_$depth$\_$transition() = True:}
    	\STATE depth$\_$counter = depth$\_$counter + 1
    \ENDIF
    \STATE FSLA.update(1)
    \STATE transition$\_$counter = transition$\_$counter + 1
	\ELSIF{beta = 0:}
		\IF{FSLA.is$\_$action$\_$switching() = True:}
			\STATE $\beta_v$ = depth$\_$counter/transition$\_$counter
			\STATE VASLA.update($\beta_v$)
			\STATE depth$\_$counter = 0
			\STATE transition$\_$counter = 0
		\ENDIF
		\STATE FSLA.update(0)
  \ENDIF
\END
\end{algorithmic}
\caption{update(beta)}  
\label{algorithm_SVDHLA_updating}
\end{algorithm}

In algorithm \ref{algorithm_SVDHLA_updating}, $beta$ notation is used for the received reinforcement signal of environment. $beta = 0$ denotes the performed action was unfavorable, conversely $beta = 1$ shows the favorable action. Beside the $beta$ notation, there new variables are defined:
\begin{enumerate}

\item \textbf{$\pmb{depth\_counter}$}: If the automaton is rewarded by doing an action, this variable will count the number of transitions to the depth state (state with the lowest number). Furthermore, it will reset if the automaton changes its action.

\item \textbf{$\pmb{transition\_counter}$}: This variable will count the number of transitions in one action. If the action changes, it will reset.

\item \textbf{$\pmb{\beta_v}$}: When the automaton wants to change its action, this variable will be used for updating the probability vector of the conducted VASLA. The following equation show how to calculate it.

\begin{equation}
\label{equ_SVDHLA_beta_update}
\beta_v = \frac{depth\_counter}{transition\_counter} 
\end{equation}

\end{enumerate}

Based on the defined variables, \ref{algorithm_SVDHLA_updating} works as follows:
\begin{itemize}

\item If SVDHLA is rewarded ($beta = 1$), one of these two occurrence may happens whether there is depth transition or not:

\begin{itemize}

\item If the depth transition happens (FSLA may reaches the depth state or stays in it), depth\_counter and transition\_counter variables will increase by one. 

\item Otherwise, the depth transition doesn't occur, therefore FSLA just receives the reward and go to the inner state. transition\_counter variable increases due to performing an action and depth\_counter variable doesn't alter. As a result, the probability of being in depth will decrease and the conducted VASLA is prone to be penalized.

\end{itemize}

\item Or else, SVDHLA is penalized ($beta = 0$). This leads the FSLA to be penalized too. In this situation, FSLA moves toward the outer state.  transition\_counter increases and depth\_counter stays the same as before. Consequently, the probability of being in depth decreases and VASLA may be penalized.

\end{itemize}

\subsubsection{An Illustrative Example}
\begin{figure*}[!t]
\centering
\subfloat[Stage one of example]{\includegraphics[width=0.5\textwidth]{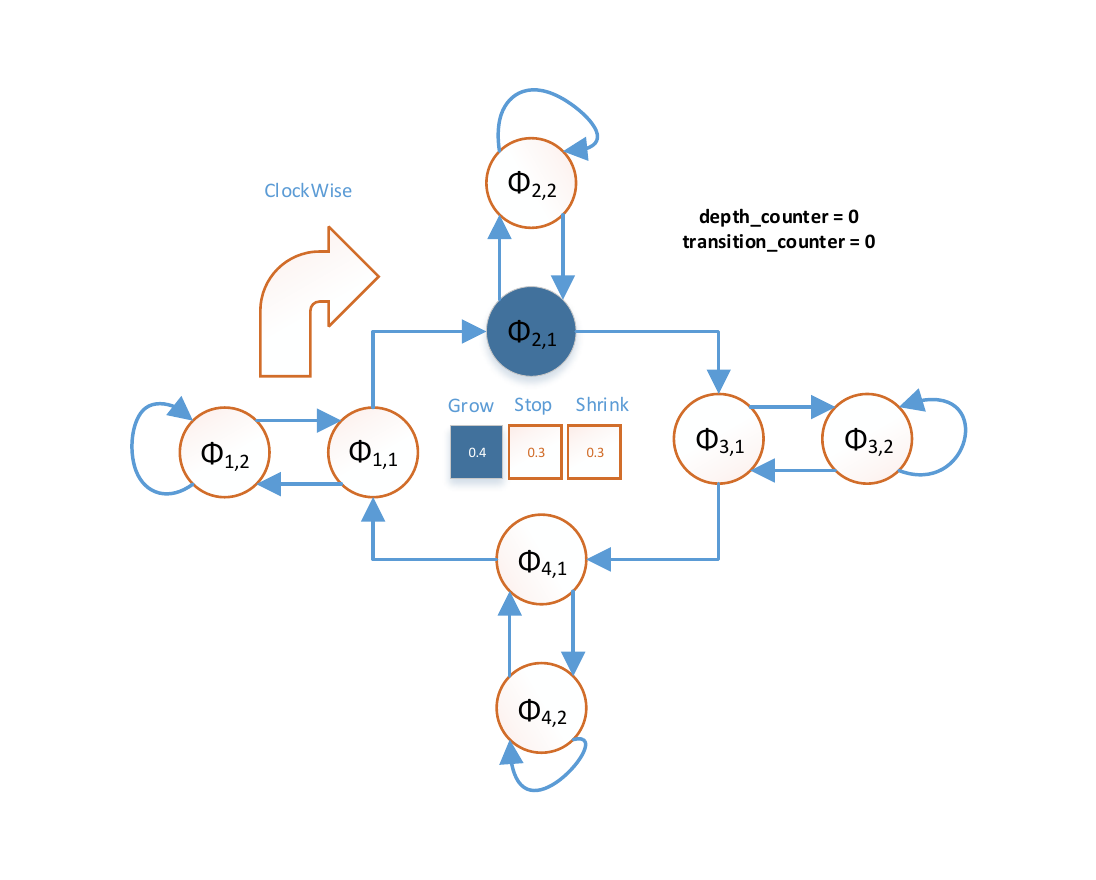}} 
\subfloat[Stage two of example]{\includegraphics[width=0.5\textwidth]{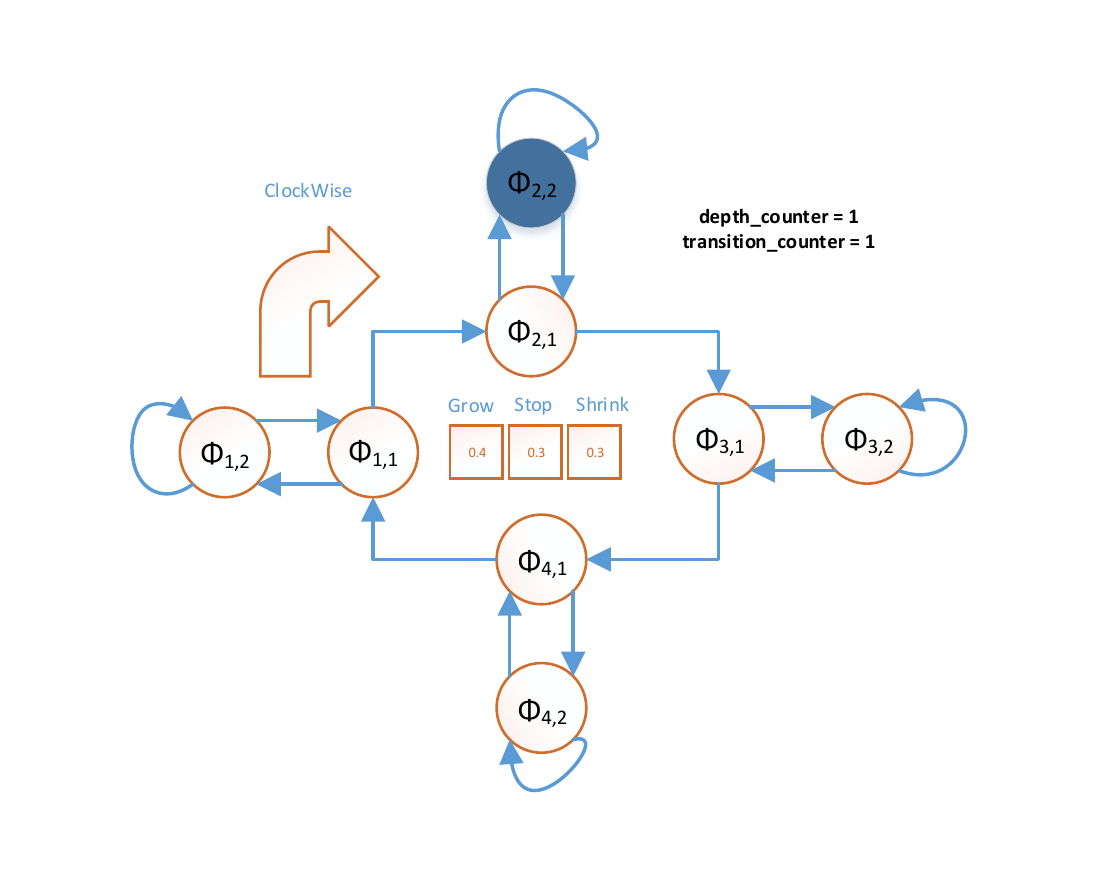}}\\ 
\subfloat[Stage three of example]{\includegraphics[width=0.5\textwidth]{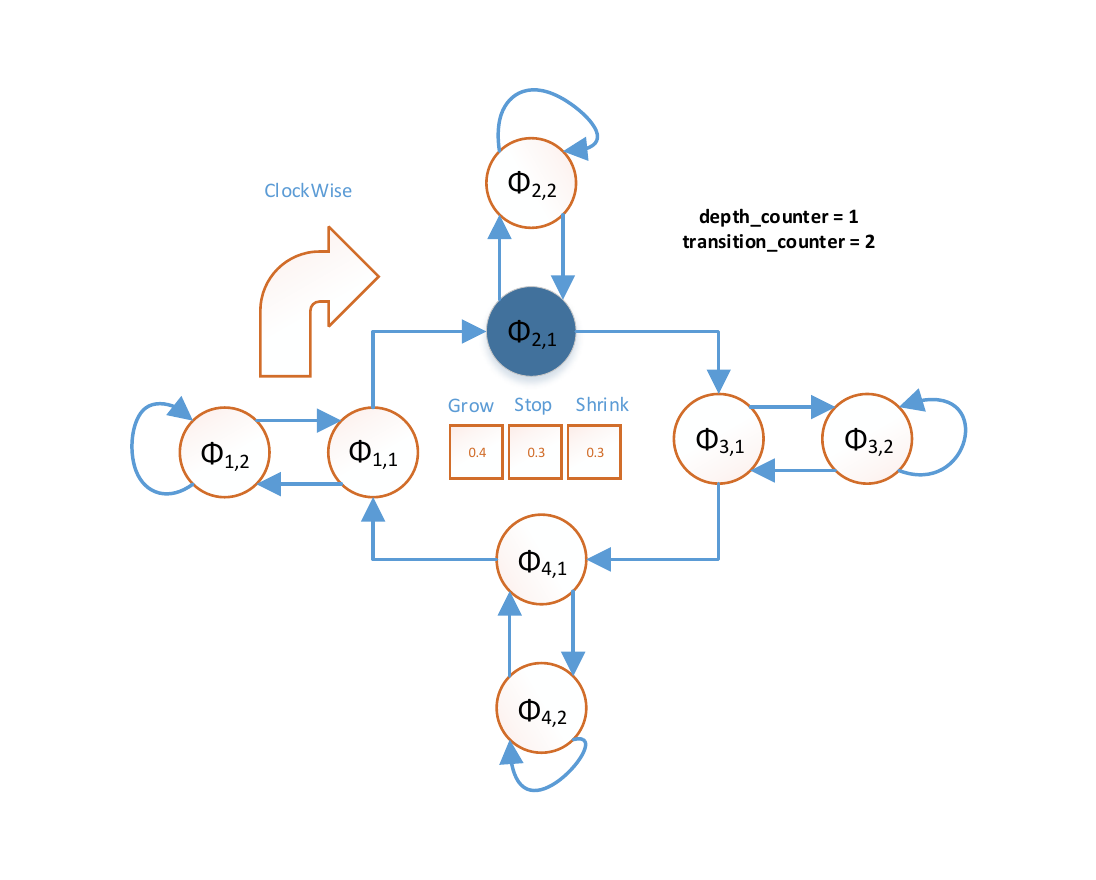}}
\subfloat[Stage four of example]{\includegraphics[width=0.45\textwidth]{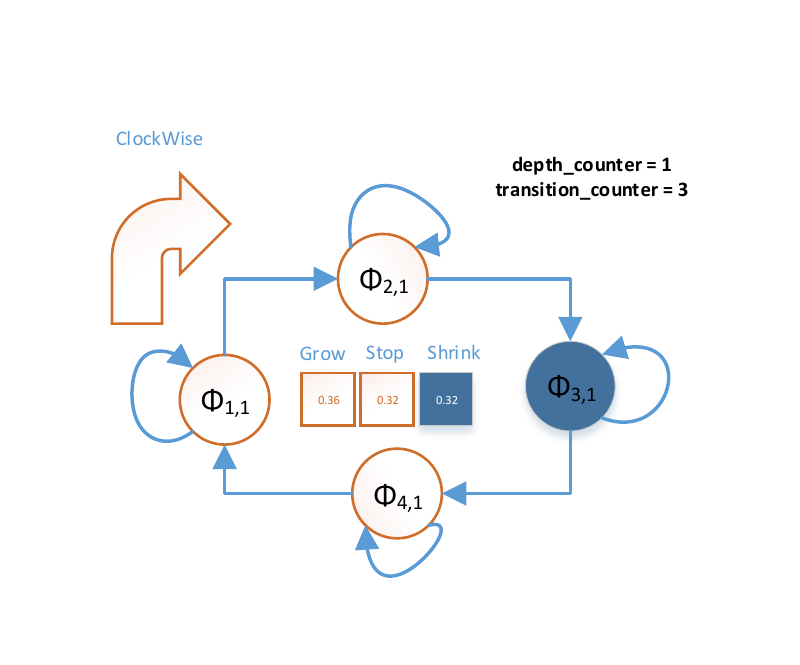}}\\ 
\caption{Four stages of an illustrative example for the SVDHLA }
\label{fig_SVDHLA_example}
\end{figure*}

This subsection devotes to an illustrative example of the proposed automaton. In this example, a $SVDHLA (K=4, N=1, \lambda_1=0.5, \lambda_2 = 0.5)$ with 4 actions, 1 initial state per each action is considered. The conducted VASLA is $L_{R-P}$ with $\lambda_1 = \lambda_2 = 0.5$. The action probability in the probability vector equals to vector $[0.4, 0.3, 0,3]$. Fig \ref{fig_SVDHLA_example} depicts the four stages of the example.

For the starting point, we suppose that the SVDHLA change its action to number 2. The automaton increases its depth and now it is in state $\phi_{(2, 1)}$ like what is shown in Fig \ref{fig_SVDHLA_example}.a. We should remark that two key variables, $transition\_counter$ and $depth\_counter$ are reset, so they are 0.

In the second stage, the SVDHLA performs action number 2. This action is successful and the automaton will be rewarded by the outer environment. Getting reward from the environment will affect both of $transition\_counter$ and $depth\_counter$ variables. Since this reward causes the FSLA moves toward the inner state ($\phi_{(2, 2)}$) and the transition is depth transition, $depth\_counter$ variable will increase by one. On the other hand, a transition occurs , so $transition\_counter$ will increase by one too. This process is depicted in Fig \ref{fig_SVDHLA_example}.b.

In the third stage, the SVDHLA performs action 2 again. In contrast to the previous stage, from the outer environment point of view this action is not successful, so the SVDHLA need to be penalized. This time FSLA moves backward to state $\phi_{(2, 1)}$. This transition is not the depth transition, so $depth\_counter$ remains unchanged. But, the automaton does a transition, therefore $transition\_counter$ variable should increase by one. Fig \ref{fig_SVDHLA_example}.c shows the result of the third stage. 

In the last stage, the SVDHLA performs action number 2. Like the previous stage, the chosen action is unsuccessful. The SVDHLA should get penalty. At this time, since the FSLA is on the on the edge state it will alter its action. Due to clock-wise policy of the FSLA, it should choose action number 3 and, changes its state to $\phi_{(3, 1)}$. We don't have any transitions to the depth, so $depth\_counter$ remains the same as before. But, a transition is done to state $\phi_{(3, 1)}$. This transition will increase $transition\_counter$ by one.

In the SVDHLA, changing the action will activate the VASLA to update the depth of the FSLA. At first, VASLA should get feedback for its previous depth-selection. Reinforcement signal equals to $\frac{depth\_counter = 1}{transition\_counter = 3}$. It shows that the previous action of the VASLA was moderate. Then, the VASLA should choose the next action with the update probability vector. This time, it chooses 'Shrink' action to decrease the depth like what is shown in Fig \ref{fig_SVDHLA_example}.d as a final stage.

This should be mentioned that this process continues until the SVDHLA find the appropriate depth for itself considering the outer environment and its changes.

\subsubsection{Capabilities}
In this part, the capabilities of the SVDHLA is discussed. These capabilities can be itemized as following:

\begin{itemize}

\item SVDHLA can transform to FSLA easily. To do this, reward and penalty parameters should set to 0. This change will disable the conducted VASLA to learn new things.

\item SVDHLA can find the appropriate depth for itself autonomously. This process can optimize the usage of the memory.

\item SVDHLA can transform into the pure-chance automaton. If $\lambda_1 = \lambda_2 = 0$ and initial depth equals to 1, pure-chance automaton will create.

\end{itemize}

\subsection{Asymmetric Variable Depth Hybrid Learning Automaton (AVDHLA)}
After introduction of SVDHLA, the more powerful model of VDHLA family will be presented in this subsection. This new model is called AVDHLA or Asymmetric Variable Depth Hybrid Learning Automaton. AVDHLA is designed to overcome some of the weaknesses of the SVDHLA including updating the depth in a symmetric manner.

Under some circumstances, it is not good to update the depth of all actions in the FSLA. This causes the automaton to get less reward from the outer environment. This problem leads us to design a new learning automaton which can update the depth of each action separately.

The new model consists of: 1-The fixed structure learning automaton 2-A number of the variable action set learning automaton. The number of the VASLA depends on the number of allowed actions.

Since some of the aspects of the proposed automaton is exactly the same as the SVDHLA, we try to summarize the common parts between theses new  proposed learning automata.

The way FSLA and VASLAs are working is the same as what is explained in sections \ref{sec:FSLA} and \ref{sec:VASLA} respectively. Also, the received reinforcement signal of these two primary automaton will be rewarded or punished due to the previous explanations.

The major difference between the proposed learning automaton and SVDHLA is about the manner of altering the depth. In the proposed model, a VASLA is conducted for each action to manage the depth of the action individually.

In the proposed automaton, consider the fixed-structure learning automaton approaches the edge state. The next punishment causes the FSLA to change the action. The conducted VASLA in the unfavorable action will active to select the new depth for itself. This way of selecting help the action to have a proper depth for the next times without affecting other action depth.

Like the SVDHLA, VASLA of each action in the FSLA should choose its action among three predefined options including: 'Grow', 'Shrink', and 'Stop'. Each VASLA in the initial step has a probability vector with equal probability for each action (It is $\frac{1}{3}$ equally).

\begin{figure}[!t]
    \centering
    \includegraphics[width=0.5\textwidth]{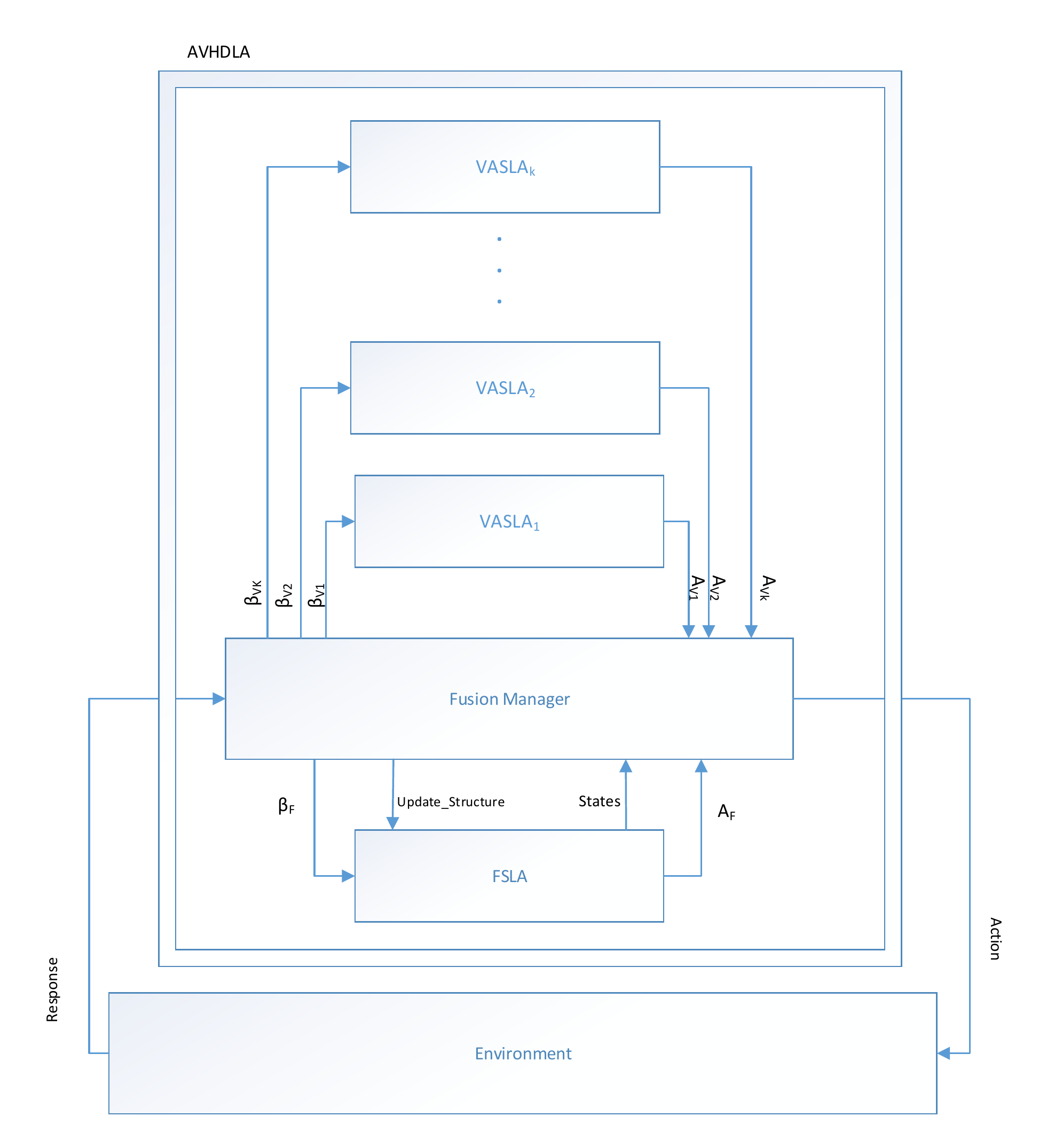}
    \caption{The architecture of AVDHLA}
    \label{fig_AVDHLA_architecture}
\end{figure}

The next subsections will discuss the input parameters, major units architecture, an illustrative example, and capabilities of the proposed learning model. To have a better understanding of the proposed model, an architecture is depicted in Fig.\ref{fig_AVDHLA_architecture}.

\subsubsection{Input Parameters}
The proposed model as a black-box can be created with the constructor $AVDHLA(K, N_1, ..., N_i, ..., N_K, \lambda_1, \lambda_2)$. To make it simpler, we can formulate it to be $AVDHLA(K, N, \lambda_1, \lambda_2)$.

The constructor is composed of these notations: $K$: the number of allowed actions, $N$: the initial depth, $\lambda_1$: the reward parameter, and $\lambda_2$: the penalty parameter. This should be mentioned that if the first constructor is used $N_i$ denotes the initial depth for the action number i $(1 \leq i \leq K)$.

\subsubsection{Units Architecture}
Like SVDHLA, the new model consists of three major units which are shown in Fig.\ref{fig_AVDHLA_architecture}. These units are:

\begin{itemize}

\item \textbf{FSLA}: This unit implements the skeleton of the proposed learning automaton. This skeleton is based on $L_{KN,K}$. Input parameters of the FSLA are: $K$ denotes the number of actions, and $N$ for the initial depth. Similar to SVDHLA, two functions are defined. These functions are named $action\_selection$ and $update(beta)$.

In addition to the previous functions, $is\_depth\_transition$, $is\_action\_switching$, and $update\_depth(new\_depth\_length)$ are defined exactly analogous to the corresponding functions in SVDHLA. Since there is a similarity, relevant descriptions are ignored.  

\item \textbf{$\pmb{VASLA_i}$}: This unit implements a variable action set learning automaton for $i^{th}$ action. This automaton has three actions ('Grow', 'Shrink', and 'Stop') for updating the depth of $i^{th}$ action. Each time the $i^{th}$ automaton is activated performs action $A_{v_i}$, and ${\beta{v_i}}$ will be the received response. It is equipped with $action\_selection$, and $update({\beta{v_i}})$ functions. In the architecture of Fig.\ref{fig_AVDHLA_architecture}, $K$ of these automata are shown $(1 \leq i \leq K)$.

\item \textbf{Fusion Manager}: This unit handles the connection of FSLA and VASLAs. As a matter of fact this unit executes $action\_selection$, and $update(beta)$ functions whenever it is needed. 

\end{itemize}

\paragraph{Action-Selection Process}
To manage the action-selection process of the FSLA and the corresponding VASLA of each FSLA's action together, a new function should be defined. Algorithm \ref{algorithm_AVDHLA_action_selection} shows this function. In compare to Algorithm \ref{algorithm_SVDHLA_action_selection}, two vectors are defined. A vector for the VASLA in which the status of each VASLA is sustained, and the depth vector in which the depth of each FSLA's action is showed at any moment.

\begin{algorithm}[!b]
\begin{algorithmic}[1]
 \renewcommand{\algorithmicrequire}{\textbf{Notation:}}
 \renewcommand{\algorithmicensure}{\textbf{Output:}}
 \REQUIRE {$FSLA$} denotes a fixed structure learning automaton with $L_{KN,K}$,\\
{$VASLA_i$} denotes the $i^{th}$ variable action set learning automaton in the VASLA vector with 3 actions ($'Grow'$, $'Shrink'$, $'Stop'$),\\
{$N_i$} denotes the depth of FSLA's $i^{th}$ action in the depth vector,\\ 
{$L$} denotes the last action made by FSLA,\\
{$V_i$} denotes the last action made by the $i^{th}$ VASLA \\
\BEGIN {1}
  \IF{FSLA.is\_action\_switching() = True:}
    \IF{N$_L$=1:}
    	\STATE V$_L$ = VASLA$_L$.action\_selection([\text{'}Grow\text{'}, \text{'}Stop\text{'}])
    \ELSE
    	\STATE V$_L$ = VASLA$_L$.action\_selection([\text{'}Grow\text{'},\text{'}Stop\text{'},\text{'}Shrink\text{'}])
    \ENDIF
	\SWITCH {V$_L$}
		\CASE {\text{'}Grow\text{'}}
  			\STATE N$_L$ = N$_L$ + 1
  			\STATE FSLA.update$\_$structure(L, N$_L$)
		\ENDCASE
		\CASE {\text{'}Shrink\text{'}}
  			\STATE N$_L$ = N$_L$ - 1
  			\STATE FSLA.update$\_$structure(L, N$_L$)
		\ENDCASE
		\CASE {\text{'}Stop\text{'}}
  			\STATE /*Do nothing about structure*/
		\ENDCASE
	\ENDSWITCH
	\STATE L = FSLA.action$\_$selection()
	\RETURN L 
	\ELSE 
		\STATE L = FSLA.action$\_$selection()
		\RETURN L 
  \ENDIF
\END
\end{algorithmic}
\caption{action$\_$selection()}  
\label{algorithm_AVDHLA_action_selection}
\end{algorithm}

To some extent, the way that the action-selection process of the ADVDHLA works is similar to the SVDHLA. unless, if the AVDHLA wants to change its action, only the depth of the corresponding action in the depth vector will change. This method of depth-selection may increase the power of the new learning automaton vastly. On the other hand, separates the depth-selection from the central controller.

\paragraph{Updating Process}
Each conducted VASLA should activate upon action-selection and when the FSLA decides to change its action, it should update the probability vector. The updating process is done through Algorithm \ref{algorithm_AVDHLA_updating}.

\begin{algorithm}[!b]
\begin{algorithmic}[1]
 \renewcommand{\algorithmicrequire}{\textbf{Notation:}}
 \renewcommand{\algorithmicensure}{\textbf{Output:}}
 \REQUIRE {$FSLA$} denotes a fixed structure learning automaton with $L_{KN,K}$,\\
{$VASLA_i$} denotes the $i^{th}$ variable action set learning automaton in the VASLA vector with 3 actions ($'Grow'$, $'Shrink'$, $'Stop'$),\\
{$L$} denotes the last action made by FSLA,\\
{$depth\_counter_i$} denotes counter for counting depth transition of the $i^{th} action$, \\
{$transition\_counter_i$} denotes counter for counting transitions of the $i^{th}$ action, \\
{$\beta_{v_i}$ denotes the reinforcement signal of the $i^{th}$ VASLA}
\BEGIN {1}
  \IF{beta = 1:}
    \IF{FSLA.is$\_$depth$\_$transition() = True:}
    	\STATE depth$\_$counter$_L$ = depth$\_$counter$_L$ + 1
    \ENDIF
    \STATE FSLA.update(1)
    \STATE transition$\_$counter$_L$ = transition$\_$counter$_L$ + 1
	\ELSIF{beta = 0:}
		\IF{FSLA.is$\_$action$\_$switching() = True:}
			\STATE $\beta_{v_L}$ = depth$\_$counter$_L$/transition$\_$counter$_L$
			\STATE VASLA$_L$.update($\beta_{v_L}$)
			\STATE depth$\_$counter$_L$ = 0
			\STATE transition$\_$counter$_L$ = 0
		\ENDIF
		\STATE FSLA.update(0)
  \ENDIF
\END
\end{algorithmic}
\caption{update(beta)}  
\label{algorithm_AVDHLA_updating}
\end{algorithm}

The major difference between the proposed Algorithm \ref{algorithm_AVDHLA_updating} for updating the internal states and probability vector and Algorithm \ref{algorithm_SVDHLA_updating} which is used for the updating process of the SVDHLA is that the AVDHLA has the specific $depth-counter$, $transition-counter$ and $\beta_v$ for the $i^{th}$ action.

\subsubsection{An Illustrative Example}
\begin{figure*}[!t]
\centering
\subfloat[Stage one of the example]{\includegraphics[width=0.5\textwidth]{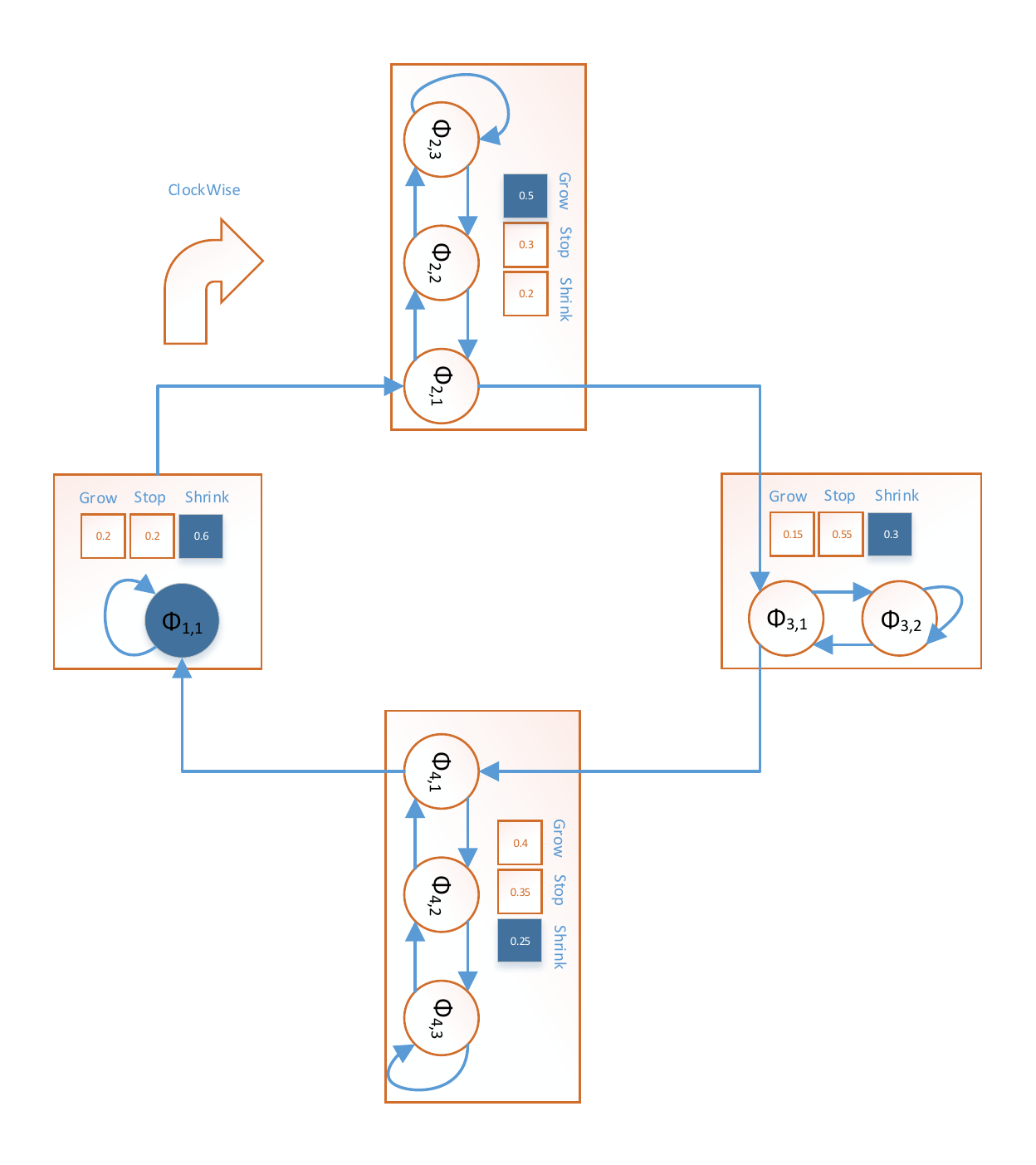}} 
\subfloat[Stage two of the example]{\includegraphics[width=0.5\textwidth]{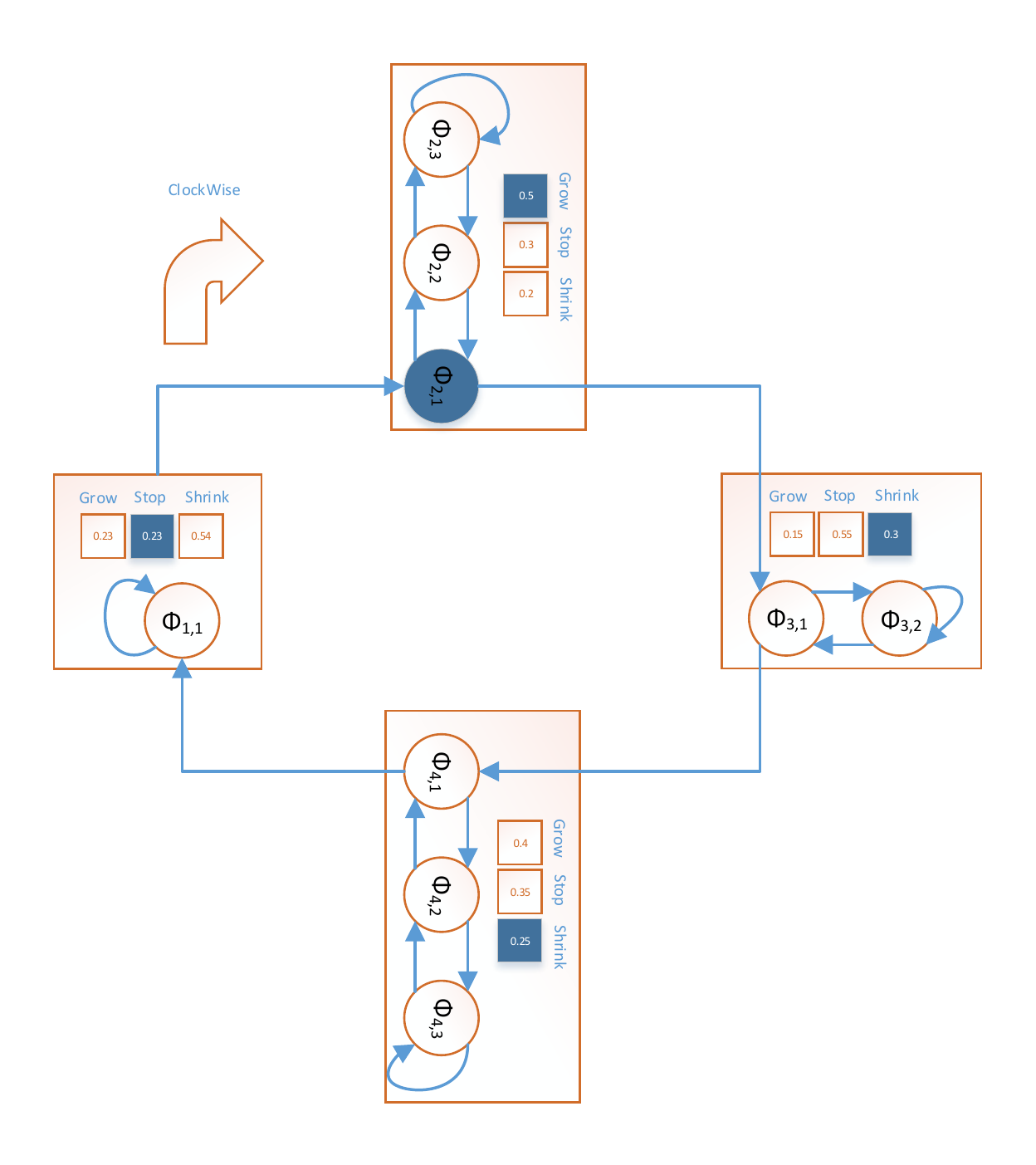}}\\ 
\subfloat[Stage three of the example]{\includegraphics[width=0.5\textwidth]{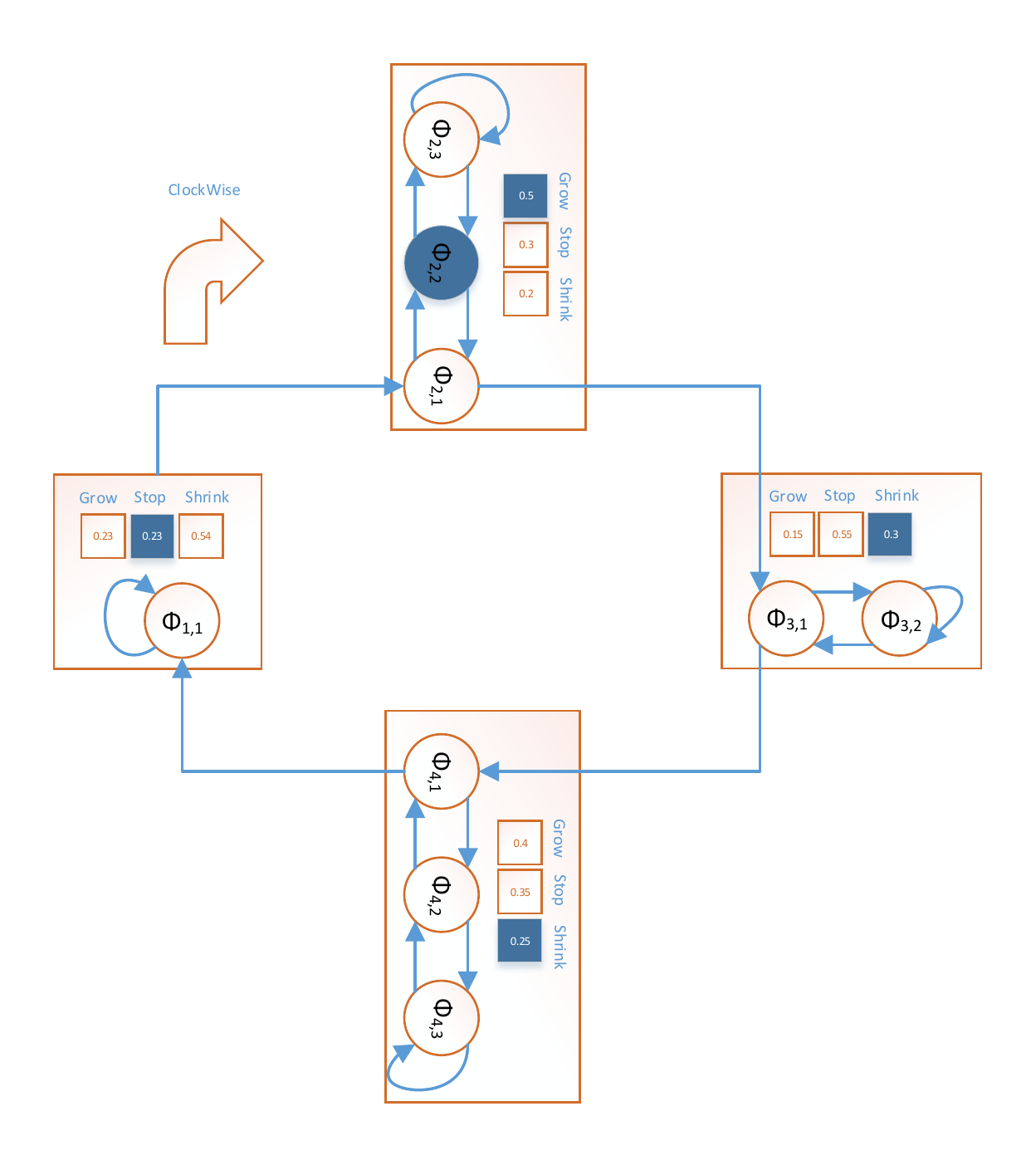}}
\subfloat[Stage four of the example]{\includegraphics[width=0.5\textwidth]{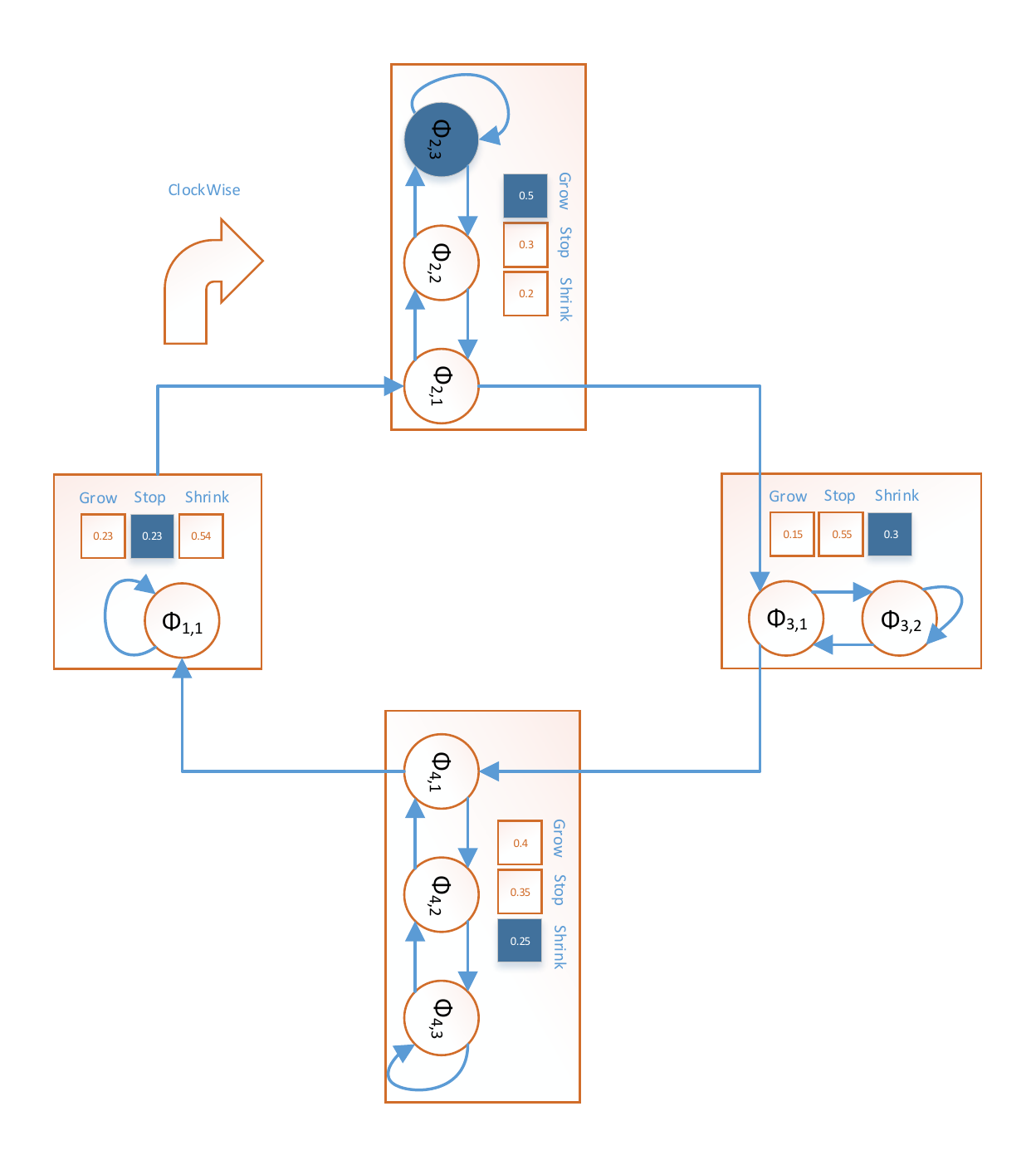}}\\ 
\caption{Four stages of an illustrative example for the AVDHLA }
\label{fig_AVDHLA_example}
\end{figure*}

To provide an example for the proposed automaton, we consider a scenario in which after passing some iterations from the initial state, the AVDHLA came to the state demonstrated in Fig \ref{fig_AVDHLA_example}.

This scenario has four stages. The used AVDHLA is constructed with $SVDHLA (K=4, N_1=1, N_2=3, N_3=2, N_4=3, \lambda_1=0.5, \lambda_2 = 0.5)$. This automaton has 4 actions. Each action has its own depth (1 four action 1, 3 for action 2, 2 for action 3, and 3 for action 4). Also, each action has a specific VASLA with a particular probability vector. The VASLAs are $L_{R-P}$ with $\lambda_1 = \lambda_2 = 0.1$. In Fig \ref{fig_AVDHLA_example}, the last action made by each VASLA is marked. 

At the first stage, the AVDHLA is in the state $\phi_{(1, 1)}$ and the last chosen action by the conducted VASLA denoted by VASLA$_1$ is 'Shrink'. We suppose that the automaton needs to be penalized. This state is shown in Fig \ref{fig_AVDHLA_example}.a.

Since the FSLA is on the $\phi_{(1, 1)}$ which is an edge state, the first step is to investigate the next action for the FSLA. Due to the clock-wise policy of the FSLA, it will choose the state $\phi_{(2, 1)}$.

In this situation, VASLA$_1$ should choose the next action too. This action-selection relates to the next depth of action 1. But, before choosing the next action, the probability vector of the VASLA$_1$ needs to be updated.

Since the depth of action 1 is 1, each transition in this action is the depth transition. Following the assumption, there was not any transitions in this action. So $\beta$ of VASLA$_1$ is 0. This shows that the last action was not good.

The probability vector of the VASLA$_1$ will transit from $[0.2, 0.2, 0.6]$ to $[0.23, 0.23, 0.54]$. This transition is the result of updating with the equation \ref{equ_unfavorable} for 'Shrink' action, and the equation \ref{equ_favorable} for the other actions. These changes are applied to Fig \ref{fig_AVDHLA_example}.b which is related to the stage 2.

The VASLA$_1$ needs to omit 'Shrink' action from its options for choosing the next action. Because, the depth of action 1 equals to 1 and if the 'shrink' action chooses, the next depth will be 0. This leads us to the invalid result. Under such a circumstance, VASLA$_1$ undertakes to choose  among 'Grow' or 'Stop' options. We assume that the 'Stop' option is chosen by the VASLA$_1$.

In the second stage, automaton performs action number 2, and get reward from the outer environment. It will go to the state $\phi_{(2, 2)}$. The $transition-counter$ variable increases by one, but the action was not depth-transition. So, the $depth-transition$ remains 0.

In the third stage, the AVDHLA performs action number 2 again. This time action is successful too. The FSLA changes the state from $\phi_{(2, 2)}$ to $\phi_{(2, 3)}$. The $transition-counter$ will get value 2, but like before $depth-transition$ still doesn't have any changes in its value. Fig \ref{fig_AVDHLA_example}.c, displays the third stage.

In the last stage, again the FSLA deserves the reward from the environment, hence update its state to $\phi_{(2, 3)}$. By contrast to the stage 3, both of the $transition-counter$ and $depth-transition$ will increase. Since there is a transition and that transition is the depth transition kind. Fig \ref{fig_AVDHLA_example}.d depicts the status of the automaton in the last stage.

This scenario will continue, and as it can be seen, the proposed automaton can synchronize itself with the outer environment to get the highest possible number of rewards.

\subsubsection{Capabilities}
The AVDHLA has wonderful capabilities that is explained in this subsection. These capabilities are defined as follows:

\begin{itemize}

\item AVDHLA can delegate the duty of depth-selection to the individual VASLA which is installed in each depth. This novel idea can expel the central control which is applied in the SVDHLA.

\item AVDHLA can transform to FSLA easily. like SVDHLA, $\lambda_1$, and $\lambda_2$ should be equal and having value 0.

\item Choosing the appropriate depth can help the AVDHLA to decrease the usage of memory. In opposition to the SVDHLA, this decrement is so smarter.

\item AVDHLA can convert into the pure-chance automaton by $\lambda_1 = \lambda_2 = 0$, and choosing 1 for $N_1$ to $N_2$.

\end{itemize}

\section{Evaluation}
This section is designed for the evaluation of the VDHLA. To evaluate the correctness and efficiency of the proposed learning automaton, we first introduce the environments, then the metrics of the evaluation are presented. At last, various experiments had been designed to assess the new learning automaton and its comparison with other kind of learning automata. Our code is available on Github \footnote{\url{https://github.com/AliNikhalat/LearningAutomata}} to help the upcoming researches.

\subsection{Environment}
The outer environment can be formulated as a triple $<\alpha, \beta, C>$ in which \cite{RLAIntroduction}:

\begin{enumerate}

\item $\alpha = \{\alpha_1, \alpha_2, ..., \alpha_r\}$ denotes a finite set of the inputs.

\item $\beta = \{\beta_1, \beta_2\}$ denotes a binary set of the outputs. For mathematical convenience $\beta_1$ and $\beta_2$ are chosen to be 0 and 1 respectively.

\item $C = \{c_1, c_2, ..., c_r\}$ denotes the penalty probability in which the element $c_i$ may characterizes the environment. $c_i$ maps to the action $\alpha_i$. In the same way, $D = \{d_1, d_2, ..., d_r\}$ denotes the reward probability. Each element of D set can be computed by $d_i = 1 - c_i$.

\end{enumerate}

The input $\alpha_n$ applied to the environment at the discrete time $t = n (n = 0, 1, 2, ...)$. The output $\beta(n)$ of the environment which can be one of the following either:

\begin{itemize}

\item $\beta(n) = 1$: The environment responses positively to the applied action and as a result the automaton deserves a reward.

\item $\beta(n) = 0$: The environment responses negatively to the applied action and consequently the automaton will punish.

\end{itemize}

By the above definitions, the negative response from the environment can be summarized as a following mathematics equation:

\begin{equation}
\label{eq_environment_nagative_response}
Pr(\beta(n) = 0 | \alpha(n) = \alpha_i) = c_i (i = 1, 2, ..., r)
\end{equation}

Meanwhile, the positive response can be shown as a following equation:
\begin{equation}
\label{eq_environment_nagative_response}
Pr(\beta(n) = 1 | \alpha(n) = \alpha_i) = 1 - c_i (i = 1, 2, ..., r)
\end{equation}

From the operational standpoint, the environment can be categorized into the \emph{stationary} and \emph{non-stationary} environments. Furthermore, non-stationary environment can be divided into either \emph{markovian switching} and \emph{state dependent} environments.

To have a better overview of the environment classification, Fig visualizes this hierarchically classification.

\begin{figure}[!t]
    \centering
    \includegraphics[width=0.5\textwidth]{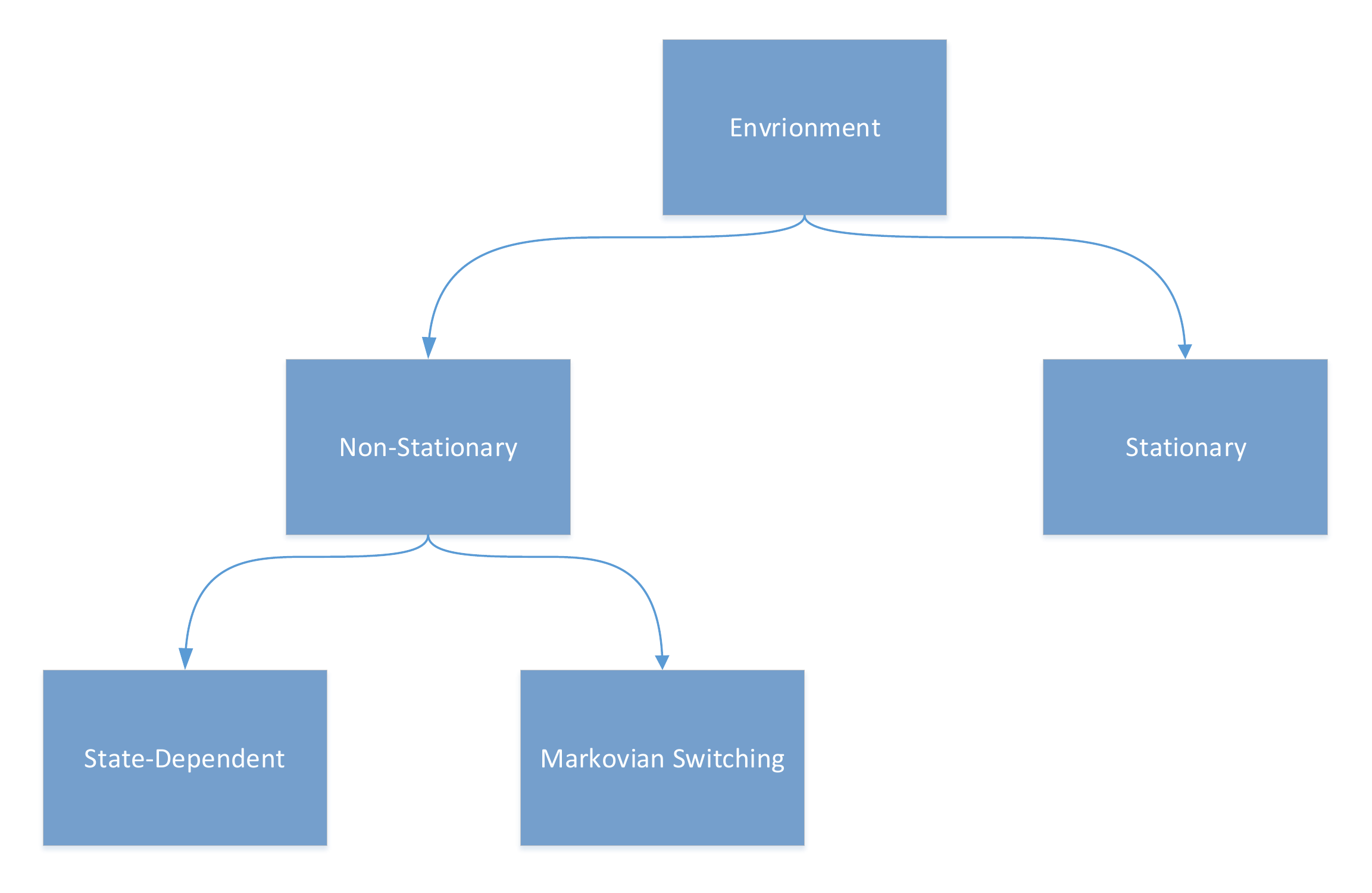}
    \caption{The hierarchical classification of the environment}
    \label{fig_SVDHLA_architecture}
\end{figure}

\subsubsection{Stationary Environment \label{section:environment}}
Referring to the previous explanations in the above section, $c_i$ denotes the penalty probability in the time $n$. If this probability do not change over the time, the environment can be considered to be the stationary environment \cite{RLAIntroduction, RLANetwork}.

\subsubsection{Non-Stationary Environment \label{section:non_stationary_environment}}
Contradiction to the stationary environment, whenever $c_i$  varies over the time, the environment is non-stationary \cite{RLAIntroduction}.

If a learning automaton with a fixed strategy is used in such an environment, it may becomes useless or performing with a lot of penalties. To tackle this problem, the automaton should have enough flexibility to track the changes of the environment.

Looking precisely into the non-stationary environments may leads us to this point that $c_i(n)$ maybe constant over an internal $[n, n + T - 1]$, and switches to a new value at the time $n + T$.

Explicitly, every penalty sets can map to a new environment. This mapping brings us to this result that the learning in a time varying environment corresponds to the learning in the multiple random environments.

To analyze the problem, some constraints should be considered on the time variation of $c_i(n)$. How the environment evolves and the previous information of the designer about the environment may helps the learning process.

Analysis of non-stationary environments may be so complex. For sake of simplification, $c(n)$ can be presumed to have a finite number of values. This leads us to this condition that the automaton performs in one of a finite set of stationary environments $E_i (i = 1, 2, .., d)$ at any instants. If the automaton switches the current environment $E_i$ to any member of the set at the next instant, it will follow a predefined rules.

Assume a condition that the automaton consists of $d$ sub-automaton and each sub-automaton  denoted by $A_i$ corresponds to the environment $E_i$. In this situation, when the $A_i$ automaton operates in the $E_i$ environment, the probability vector of $A_i$ will update according to the response of environment $E_i$.

This way of thinking about the non-stationary environments can help us to analyze each environment individually. The complex non-stationary environment divides into the finite numbers of stationary environments which can study separately.

Two major weaknesses of this model of the simplification is:

\begin{enumerate}

\item Possibility of  assigning each $A_i$ sub-automaton to the $E_i$ environment is doubtful. This can be feasible only when the total number of environments in the environment set is not too large. Beside that, since the sub-automaton $Ai$ is updated only when the environment $E_i$ is active, updating in this way occurs infrequently. So the corresponding action probabilities might suffers from this approach.

\item We have this assumption that the automaton has total awareness of the environment in which it operates at every instant. This obligation will not satisfy in the practical environments.

\end{enumerate}

By knowing the above problems about the non-stationary environments, two classes of the non-stationary environments (MSE, State-dependent) have been described in the following paragraphs. In each case the automaton is presumed to operate in one of a finite number of environments $E_1, E_2, ..., E_d$.

\paragraph{MSE}
MSE or \emph{Markovian Switching Environment} is an environment where each $E_i (1 \leq i \leq d)$ environment is itself states of a Markov chain. If the chain is ergodic an automaton that is connected to such an environment will be in each of the component environments with a fixed probability corresponding to the asymptotic probability distribution of the ergodic chain \cite{RLAIntroduction, RLANetwork}.

\paragraph{State-Dependent Environment}
If the environment $E_i (1 \leq i \leq d)$ has finite number of sub-environments, and the state of each environment alters with the stage number n, such a composite environment is referred to a \emph{State-Dependent Environment} \cite{RLAIntroduction, RLANetwork}.

In this environment, changes of the states may relay on the $n^{th}$ stage explicitly or implicitly. Generally, this kind of environments is described by the $c_i (1 \leq i \leq r)$ penalty probabilities.

State-dependent environment has three well-known models (A, B, C). In this study, B, C models are ignored. Hence model A is explained subtly.

In the A model of such an environment, if the automaton performs action $\alpha_i$ in the $n^{th}$ stage, two events will occur:

\begin{enumerate}

\item The $c_i$ probability of the $i^{th}$ action will increase.
\item The $c_i$ probability of the other actions ($j \neq i$) decreases.

\end{enumerate}  

These occurrences implies that the performed action becomes worse for the next stages, while other actions improve with time.

From mathematical standpoint, this environment is described as below equations:

\begin{equation}
\label{equ_environment_state_dependent}
   \begin{cases}
      c_i(n) = c_i(n) + \theta_i(n) \\
      c_j(n) = c_j(n) - \phi_j(n) & j \neq i \\
   \end{cases}
\end{equation}

In the above equation, $\theta_i(n)$ and $\phi_j(n)$ $(i, j = 1, 2, ..., r)$ both are non-negative functions of the $n^{th}$ stage. Generally, $\theta_i(n)$ and $\phi_j(n)$ might be the functions of $p(n)$ as well as $c_i(n)$.

For making the model simpler, $\theta_i(n)$ is considered to be the constant value without any dependencies to the time $n$ . The following equation is used for $\theta_i(n)$:

\begin{equation}
\label{equ_environment_state_dependent_simplification_theta}
   \theta_i(n) = 
   \begin{cases}
      \theta_i & c_i(n) + \theta_i(n) \leq 1 \\
      1 - c_i(n) & o.w \\
   \end{cases}
\end{equation}
And $\phi_j(n)$ is defined to be constant too. Relation of $\phi_j(n)$ to the time $n$ is ignored. The Simplified equation is presented below: 

\begin{equation}
\label{equ_environment_state_dependent_simplification_phi}
   \phi_j(n) = 
   \begin{cases}
      \phi_j & c_j(n) - \phi_j(n) \geq 0 \\
      1 - c_j(n) & o.w \\
   \end{cases}
\end{equation}

\subsection{Metrics \label{section:evaluate_metric}}
Metrics for the evaluation of the proposed learning automaton are introduced here. These metrics will be applied to show the efficiency and the performance of the VDHLA due to the designed experiments in the presented environments.

In the following items, these metrics are defined \cite{RLAIntroduction, RLANetwork}:

\begin{itemize}

\item \textbf{Total Number of Rewards (TNR)}: This evaluation metric shows the total number of rewards that the automaton is received by the environment ($\beta = 1$). In the $i^{th}$ iteration, if the automaton gets the positive feedback from the environment, TNR will be increased by 1. Generally, the higher values of TNR shows the better performance of the learning automaton. The cumulative relation of the TNR is shown in the following equation:

\begin{equation}
\label{equ_environment_TNR}
   TNR_i = \sum_{j \leq i} \beta_j
\end{equation}

\item \textbf{Total Number of Action Switching (TNAS)}: This metric is used to count how many times in an environment, the automaton should change the chosen action. This metric is so important. Because, some of learning automata cost a lot for changing the action.
As a matter of fact, TNAS$_i$ shows the aggregation of the action switching of the automaton at the $i^{th}$ stage. The equation of TNAS$_i$ is presented in the following:

\begin{equation}
\label{equ_environment_TNAS}
   TNAS_i = \sum_{j \leq i} s_j
\end{equation}

We should remark that the lower number of TNAS is better for the automaton.

\item \textbf{Probability of Choosing the Favorable Action ($P(\alpha_i)$)}: If the automaton has a situation in which it can choose between more than one action, and one them is favorable, the probability of choosing the favorable action should be considered as an evaluation metric.

This parameter equals to the total number of times that $\alpha_i$ is chosen divided by the total number of times that other actions have been chosen.

\begin{equation}
\label{equ_environment_Probability}
   P(\alpha_i) = \frac{Number\:of\:times\:that\;\alpha_i\:is\:chosen}{Total\:number\:of \:action\:selection}
\end{equation}

The higher values for $P(\alpha_i)$ shows that the automaton moves toward the best possible action. Hence, this parameter has an important role in the experiments.

\end{itemize} 

\subsection{Experiments Results}
We have designed lots of experiments to test the proposed automaton in the all possible aspects. These experiments are conducted in various environments which are introduced in the section \ref{section:environment}.

For the first time in the literature, an automaton is tested in the \textit{Markovian Switching} and \textit{State-Dependent} environments. This novel approach will help researchers to see strengths and weaknesses of the learning automaton in such environments.

Since three kinds of environments had been introduced, three major categories of the experiments are designed. These categories are:

\begin{enumerate}

\item \textbf{Experiment1 (Ex1)}: The main goal of this experiment is evaluating the proposed automaton in an environment with the fixed penalty probability ($c_i$).

\item \textbf{Experiment2 (Ex2)}: This category of the experiments is dedicated to test the proposed automaton in the markovian switching environments with different properties for each environment.

\item \textbf{Experiment3 (Ex3)}: In this category of the experiments, the proposed automaton is assessed in different kinds of the state-dependent environments.

\end{enumerate}

Finally we should mention this point that due to the huge numbers of experiments, a small portion of them which are more important than the others are excerpted in this part. Rest of them are explained in the appendixes comprehensively.  

\subsubsection{Experiment 1}
This experiment is designed to evaluate the proposed learning automaton in the stationary environments with the fixed penalty probability. To have a better overview on the proposed automaton in the stationary environments, Experiment 1 is divided in three sub-experiments. 

\begin{itemize}

\item Experiment 1.1 for investigating the learning capability of the automaton.

\item Experiment 1.2 is designed to compare the proposed automaton with FSLA.

\item Experiment 1.3 studies the proposed automaton versus VSLA.

\item Experiment 1.4 undertakes to compare SVDHLA and AVDHLA as two major members of VDHLA family.

\item Experiment 1.5 dedicates to AVDHLA for evaluating it with various values of the initial depth.

\end{itemize}

\paragraph{Experiment 1.1} 
The learning capability of the proposed automaton should be considered before doing any other experiments. To do this, Experiment 1.1 is designed to see whether the proposed automaton performs better than \textit{Pure Chance Automaton (PCA)} which chooses an action randomly in each instant or not.

To examine the learning capability of both SVDHLA and AVDHLA, an environment with considering probability of being rewarded for corresponding actions are designed. The probability of action 1 to get reward from the environment is 0.1 and the second one is 0.9. 

For sake of explanation, the first action (action $\alpha_1$) has the probability of 0.1. This means that the automaton in this environment will be rewarded in 10\% of times by choosing $\alpha_1$. Meanwhile, choosing the second one (action $\alpha_2$) brings the performed action of the automaton with reward in 90\% of times. 

In such an environment, the automaton must incline to the second action more than the first one to get higher number of rewards and decreases its number of action switching.

This experiment is done for SVDHLA and AVDHLA in 1000 iterations separately. Both of them have two actions and the inner VASLAs are $L_{R\_{\epsilon P}}$ with $\lambda_1 = 0.1$ and $\lambda_2 = 0.01$. 

\begin{figure}[!t]
\centering
\subfloat[TNR of SVDHLA]{\includegraphics[width=0.25\textwidth]{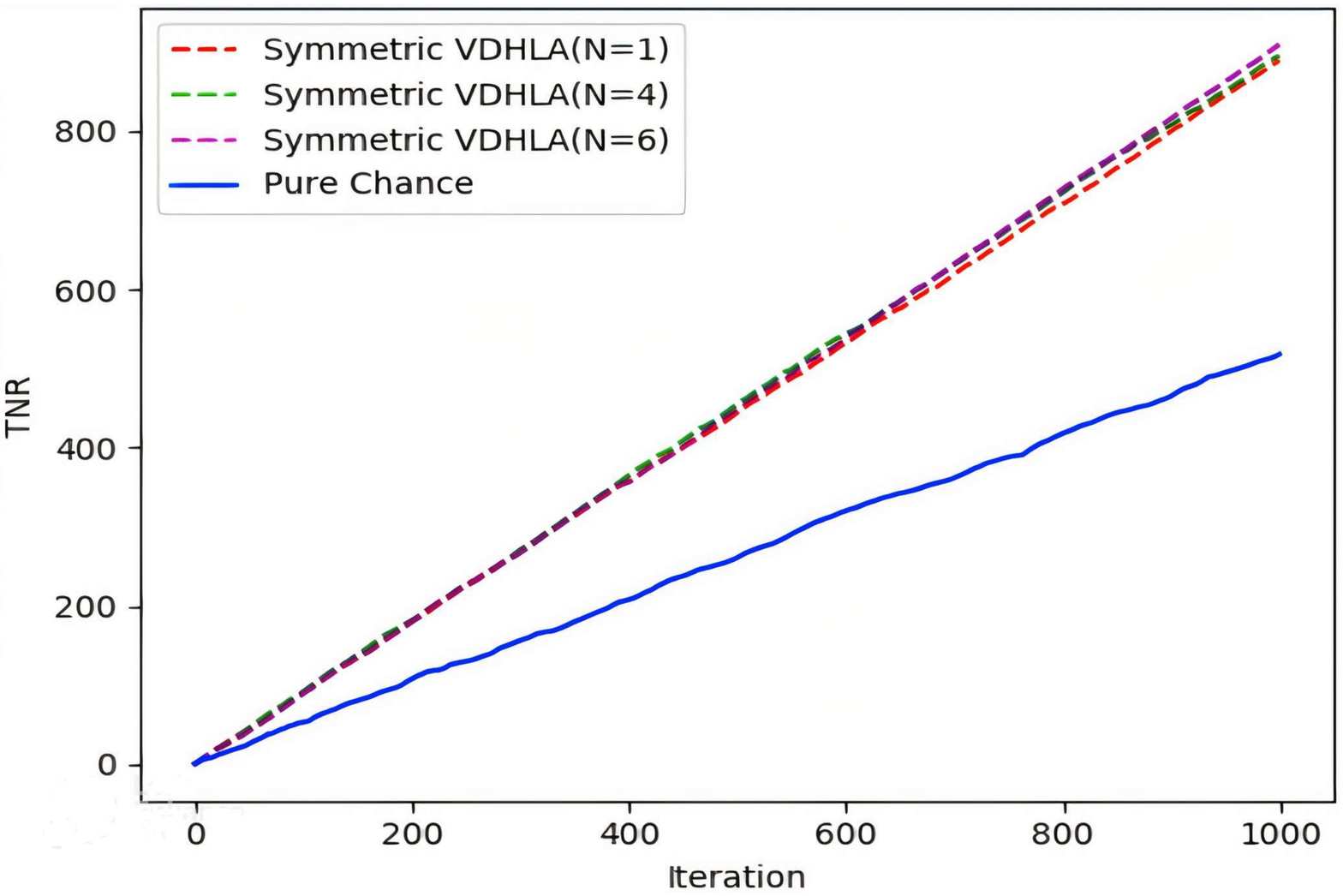}} 
\subfloat[TNAS of SVDHLA]{\includegraphics[width=0.25\textwidth]{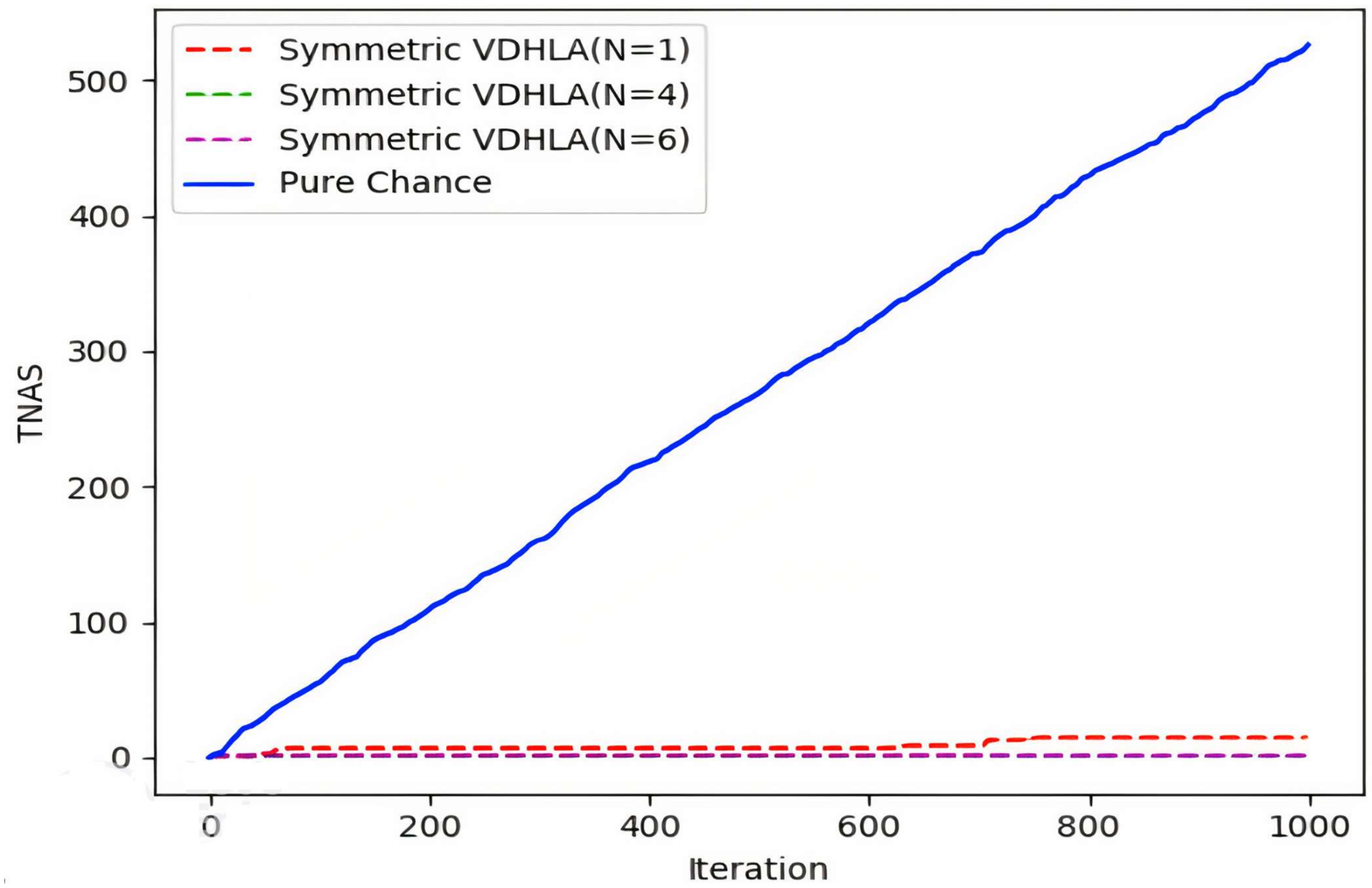}}\\ 
\subfloat[TNR of AVDHLA]{\includegraphics[width=0.25\textwidth]{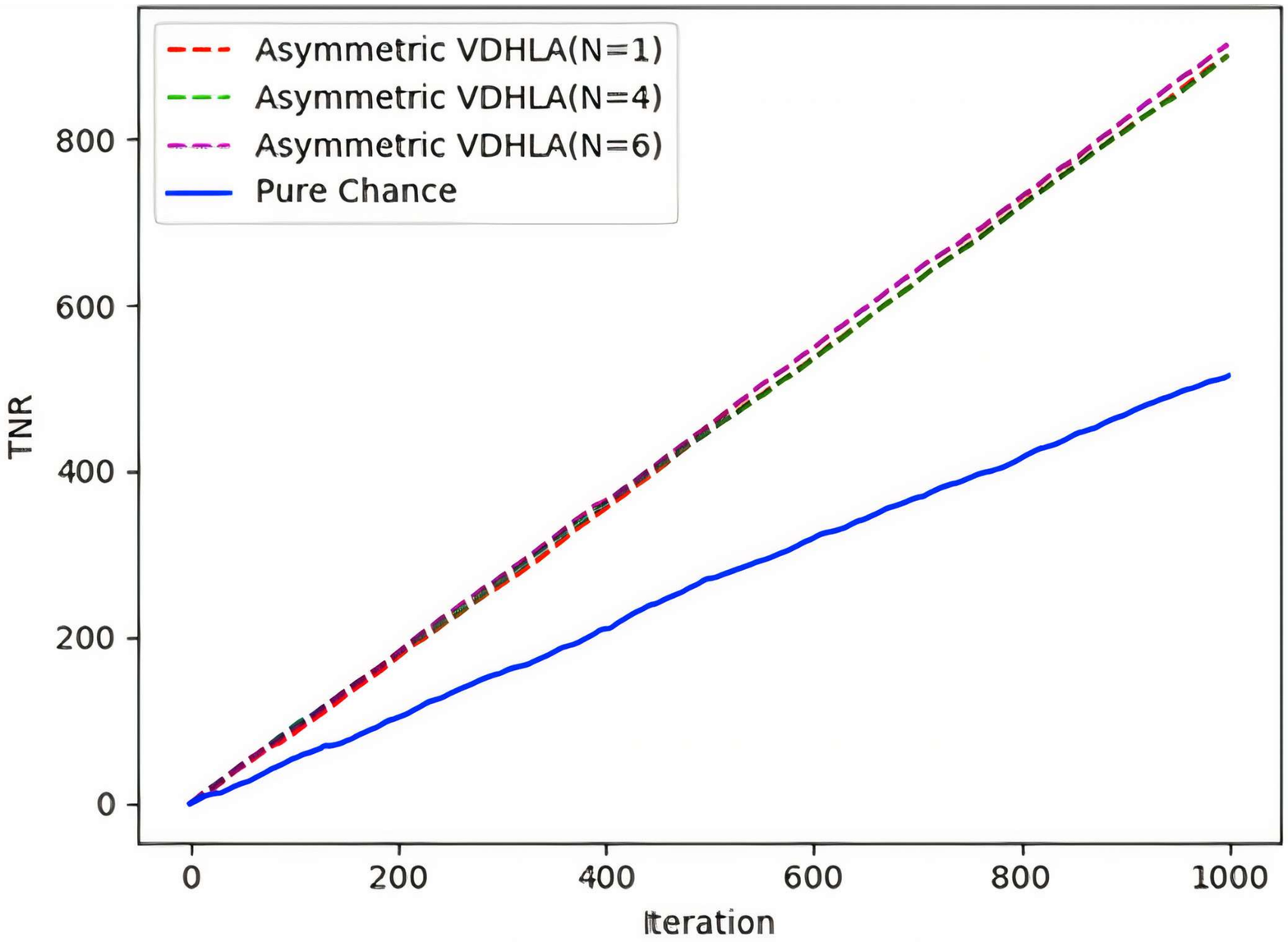}} 
\subfloat[TNAS of AVDHLA]{\includegraphics[width=0.25\textwidth]{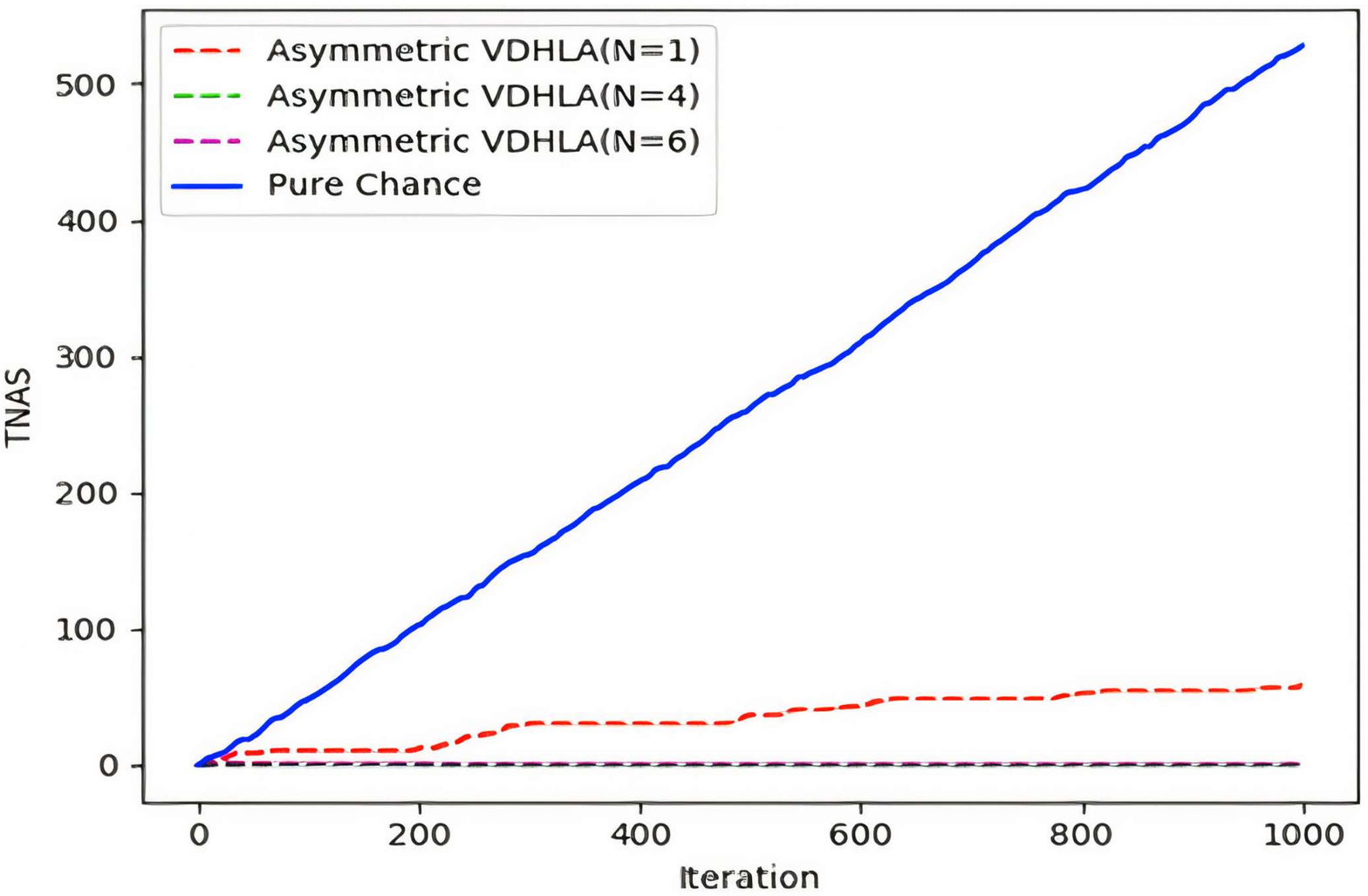}}\\ 
\caption{Experimental results of Ex 1.1 with reward probability vector of (0.1, 0.9)}
\label{fig_VDHLA_EX_1_1_1}
\end{figure}

The experimented SVDHLAs have 1, 4, 6 initial depth and simultaneously, AVDHLAs have $(1,1)$, $(4,4)$, $(6,6)$ initial depth. Fig \ref{fig_VDHLA_EX_1_1_1} shows the result from the TNR and TNAS aspects. 

As these figures show both of the proposed VDHLA kind learning automata (SVDHLA and AVDHLA) outperform the pure chance learning automaton with respect to TNR and TNAS. This leads us to this result that VDHLA is able to learn obviously.

According to the results of this experiment, VDHLA can detect the action with the higher probability of reward very well. Also, the VDHLA with the higher depth can adopt itself with the environment faster than the others with lower values. 

The main reason of this faster adoption is the ability of the automaton to increase its depth for the favorable action. Actually, the automaton with the higher values of depth has a better initial condition therefore needs less effort to change the depth. Eventually, they will get more rewards and decrease their action switching number toward the favorable action.  

Two other configurations are considered in Appendix \ref{section:appendix_exp_1_1} for evaluating the learning capability of VDHLA. The first one has closer probability (0.3 for action $\alpha_1$ and 0.7 for action $\alpha_2$). The next one has the same reward probability (0.5 for action $\alpha_1$ and 0.5 for action $\alpha_2$).

\paragraph{Experiment 1.2} 
This experiment is conducted to compare the proposed automaton with fixed-structure learning automaton. The considered environment has just one favorable action with the probability of 0.8. Depending on the number of other actions ($K$), 0.2 will divide between them. The following equation summarizes the configuration of the environment from the probability of being rewarded aspect. 

\begin{equation}
\label{equ_environment_ex_1_2_configuration}
   P(\alpha_i) = 
   \begin{cases}
      0.8 & i = 1 \\
      \frac{0.2}{K - 1} & i = 2, ..., K \\
   \end{cases}
\end{equation}

To make the experiment more productive, 9 actions are considered for VDHLA and FSLA. The considered initial depth for SVDHLA is 1, 3, 5, 7 for configuration number 1 to 4 respectively. Also, the initial depth for each action of AVDHLA is the same as SVDHLA through the considered configurations. 

The Number of iterations in SVDHLA experiment is 1000, and the conducted VASLA in SVDHLA is $L_{R\_I}$ with $\lambda_1 = 0.1$ and $\lambda_2 = 0$. The result of this experiment for SVDHLA can be seen in Table \ref{table_ex_1_2_SVDHLA_results}.

\begin{table}[!h]
\renewcommand{\arraystretch}{1.3}
\caption{Experimental Results of Experiment 1.2 for SVDHLA}
\label{table_ex_1_2_SVDHLA_results}
\centering
\resizebox{\columnwidth}{!}{
\begin{tabular}{lcccccccc}
\hline
Model & \multicolumn{2}{l}{\;\;\;\;Config 1} & \multicolumn{2}{l}{\;\;\;\;Config 2} & \multicolumn{2}{l}{\;\;\;\;Config  3} & \multicolumn{2}{l}{\;\;\;\;Config 4} \\
& TNR & TNAS & TNR & TNAS & TNR & TNAS & TNR & TNAS\\
\hline\hline
SVDHLA & \textbf{802} & \textbf{18} & \textbf{790} & \textbf{27} & \textbf{808} & \textbf{1} & \textbf{793} & \textbf{1} \\
FSLA & 290 & 710 & 730 & 85 & 801 & 3 & 778 & 2 \\
\hline
\end{tabular}}
\end{table}

The experiment is repeated for AVDHLA. This time the number of iterations increase to 10000 because AVDHLA needs more time to adopt itself with the environment. Also, inner VASLAs are changed to $L_{R\_{\epsilon P}}$ with  $\lambda_1 = 0.1$ and $\lambda_2 = 0.01$. The results is shown in Table \ref{table_ex_1_2_AVDHLA_results}.

\begin{table}[!h]
\renewcommand{\arraystretch}{1.3}
\caption{Experimental Results of Experiment 1.2 for AVDHLA}
\label{table_ex_1_2_AVDHLA_results}
\centering
\resizebox{\columnwidth}{!}{
\begin{tabular}{lcccccccc}
\hline
Model & \multicolumn{2}{l}{\;\;\;\;Config 1} & \multicolumn{2}{l}{\;\;\;\;Config 2} & \multicolumn{2}{l}{\;\;\;\;Config  3} & \multicolumn{2}{l}{\;\;\;\;Config 4} \\
& TNR & TNAS & TNR & TNAS & TNR & TNAS & TNR & TNAS\\
\hline\hline
AVDHLA & \textbf{7860} & \textbf{167} & \textbf{7675} & \textbf{446} & \textbf{8048} & \textbf{42} & \textbf{8074} & \textbf{4} \\
FSLA & 3181 & 6819 & 7487 & 729 & 7893 & 89 & 7990 & 24 \\
\hline
\end{tabular}}
\end{table}

The results expose that VDHLA family (SVDHLA and AVDHLA) overcomes FSLA in both TNR and TNAS aspects. It shows that VDHLA is less sensitive than FSLA with respect to the initial depth. if the depth of FSLA increases, it can get rewards better and the total number of action switching decreases. But since VDHLA can increase its depth autonomously, this problem is solved in this automaton. 

\begin{figure}[!t]
\centering
\subfloat[SVDHLA]{\includegraphics[width=0.25\textwidth]{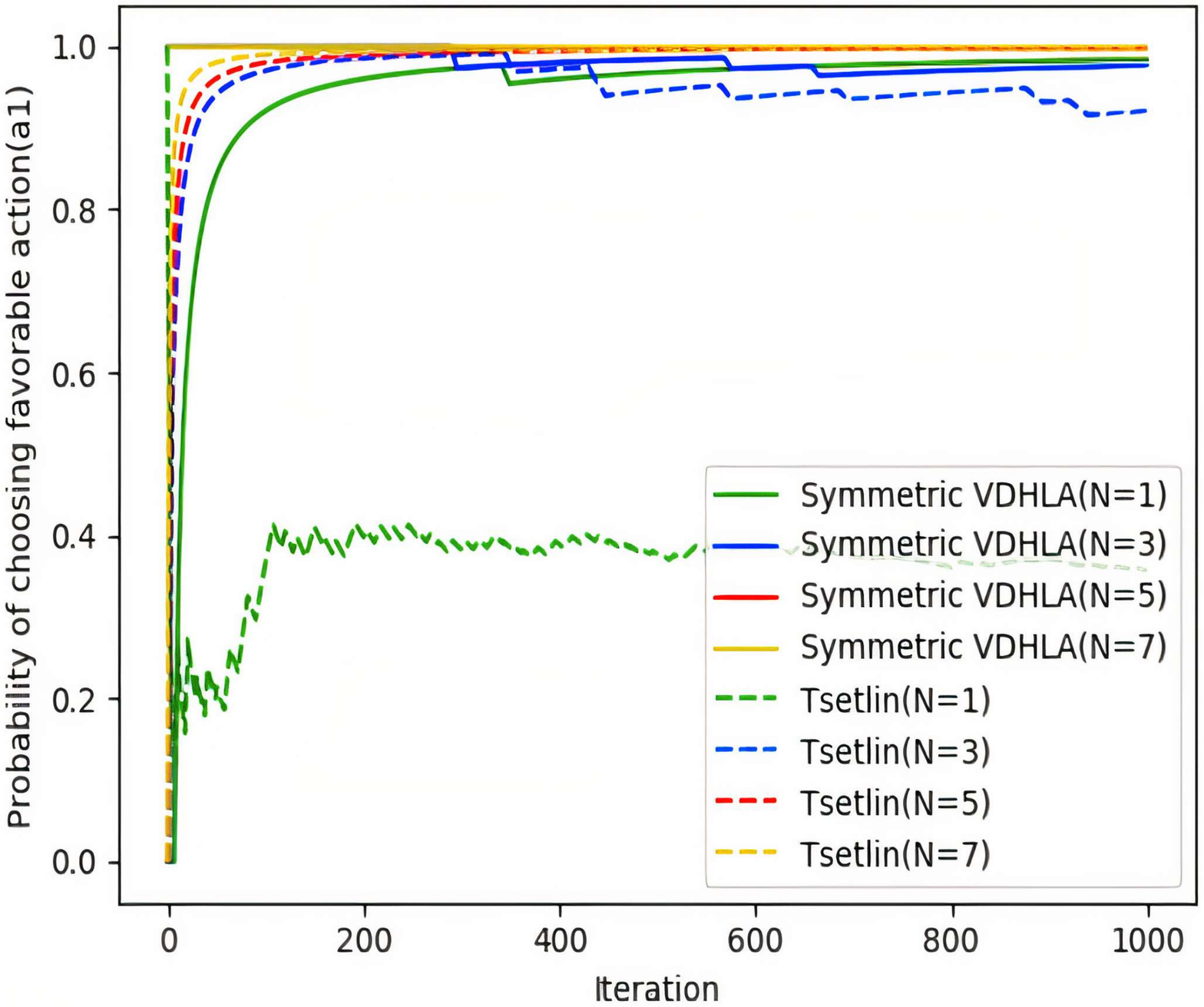}} 
\subfloat[AVDHLA]{\includegraphics[width=0.25\textwidth]{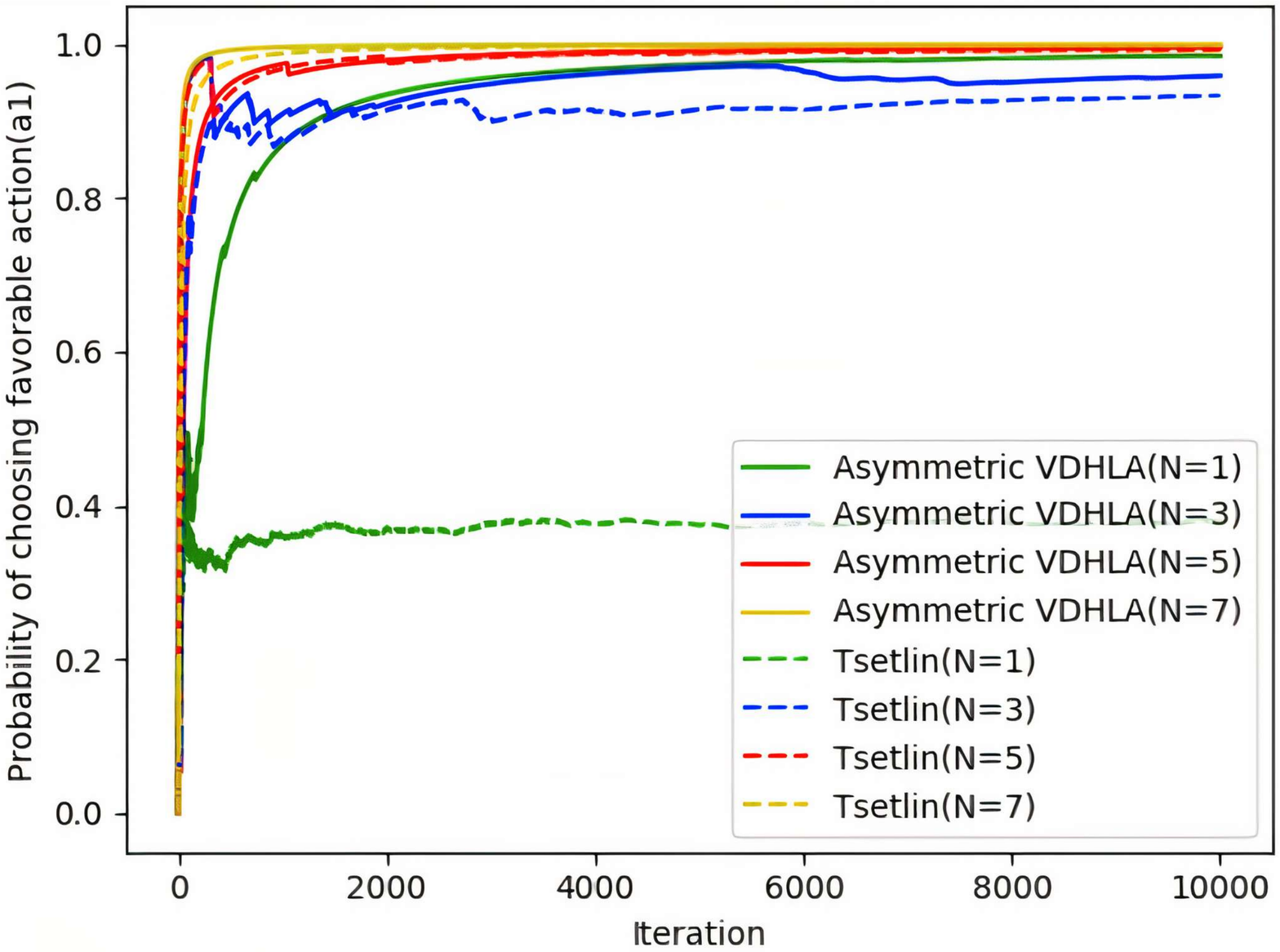}}\\ 
\caption{Experimental results of Ex 1.2 with respect to the probability of choosing the favorable action for VDHLA with K = 9}
\label{fig_VDHLA_EX_1_2}
\end{figure}

As well as TNR and TNAS metrics, we consider the probability of choosing the favorable action. In this experiment, favorable action is $\alpha_1$. The results can be seen in Fig.\ref{fig_VDHLA_EX_1_2}. By looking at this figure, we can see that VDHLA family impressively finds its appropriate depth in terms of the favorable action. It means that VDHLA is robust in terms of changing memory depth which is a great benefit of this model.

we should remark that more experiments have been performed to have a better comparison between VDHLA family and FSLA. These experiments exist in Appendix \ref{section:appendix_exp_1_2}.

\paragraph{Experiment 1.3} 
An experiment is needed to be designed to compare VDHLA family with VSLA. In order to do this, an environment with the probability of being rewarded for the specific action in 80 \% of times is considered. In such an environment the others have roughly 20\% chance to get reward. Equation \ref{equ_environment_ex_1_2_configuration} is applied to create this environment.

Besides, Number of allowed actions for an automaton to be in this environment are 4. 5 configurations are used for VSLAs is used to compare with VDHLA. Each configuration is tested with two values for $\lambda_1$ and $\lambda_2$ (denoted by a and b). These configurations are displayed in Table \ref{table_ex_1_3_configurations} from number 1 to 5.

\begin{table}[!h]
\renewcommand{\arraystretch}{1.3}
\caption{The Setup of Experiment 1.3}
\label{table_ex_1_3_configurations}
\centering
\resizebox{\columnwidth}{!}{
\begin{tabular}{lccccccccc}
\hline
  & \multicolumn{1}{l}{$P$} & \multicolumn{2}{l}{\hspace{3em} $L_{R\_I}$} & \multicolumn{2}{l}{\hspace{3em} $L_{P\_I}$} & \multicolumn{2}{l}{\hspace{3em} $L_{R\_P}$} & \multicolumn{2}{l}{\hspace{3em} $L_{R\_{\epsilon P}}$} \\
  & Config1 & Config2.a & Config2.b & Config3.a & Config3.b & Config4.a & Config4.b & Config5.a & Config5.b \\
\hline\hline
 K & 5 & 5 & 5 & 5 & 5 & 5 & 5 & 5 & 5 \\
 N & 5 & 5 & 5 & 5 & 5 & 5 & 5 & 5 & 5 \\
 $\lambda_1$ & 0 & 0.1 & 0.01 & 0 & 0 & 0.1 & 0.01 & 0.1 & 0.01 \\
 $\lambda_2$ & 0 & 0 & 0 & 0.1 & 0.01 & 0.1 & 0.01 & 0.01 & 0.001 \\
\hline
\end{tabular}}
\end{table}

The results of this experiment with respect to TNR and TNAS are shown in Table \ref{table_ex_1_3_result} considering both of SVDHLA and AVDHLA together for 1000 iterations. These outstanding result shows the power of the proposed automaton to collect more rewards with lower number of action switching over VSLA in all cases. It means that VDHLA unlike the VSLA doesn't have any dependencies to updating schemes. Also the lower number of action switching narrates that the proposed automaton learns to find the favorable action.

\begin{table}[!h]
\renewcommand{\arraystretch}{1.3}
\caption{Experimental results of Ex 1.3 with respect to TNR and TNAS}
\label{table_ex_1_3_result}
\centering
\resizebox{\columnwidth}{!}{
\begin{tabular}{lcccccccccccccccccc}
\hline
  Model & \multicolumn{2}{l}{\hspace{1em} Config1} & \multicolumn{2}{l}{\hspace{1em} Config2.a} & \multicolumn{2}{l}{\hspace{1em} Config2.b} & \multicolumn{2}{l}{\hspace{1em} Config3.a} & \multicolumn{2}{l}{\hspace{1em} Config3.b} & \multicolumn{2}{l}{\hspace{1em} Config4.a} & \multicolumn{2}{l}{\hspace{1em} Config4.b} & \multicolumn{2}{l}{\hspace{1em} Config5.a} &  \multicolumn{2}{l}{\hspace{1em} Config5.b}\\
 & TNR & TNAS & TNR & TNAS & TNR & TNAS & TNR & TNAS & TNR & TNAS & TNR & TNAS & TNR & TNAS & TNR & TNAS & TNR & TNAS \\
\hline\hline
 SVDHLA & \textbf{811} & \textbf{20} & \textbf{805} & \textbf{5} & \textbf{784} & \textbf{35} & \textbf{805} & \textbf{5} & \textbf{770} & \textbf{20} & \textbf{806} & \textbf{5} & \textbf{775} & \textbf{5} & \textbf{784} & \textbf{15} & \textbf{796} & \textbf{5}  \\
 AVDHLA & 742 & 57 & 795 & 18 & 791 & 9 & 813 & 0 & 789 & 9 & 790 & 8 & 803 & 3 & 808 & 7 & 800 & 13  \\
 VSLA & \textbf{203} & \textbf{812} & \textbf{794} & \textbf{38} & \textbf{629} & \textbf{345} & \textbf{230} & \textbf{816} & \textbf{214} & \textbf{792} & \textbf{462} & \textbf{638} & \textbf{355} & \textbf{750} & \textbf{721} & \textbf{109} & \textbf{545} & \textbf{423}  \\
\hline
\end{tabular}}
\end{table}

\begin{figure*}[!t]
\centering
\subfloat[Config 1]{\includegraphics[width=0.33\textwidth]{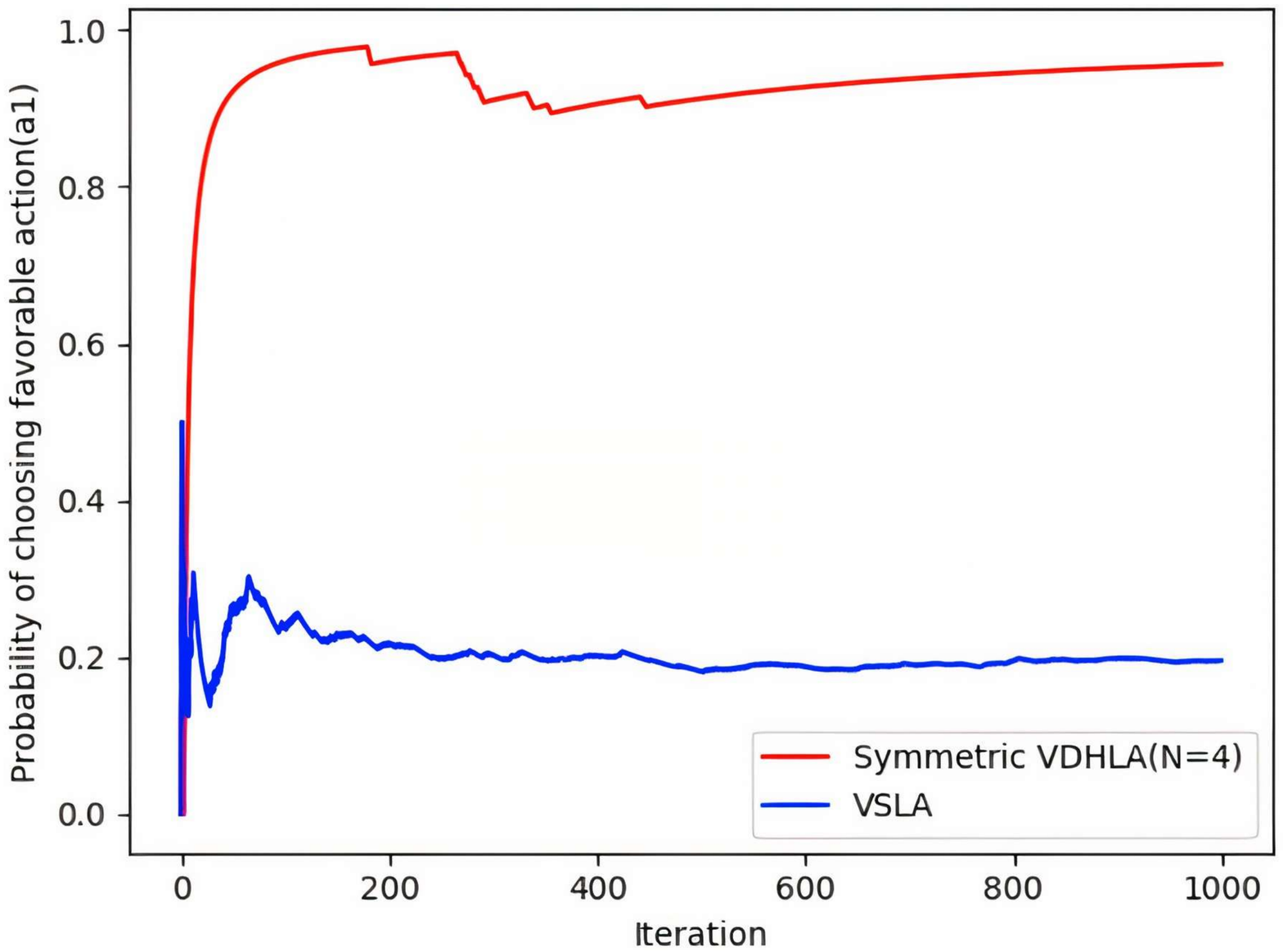}} 
\subfloat[Config 2]{\includegraphics[width=0.33\textwidth]{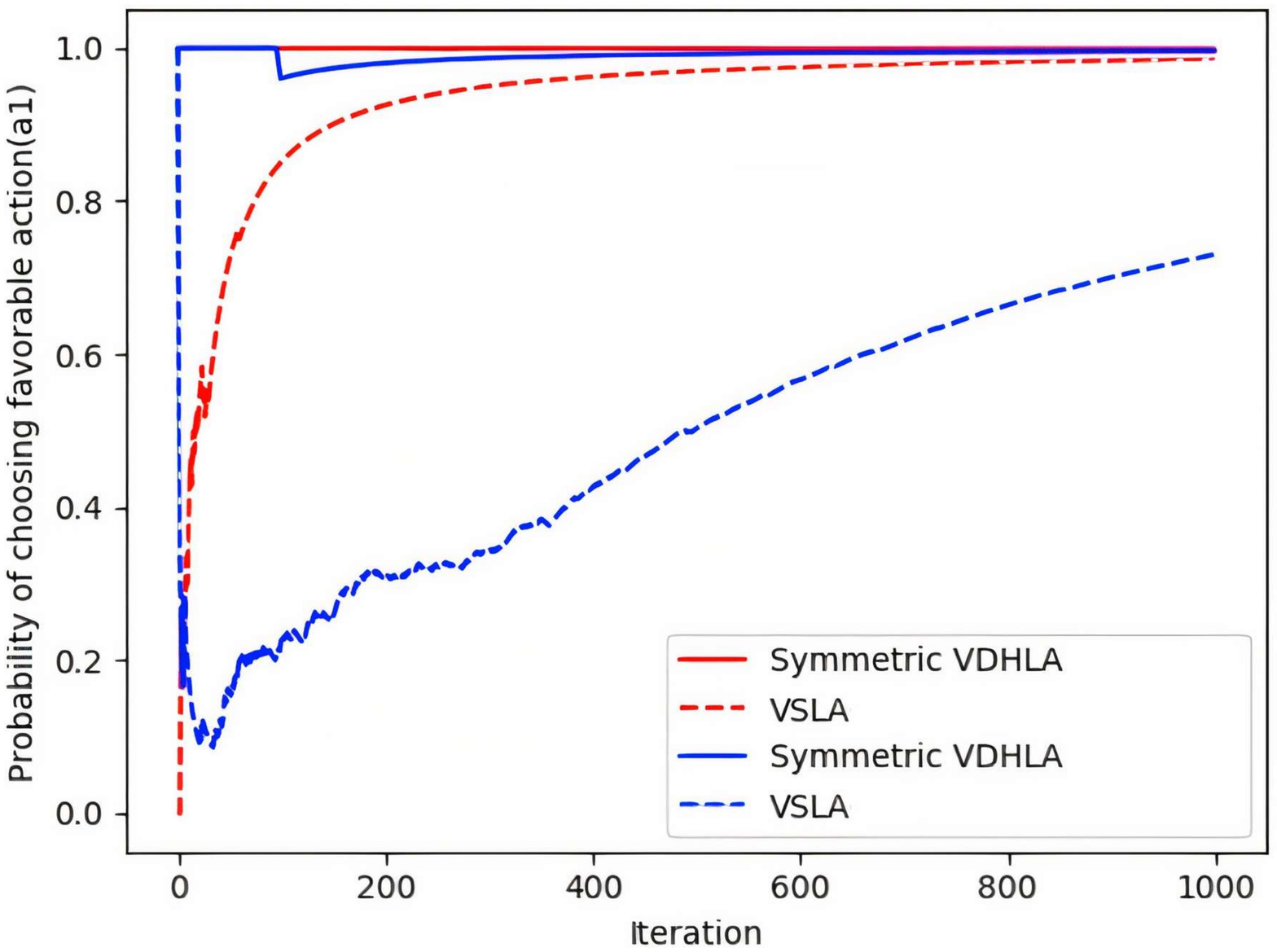}}
\subfloat[Config 3]{\includegraphics[width=0.33\textwidth]{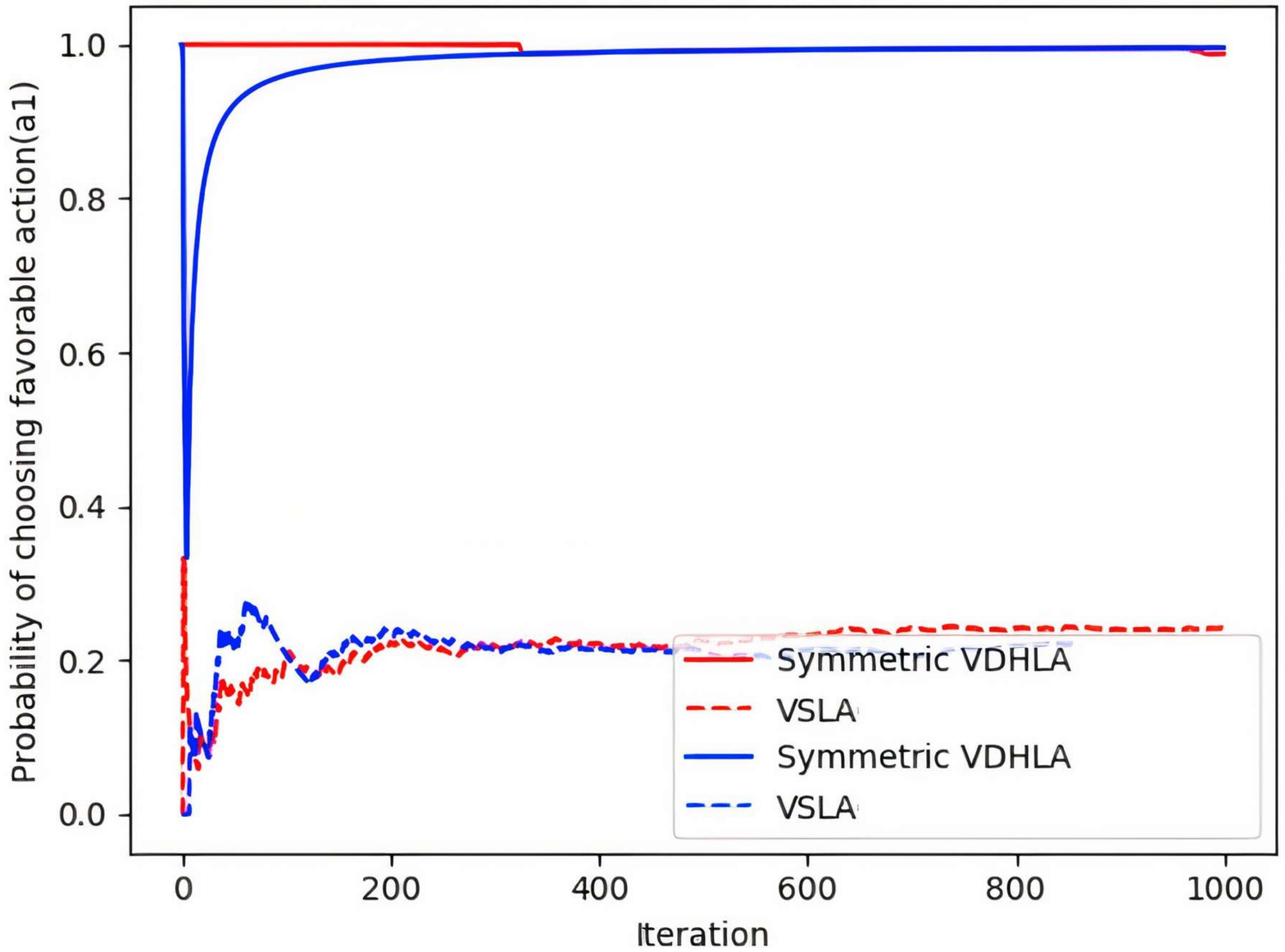}}\\ 
\subfloat[Config 4]{\includegraphics[width=0.33\textwidth]{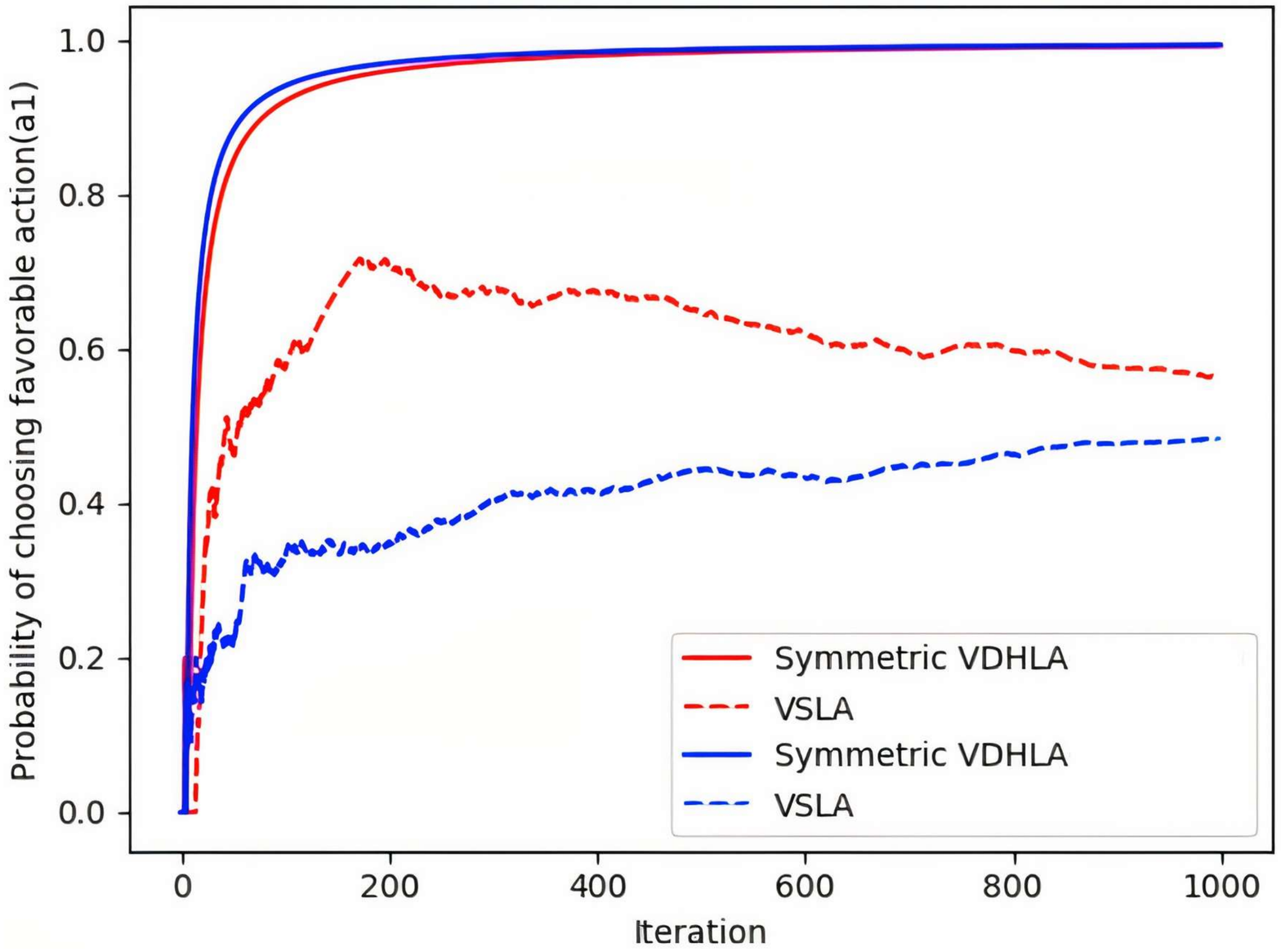}} 
\subfloat[Config 5]{\includegraphics[width=0.33\textwidth]{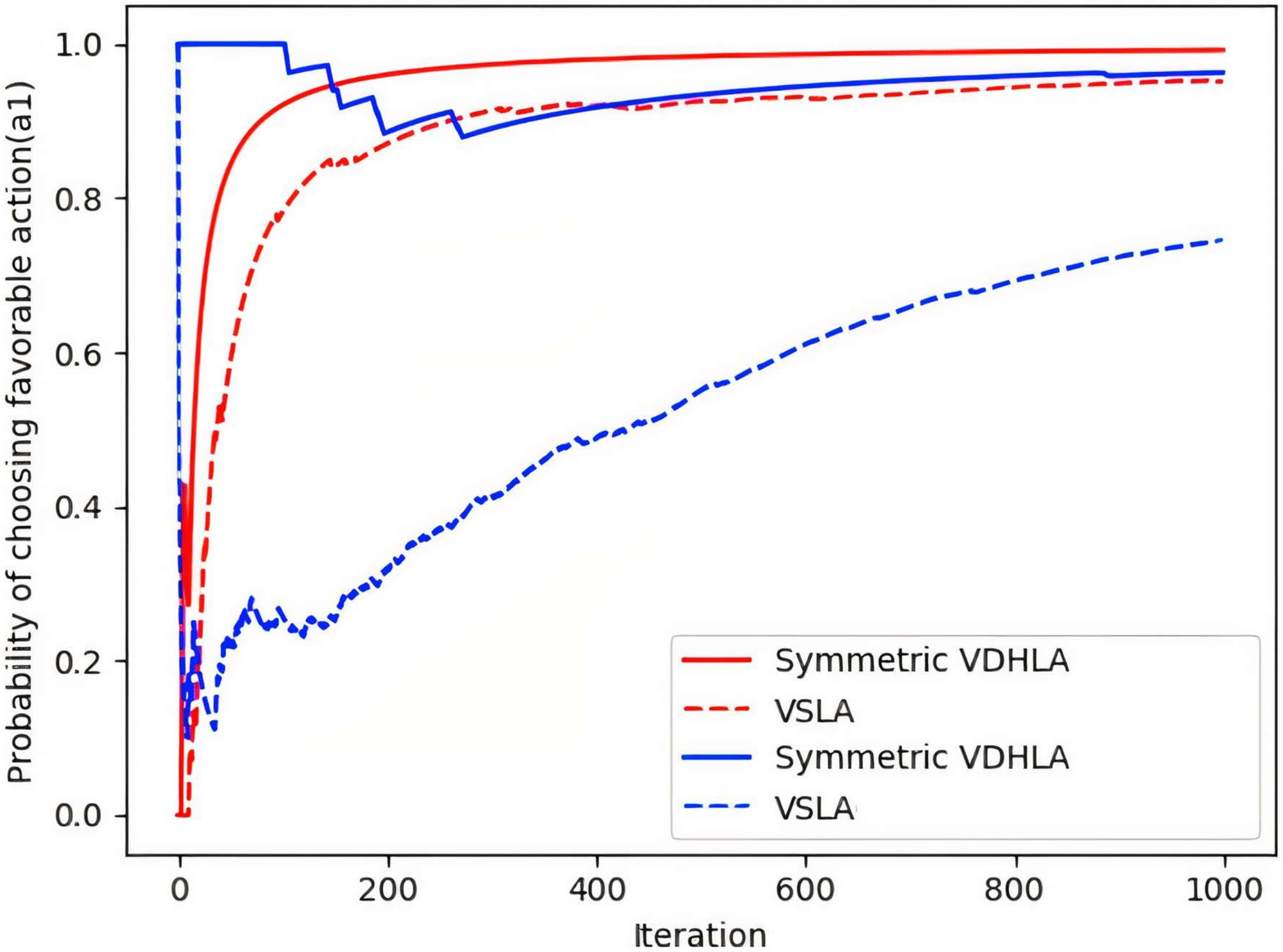}}\\ 
\caption{Experimental results of Ex 1.3 with respect to the probability of choosing the favorable action for SVDHLA; Config a denoted by red and Config b denoted by blue}
\label{fig_SVDHLA_EX_1_3}
\end{figure*}

\begin{figure*}[!t]
\centering
\subfloat[Config 1]{\includegraphics[width=0.33\textwidth]{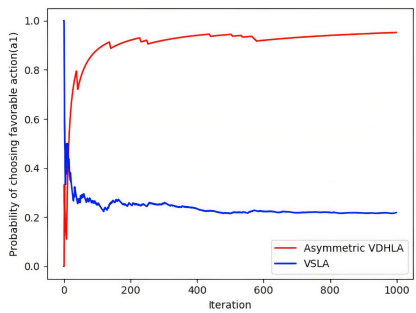}} 
\subfloat[Config 2]{\includegraphics[width=0.33\textwidth]{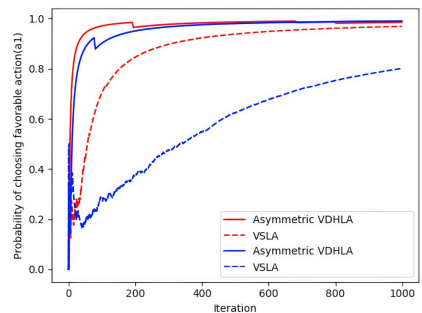}}
\subfloat[Config 3]{\includegraphics[width=0.33\textwidth]{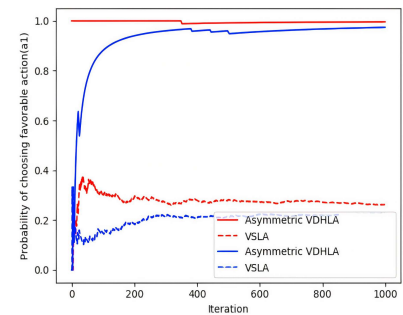}}\\ 
\subfloat[Config 4]{\includegraphics[width=0.33\textwidth]{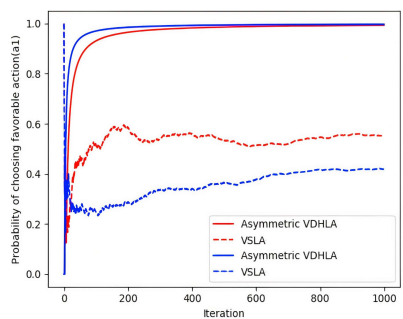}} 
\subfloat[Config 5]{\includegraphics[width=0.33\textwidth]{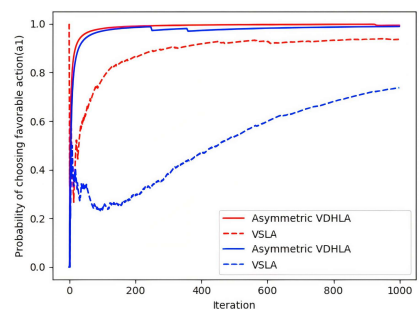}}\\ 
\caption{Experimental results of Ex 1.3 with respect to the probability of choosing the favorable action for AVDHLA; Config a denoted by red and Config b denoted by blue}
\label{fig_AVDHLA_EX_1_3}
\end{figure*}

Moreover, in the probability of choosing the favorable action metric, VSLA inferior to VDHLA. This comes from our measuring of this metric in Fig.\ref{fig_SVDHLA_EX_1_3} for SVDHLA and Fig.\ref{fig_AVDHLA_EX_1_3} for AVDHLA. The main reason is that VDHLA inclines to the action $\alpha_1$ swiftly, therefore less likely to get a penalty from the environment. Also, this part of experiment leads us to this result that probability of choosing the favorable action in VDHLA by contradiction to VSLA is regardless to updating scheme. 

\paragraph{Experiment 1.4}
This experiment is conducted to compare the introduced kinds of VDHLA family in various configurations. At first, we develop an experiment with 4 configurations for different values of initial depth. 1, 3, 5. 7 are considered for the initial depth of SVDHLA and each action's depth of AVDHLA through configuration 1 till 4. It also consists of 3 values $K = 2, K = 5, K = 9$ for the number of allowed action. This experiment is done in 2 phases:

\begin{enumerate}

\item Phase 1: The inner VASLAs are $L_{R\_{\epsilon P}}$ with $\lambda_1 = 0.1$ and $\lambda = 0.01$. This phase is explained in the following.

\item Phase 2: The inner VASLAs are $L_{R\_I}$ with $\lambda_1 = 0.1$ and $\lambda = 0$. This phase is explained in Append \ref{section:appendix_exp_1_4}.

\end{enumerate}

The results of this experiment with considered values of $K$ are in Table \ref{table_ex_1_4_results} for 10000 iterations. By looking at results, we can conclude that none of them are superior to the other with respect to the TNR and TNAS. There is a little difference in TNAS metric. AVDHLA has lower number of TNAS, because it can choose the depth of each action asymmetrically. Therefore, on choosing the action which is not favorable, it doesn't change the depth of all action together, just decreases the depth of unfavorable action. Beside that, it increases the depth of favorable action, so as expected TNR increases and TNAS decreases.

Furthermore, by increasing the number of $K$, the automaton can find the favorable action harder. This is obvious in the results of the experiment with $K = 9$. The other parameter that can be discussed is $N$. Increasing the initial depth will improve the performance of the automaton with respect to the TNR and TNAS. The main reason is that with higher N, VDHLA doesn't change its decision roughly.

\begin{table}[!h]
\renewcommand{\arraystretch}{1.3}
\caption{Experimental Results of Experiment 1.4 with Respect to TNR and TNAS}
\label{table_ex_1_4_results}
\centering
\resizebox{\columnwidth}{!}{
\begin{tabular}{lcccccccc}
\hline
Model & \multicolumn{2}{l}{\;\;\;\;Config 1} & \multicolumn{2}{l}{\;\;\;\;Config 2} & \multicolumn{2}{l}{\;\;\;\;Config  3} & \multicolumn{2}{l}{\;\;\;\;Config 4} \\
& TNR & TNAS & TNR & TNAS & TNR & TNAS & TNR & TNAS\\
\hline
&&&&K = 2&&&&\\
\hline
AVDHLA & \textbf{7923} & \textbf{49} & \textbf{7928} & \textbf{166} & \textbf{7990} & \textbf{6} & \textbf{8001} & \textbf{2} \\
SVDHLA & 7871 & 238 & 7929 & 110 & 8014 & 20 & 8000 & 5 \\
\hline
&&&&K = 5&&&&\\
\hline
AVDHLA & \textbf{7951} & \textbf{94} & \textbf{7861} & \textbf{120} & \textbf{7930} & \textbf{27} & \textbf{8067} & \textbf{9} \\
SVDHLA & 7928 & 100 & 8010 & 10 & 7920 & 67 & 8020 & 12 \\
\hline
&&&&K = 9&&&&\\
\hline
AVDHLA & \textbf{7878} & \textbf{208} & \textbf{7936} & \textbf{87} & \textbf{8036} & \textbf{15} & \textbf{7948} & \textbf{6} \\
SVDHLA & 7992 & 18 & 7949 & 36 & 8014 & 9 & 7996 & 18 \\
\hline
\end{tabular}}
\end{table}

\paragraph{Experiment 1.5}
This experiment devotes to compare the AVDHLA with SVDHLA in terms of asymmetric chosen depth for AVDHLA. To do this experiment, we assume that both of automata have 5 actions and the inner VASLAs are $L_{R\_{\epsilon P}}$ with $\lambda_1 = 0.1$ and $\lambda = 0.01$. This experiment has three potential scenarios as follows:

\begin{enumerate}

\item The initial depth of AVDHLA from the favorable action is high. As well, other actions in such an automaton have a lower initial depth. Since the first action in the desire environment is $\alpha_1$, therefore AVDHLA's initial depth vector is $[10, 3, 3, 3, 3]$.

\item The initial depth of AVDHLA for all actions have the same value. So, $[3, 3, 3, 3, 3]$ vector is considered.

\item The initial depth of AVDHLA from the favorable action is low. Beside, other actions have a higher initial depth. Therefore, vector $[3, 10, 10, 10, 10]$ is used as a initial depth.

\end{enumerate}

The outstanding results of this experiments which can be seen in Table \ref{table_ex_1_5_results} shows how the AVDHLA is affected by the initial depth in compare to SVDHLA. If the initial depth of AVDHLA sets accurately, this automaton will be so powerful in the stationary environment with respect to TNR and TNAS. First, we infer from this experiment that how strong AVDHLA would be if the initial depth is chosen correctly. Second, the wrong chosen depth for the initial depth can be compensated because the proposed automaton can adopt its depth by interacting with the outer environment. The penalty of this wrong choice leads to increasing the number of action switching. Consequently, AVDHLA is worthy of trust even with wrong chosen initial depth. 

\begin{table}[!h]
\renewcommand{\arraystretch}{1.3}
\caption{Experimental Results of Experiment 1.5 with Respect to TNR and TNAS}
\label{table_ex_1_5_results}
\centering
\resizebox{\columnwidth}{!}{
\begin{tabular}{lcccccccc}
\hline
Model & \multicolumn{2}{l}{\;\;\;\;Depth = 1} & \multicolumn{2}{l}{\;\;\;\;Depth = 3} & \multicolumn{2}{l}{\;\;\;\;Depth = 5} & \multicolumn{2}{l}{\;\;\;\;Depth = 7} \\
& TNR & TNAS & TNR & TNAS & TNR & TNAS & TNR & TNAS\\
\hline
SVDHLA & \textbf{7881} & \textbf{75} & \textbf{7966} & \textbf{60} & \textbf{7932} & \textbf{5} & \textbf{8006} & \textbf{10} \\
&\multicolumn{7}{l}{\textbf{AVDHLA with [10, 3, 3, 3, 3] Initial Depth - Scenario 1}}&\\
\hline
\hline
&&{TNR}&&&&{TNAS}&&\\
AVDHLA &&{8037}&&&&{1}&& \\
\hline
&\multicolumn{7}{l}{\textbf{AVDHLA with [3, 3, 3, 3, 3] Initial Depth - Scenario 2}}&\\
\hline
\hline
&&{TNR}&&&&{TNAS}&&\\
AVDHLA &&{7928}&&&&{74}&& \\
\hline
&\multicolumn{7}{l}{\textbf{AVDHLA with [3, 10, 10, 10, 10] Initial Depth - Scenario 3}}&\\
\hline
\hline
&&{TNR}&&&&{TNAS}&&\\
AVDHLA &&{7848}&&&&{110}&& \\
\hline
\end{tabular}}
\end{table}

\subsubsection{Experiment 2}
\begin{figure}[!t]
    \centering
    \includegraphics[width=0.5\textwidth]{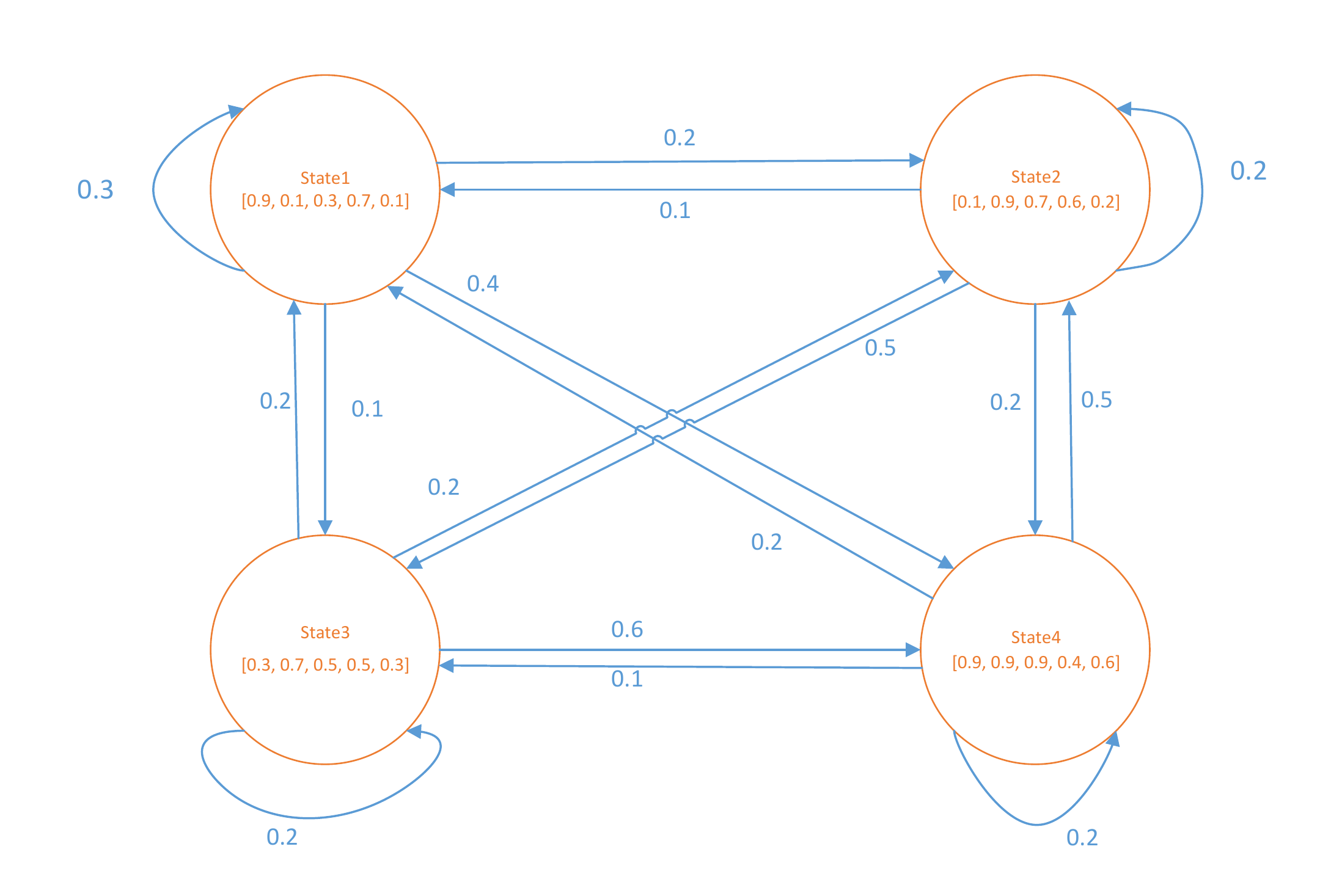}
    \caption{The designed environment for Markovian Switching experiment}
    \label{fig_markovian_switching}
\end{figure}

This experiment is conducted to study the performance of the proposed automaton in the complex Markovian Switching environment with respect to TNR and TNAS. To reach this objective, a markov chain with 4 states is considered. To have a better overview, Fig.\ref{fig_markovian_switching} shows the desired environment. In such an environment, we considered the transition matrix ($T$), and the probability of being rewarded in each state denoted by reward matrix ($R$)  as follows respectively.

\begin{equation}
\label{equ_exp_2_transition}
T = \left(\begin{IEEEeqnarraybox}[][c]{,c/c/c/c/c/c/c,}
0.3&&0.2&&0.1&&0.4\\
0.1&&0.2&&0.5&&0.2\\
0.2&&0.2&&0.2&&0.6\\
0.2&&0.5&&0.1&&0.2
\end{IEEEeqnarraybox}\right)
\end{equation}

\begin{equation}
\label{equ_exp_2_reward}
R = \left(\begin{IEEEeqnarraybox}[][c]{,c/c/c/c/c/c/c/c/c,}
0.9&&0.1&&0.3&&0.7&&0.1\\
0.1&&0.9&&0.7&&0.6&&0.2\\
0.3&&0.7&&0.5&&0.5&&0.3\\
0.9&&0.9&&0.9&&0.4&&0.6
\end{IEEEeqnarraybox}\right)
\end{equation}

In addition, inner VASLAs are $L_{P\_I}$ with $\lambda_1 = 0$ and $\lambda_2 = 0.01$. The number of allowed action is 5, and the initial depth is 1, 3, 5, 7 for configuration 1 to 4 respectively.

Before analyzing the results which are shown in Table \ref{table_ex_markovian_switching_results_3_appendix} for 1000 iterations, let's turn the Markovian Switching environment into the stationary environment using the steady state analysis of the markov chain.

\begin{IEEEeqnarray}{c}
\label{equ_markovian_3_conversion_appendix}
v_{\infty}T = v_{\infty} \IEEEyesnumber\IEEEyessubnumber*\\\\
v_{\infty}
\left(\begin{IEEEeqnarraybox}[][c]{,c/c/c/c/c/c/c,}
0.3&&0.2&&0.1&&0.4\\
0.1&&0.2&&0.5&&0.2\\
0.2&&0.2&&0.2&&0.6\\
0.2&&0.5&&0.1&&0.2
\end{IEEEeqnarraybox}\right)
= v_{\infty}\\\\
\left(\begin{IEEEeqnarraybox}[][c]{,c/c/c/c/c/c/c,}
x&&y&&w&&z
\end{IEEEeqnarraybox}\right)
\left(\begin{IEEEeqnarraybox}[][c]{,c/c/c/c/c/c/c,}
0.3&&0.2&&0.1&&0.4\\
0.1&&0.2&&0.5&&0.2\\
0.2&&0.2&&0.2&&0.6\\
0.2&&0.5&&0.1&&0.2
\end{IEEEeqnarraybox}\right)
=\left(\begin{IEEEeqnarraybox}[][c]{,c/c/c/c/c/c/c,}
x&&y&&w&&z
\end{IEEEeqnarraybox}\right)\\\\
\begin{cases}
	0.3 x + 0.1 y + 0.2 w + 0.2 z = x \\
    0.2 x + 0.2 y + 0.2 w + 0.5 z = y \\
    0.1 x + 0.5 y + 0.2 w + 0.1 z = w \\
    0.4 x + 0.2 y + 0.6 w + 0.2 z = z \\
    x + y + w + z = 1
\end{cases} \\\\
\begin{cases}
	x = 0.19\\
    y = 0.28\\
    w = 0.23\\
    z = 0.28
\end{cases}
\end{IEEEeqnarray}

Accordingly, the above conversion turns Markovian Switching environment into the stationary environment with the probability of being rewarded 0.19, 0.28, 0.23 and, 0.28 for action 1 till action 4 respectively.

\begin{table}[!h]
\renewcommand{\arraystretch}{1.3}
\caption{Experimental Results of Experiment 2 with Respect to TNR and TNAS}
\label{table_ex_markovian_switching_results_3_appendix}
\centering
\resizebox{\columnwidth}{!}{
\begin{tabular}{lcccccccc}
\hline
Model & \multicolumn{2}{l}{\;\;\;\;Config 1} & \multicolumn{2}{l}{\;\;\;\;Config 2} & \multicolumn{2}{l}{\;\;\;\;Config  3} & \multicolumn{2}{l}{\;\;\;\;Config 4} \\
& TNR & TNAS & TNR & TNAS & TNR & TNAS & TNR & TNAS\\
\hline
\hline
AVDHLA & \textbf{6948} & \textbf{73} & \textbf{6407} & \textbf{264} & \textbf{6649} & \textbf{151} & \textbf{6897} & \textbf{52} \\
SVDHLA & 6373 & 316 & 6815 & 76 & 6622 & 41 & 6403 & 77 \\
FSLA & \textbf{5863} & \textbf{4137} & \textbf{6113} & \textbf{935} & \textbf{6565} & \textbf{242} & \textbf{6680} & \textbf{58} \\
\hline
\end{tabular}}
\end{table}

According to the random selection of values such a the transition matrix and the reward matrix, the proposed automaton can adopt itself with the Markovian Switching environment with respect to TNR and TNAS. In addition, AVDHLA outperforms SVDHLA and FSLA in getting reward and minimizing the action switching. The main reason to reach this goal is the ability of AVDHLA in choosing the right depth through the actions with higher probability of reward. We should remark this point that all the actions approximately have the same probability of getting reward, hence the initial depth doesn't play an important role in this experiment. Finally, we should conclude that both AVDHLA and SVDHLA can worth of trust in Markovian Switching environment.

As well as this experiment, accurate analysis of VDHLA in Markovian Switching environments are performed in Appendix \ref{section:appendix_markovian_switching_exp}. In this Appendix, various conditions are considered to compare the proposed automaton with FSLA in conjunction with TNR and TNAS.  

\subsubsection{Experiment 3}
This experiment is devoted to study the performance of the proposed automaton in State-Dependent environment with respect to TNR and TNAS. To aim this, this experiment is divided into two parts:

\begin{enumerate}

\item Experiment 3.1 for comparing VDHLA to FSLA in order to see the difference between these automata in getting penalty and reward.

\item Experiment 3.2 for comparing VDHLA with VSLA according to find the strengths and weaknesses. 

\end{enumerate}

For above experiments, three tuples of ($\theta$, $\phi$) are considered: $(0.0002, 0.00002)$, $(0.0005, 0.0005)$, and $(0.00002, 0.0002)$ for scenarios 1 till 3 correspondingly.

\paragraph{Experiment 3.1}
The proposed model should be compared to FSLA in State-Dependent environment. The motivation is to analyze the effect of $\theta$ and $\phi$ parameters which are introduced in section \ref{section:non_stationary_environment}.

In order to do this experiment, inner VASLA are $L_{R\_I}$ with $\lambda_1 = 0.01$ and $\lambda_2 = 0$. The considered initial depth are 1, 3, 5, 7 for configuration 1 to 4 respectively. 

The initial reward probability vector of the desired environment is $[0.9, 0.1]$. This means that at the first stage, action 1 has the probability of 0.9 and action 2 has the probability of 0.1 for being rewarded. In the next stages this reward probability vector increases or decreases through the values of $\theta$ and $\phi$.

\begin{table}[!h]
\renewcommand{\arraystretch}{1.3}
\caption{Experimental Results of the Experiment 3.1 with Respect to TNR and TNAS}
\label{table_ex_3_1_results}
\centering
\resizebox{\columnwidth}{!}{
\begin{tabular}{lcccccccc}
\hline
Model & \multicolumn{2}{l}{\;\;\;\;Config 1} & \multicolumn{2}{l}{\;\;\;\;Config 2} & \multicolumn{2}{l}{\;\;\;\;Config  3} & \multicolumn{2}{l}{\;\;\;\;Config 4} \\
& TNR & TNAS & TNR & TNAS & TNR & TNAS & TNR & TNAS\\
\hline
&& \multicolumn{5}{l}{Scenario 1 - $(\theta = 0.0002, \phi = 0.00002)$} && \\
\hline
AVDHLA & \textbf{7551} & \textbf{33} & \textbf{7421} & \textbf{216} & \textbf{7422} & \textbf{90} & \textbf{7469} & \textbf{73} \\
SVDHLA & 7452 & 132 & 7465 & 320 & 7420 & 80 & 7496 & 70 \\
FSLA & \textbf{6319} & \textbf{3681} & \textbf{7346} & \textbf{361} & \textbf{7433} & \textbf{60} & \textbf{7424} & \textbf{7} \\
\hline
&& \multicolumn{5}{l}{Scenario 2 - $(\theta = 0.0005, \phi = 0.0005)$} && \\
\hline
AVDHLA & \textbf{4566} & \textbf{2384} & \textbf{4690} & \textbf{2316} & \textbf{4650} & \textbf{2043} & \textbf{4696} & \textbf{2107} \\
SVDHLA & 4655 & 1876 & 4374 & 2790 & 4539 & 2866 & 4510 & 2956 \\
FSLA & \textbf{3604} & \textbf{6396} & \textbf{4677} & \textbf{2888} & \textbf{4465} & \textbf{2397} & \textbf{4609} & \textbf{2117} \\
\hline
&& \multicolumn{5}{l}{Scenario 3 - $(\theta = 0.00002, \phi = 0.0002)$} && \\
\hline
AVDHLA & \textbf{4224} & \textbf{3630} & \textbf{4162} & \textbf{4071} & \textbf{4168} & \textbf{3413} & \textbf{4385} & \textbf{2553} \\
SVDHLA & 4103 & 3689 & 4103 & 2934 & 4032 & 2538 & 4233 & 3447 \\
FSLA & \textbf{2865} & \textbf{7135} & \textbf{4149} & \textbf{4149} & \textbf{4110} & \textbf{3585} & \textbf{4066} & \textbf{2900} \\
\hline
\end{tabular}}
\end{table}

In this experiment, action 1 is favorable at the first stage. After 1000 iterations, this action is also favorable although $\theta$ and $\phi$ affect the reward probability vector. The results in Table \ref{table_ex_3_1_results} reveals the superiority of VDHLA over FSLA with respect to TNR and TNAS. Also, in VDHLA family, AVDHLA performs better than SVDHLA in conjunction with measured parameters. Because AVDHLA can track the environment by decreasing the depth of action 1 when it is needed.

\paragraph{Experiment 3.2}
This experiment is in charge of evaluating the proposed automaton with respect to TNR and TNAS. To reach this goal, $K = 5$ and $N = 4$ is considered for VDHLA. On the other side for VSLA, the setup of this experiment is exactly the same as equation \ref{equ_environment_Probability}. The inner VASLAs are the same as VSLA ( except the number of actions) with the configurations that are shown in Table \ref{table_ex_3_2_configurations}.

\begin{table}[!h]
\renewcommand{\arraystretch}{1.3}
\caption{The Setup of VSLA in Experiment 3.2}
\label{table_ex_3_2_configurations}
\centering
\resizebox{\columnwidth}{!}{
\begin{tabular}{lccccccccc}
\hline
  & \multicolumn{1}{l}{$P$} & \multicolumn{2}{l}{\hspace{3em} $L_{R\_I}$} & \multicolumn{2}{l}{\hspace{3em} $L_{P\_I}$} & \multicolumn{2}{l}{\hspace{3em} $L_{R\_P}$} & \multicolumn{2}{l}{\hspace{3em} $L_{R\_{\epsilon P}}$} \\
  & Config1 & Config2.a & Config2.b & Config3.a & Config3.b & Config4.a & Config4.b & Config5.a & Config5.b \\
\hline\hline
 K & 5 & 5 & 5 & 5 & 5 & 5 & 5 & 5 & 5 \\
 N & 4 & 4 & 4 & 4& 4 & 4 & 4 & 4 & 4 \\
 $\lambda_1$ & 0 & 0.1 & 0.01 & 0 & 0 & 0.1 & 0.01 & 0.1 & 0.01 \\
 $\lambda_2$ & 0 & 0 & 0 & 0.1 & 0.01 & 0.1 & 0.01 & 0.01 & 0.001 \\
\hline
\end{tabular}}
\end{table}

According to the result of this experiment which is shown in table \ref{table_ex_1_3_result}, in all configuration VDHLA outperforms VSLA in conjunction with TNR and TNAS. When both of $\lambda_1$ and $\lambda_2$ equal to 0, since VDHLA turns into FSLA, it doesn't change the decision easily. As a result, it will get more rewards in addition to the less number of action switching. Also, between various configurations, $L_{R\_I}$ is superior. The main reason is that this scheme is not sensitive to penalty. Consequently, getting penalty from such an environment in which the reward probability changes doesn't affect the probability vector. Eventually, VDHLA is superior to VSLA in State-Dependent environment with respect to the define experiment metrics.

\begin{table}[!h]
\renewcommand{\arraystretch}{1.3}
\caption{Experimental results of Experiment 3.2 with respect to TNR and TNAS}
\label{table_ex_1_3_result}
\centering
\resizebox{\columnwidth}{!}{
\begin{tabular}{lcccccccccccccccccc}
\hline
  Model & \multicolumn{2}{l}{\hspace{1em} Config1} & \multicolumn{2}{l}{\hspace{1em} Config2.a} & \multicolumn{2}{l}{\hspace{1em} Config2.b} & \multicolumn{2}{l}{\hspace{1em} Config3.a} & \multicolumn{2}{l}{\hspace{1em} Config3.b} & \multicolumn{2}{l}{\hspace{1em} Config4.a} & \multicolumn{2}{l}{\hspace{1em} Config4.b} & \multicolumn{2}{l}{\hspace{1em} Config5.a} &  \multicolumn{2}{l}{\hspace{1em} Config5.b}\\
 & TNR & TNAS & TNR & TNAS & TNR & TNAS & TNR & TNAS & TNR & TNAS & TNR & TNAS & TNR & TNAS & TNR & TNAS & TNR & TNAS \\
\hline
&&&&&&& \multicolumn{5}{l}{Scenario 1 - $(\theta = 0.0002, \phi = 0.00002)$} &&&&&&& \\
\hline
 SVDHLA & \textbf{777} & \textbf{5} & \textbf{792} & \textbf{1} & \textbf{790} & \textbf{0} & \textbf{776} & \textbf{15} & \textbf{783} & \textbf{10} & \textbf{773} & \textbf{0} & \textbf{790} & \textbf{30} & \textbf{779} & \textbf{1} & \textbf{789} & \textbf{0}  \\
 AVDHLA & 659 & 76 & 701 & 14 & 667 & 61 & 638 & 83 & 668 & 50 & 676 & 80 & 719 & 12 & 693 & 30 & 689 & 60  \\
 VSLA & \textbf{197} & \textbf{798} & \textbf{700} & \textbf{38} & \textbf{578} & \textbf{291} & \textbf{235} & \textbf{817} & \textbf{211} & \textbf{793} & \textbf{439} & \textbf{619} & \textbf{379} & \textbf{701} & \textbf{667} & \textbf{119} & \textbf{547} & \textbf{371}  \\
\hline
&&&&&&& \multicolumn{5}{l}{Scenario 2 - $(\theta = 0.0005, \phi = 0.0005)$} &&&&&&& \\
\hline
 SVDHLA & \textbf{580} & \textbf{98} & \textbf{574} & \textbf{88} & \textbf{596} & \textbf{87} & \textbf{565} & \textbf{130} & \textbf{560} & \textbf{110} & \textbf{542} & \textbf{193} & \textbf{558} & \textbf{94} & \textbf{548} & \textbf{87} & \textbf{563} & \textbf{153}  \\
 AVDHLA & 570 & 136 & 575 & 157 & 587 & 94 & 584 & 122 & 566 & 162 & 566 & 92 & 565 & 130 & 563 & 97 & 587 & 115  \\
 VSLA & \textbf{339} & \textbf{821} & \textbf{532} & \textbf{34} & \textbf{563} & \textbf{419} & \textbf{366} & \textbf{801} & \textbf{370} & \textbf{800} & \textbf{514} & \textbf{640} & \textbf{516} & \textbf{680} & \textbf{556} & \textbf{263} & \textbf{544} & \textbf{415}  \\
\hline
&&&&&&& \multicolumn{5}{l}{Scenario 3 - $(\theta = 0.00002, \phi = 0.0002)$} &&&&&&& \\
\hline
 SVDHLA & \textbf{792} & \textbf{5} & \textbf{797} & \textbf{0} & \textbf{804} & \textbf{1} & \textbf{781} & \textbf{25} & \textbf{806} & \textbf{5} & \textbf{786} & \textbf{10} & \textbf{812} & \textbf{5} & \textbf{790} & \textbf{0} & \textbf{812} & \textbf{1}  \\
 AVDHLA & 762 & 29 & 761 & 19 & 805 & 7 & 757 & 9 & 809 & 5 & 797 & 9 & 796 & 8 & 795 & 2 & 779 & 7  \\
 VSLA & \textbf{256} & \textbf{801} & \textbf{747} & \textbf{39} & \textbf{656} & \textbf{332} & \textbf{289} & \textbf{817} & \textbf{289} & \textbf{813} & \textbf{494} & \textbf{634} & \textbf{469} & \textbf{680} & \textbf{721} & \textbf{158} & \textbf{586} & \textbf{394}  \\
\hline
\end{tabular}}
\end{table} 

\section{Application}
In this section, both of SVDHLA and AVDHLA are devoted to design a novel defense mechanism, namely Nik \cite{RJahromi2023Nik}, against the selfish mining \cite{REyal2015Miner, REyal2018Majority} attack in Bitcoin\cite{RNakamoto2008Bitcoin, RWang2019Survey, RBabaioff2012Bitcoin}. At first, the related concepts such as mining process and the selfish mining attack are introduced briefly. Then, the performed experiments to see the performance of the proposed automaton in such a complex environment are presented. Additionally, the developed simulator can be found on Github \footnote{\url{https://github.com/AliNikhalat/SelfishMining}}.

\subsection{Concepts}
Bitcoin \cite{RNakamoto2008Bitcoin} is a decentralized cryptocurrency that is introduced by Satoshi Nakamoto in 2009. It has received a lot of attentions because of the decentralization nature exhibited \cite{RNakamoto2008Bitcoin, RWang2019Survey}.

The transactions in Bitcoin network are recorded through units called block. Creating a new block needs an effort to solve a cryptopuzzle with a dedicated reward. The participants who put their resources to solve such a puzzle are called miners \cite{RNakamoto2008Bitcoin, RWang2019Survey, REyal2018Majority, RJahromi2023Nik}.

Incentive mining process guarantees the safety of Bitcoin. It means each miner will be rewarded based on shared resource. If this assumption works correctly, the decentralization of the network will be sustained \cite{RWang2019Survey}.

But handling Bitcoin in a decentralized manner is a challenging problem. Some attacks like the selfish mining \cite{REyal2018Majority} will threat the most important property of Bitcoin by keeping newly discovered block private and revealing them whenever they get rewards more than their fair-share.

Revealing kept blocks by selfish miners will make a fork in the chain of blocks. Under such a circumstance, the honest branch of the fork which is the result of the right work will be discarded. Consequently, a consensus is reached on the selfish branch of the fork \cite{RWang2021Blockchain, RZhang2017Publish}.

Our novel approach with the proposed automaton will overcome this frustrating problem in Bitcoin. To reach the ultimate goal, we can reduce this problem to a simpler problem of making decision among the created branches of the fork in each distributed miner.

\subsection{Proposed Defense}
In this section, our novel defense which takes the advantages of powerful learning automata will be described. A learning automaton is conducted as a decision maker to increase the safety in each node. It needs some predefined definitions to help the node to select one branch of fork (Maybe some of them are produced by the selfish miners) accurately. These definitions are illustrated around the traits of a branch in the fork as follows: 

\begin{algorithm}[!t]
\begin{algorithmic}[1]
 \renewcommand{\algorithmicrequire}{\textbf{Notation:}}
 \renewcommand{\algorithmicensure}{\textbf{Output:}}
\BEGIN {1}
  \IF{($K=K_{max}$)}
    \STATE L = LA.choose \_ action([\text{'}Stop\text{'}, \text{'}Shrink\text{'}])
  \ELSIF{($K=K_{min}$)}
    \STATE L = LA.choose \_ action([\text{'}Grow\text{'}, \text{'}Stop\text{'}])
  \ELSE
    \STATE L = LA.choose \_  action([\text{'}Grow\text{'}, \text{'}Stop\text{'}, \text{'}Shrink\text{'}])
  \ENDIF
\SWITCH {$L$}
\CASE {\text{'}Grow\text{'}}
  \STATE $K = K + 1$
\ENDCASE
\CASE {\text{'}Shrink\text{'}}
  \STATE $K = K - 1$
\ENDCASE
\CASE {\text{'}Stop\text{'}}
  \STATE /*Do nothing about $K$*/
\ENDCASE
\ENDSWITCH
\END
\end{algorithmic}
\caption{UpdateFailSafe(LA, K, $K_{min}$, $K_{max}$)} 
\label{algorithm_update_fail_safe}
\end{algorithm}

\begin{itemize}

\item Length of a branch which denotes by $L$ is the number of blocks in that branch.

\item Weight of a branch is denoted by $W$. From the first block till the last one of that branch, it compare with other blocks with the same height of the other branches. The weight of the branch with the most recent creation time will increase by one. 

\item Fail-safe parameter which denotes by $K$ will help the miner to choose a branch by $L$ or $W$. If the length of one branch of that fork is longer than the others by $K$, that branch will be chosen. Otherwise $W$ parameter is considered for choosing the fork.

\item Decision making time denotes by $\tau$ is a time for a miner to check if any forks exist. If such a fork exists, the miner should choose one of them considering $K$ parameter.

\item Time parameter defined to configure the next value for $K$ using the automaton. This parameter denotes by $\theta$. Each $\theta$ consists of several $\tau$.

\end{itemize}

The considered algorithm to help the miner on making a decision is follows:

\begin{enumerate}

\item Calculate $L$ of each branch.

\item Calculate $W$ of each branch.

\item After sorting by the longest chain, If the difference of the first longest branch from the second longest is greater than $K$, miner must choose the longest branch. Otherwise, it must choose the heaviest chain by the calculated $W$ parameter.
\item If $\tau$ reaches the end, the learning automaton should choose the next value for $K$. Usually, $K$ swings between $K_{min}$ and $K_{max}$. The automaton has three options: 1-\textbf{Grow} for increasing $K$ by one, 2-\textbf{Stop} To leave $K$ unchanged, 3-\textbf{Shrink} for decreasing $K$ by one. Such a algorithm is related to updating the fail-safe parameter that is shown in Algorithm \ref{algorithm_update_fail_safe}.

\item If $\theta$ reaches the end, the automaton should get feedback from the environment. Actually, we have designed a virtual environment for the automaton to inform about its decision. The reinforcement signal is calculated by the number of decisions which has made by $W$ divides by total number of decisions. Total number of decisions  are consisted of number of height decision ($L$) plus number of weight decision ($W$). The following equation will demonstrate $\beta$ parameter of the automaton.

\begin{equation}
\label{application_beta_calculation}
\beta = \frac{Number\;of\;Weight\;Decisions}{Total\;Number\;of\;Decisions}
\end{equation}   

\end{enumerate}

\subsection{Evaluation}
Two metrics are considered to evaluate the performance of the automaton against the selfish mining attack as well as what is defined in the section \ref{section:evaluate_metric}. These metrics are:

\begin{enumerate}

\item \textbf{Relative Revenue}: This metric is used to measure revenue of miner based on the revenue of the others. In fact, we want to assess the margin between the revenue of the miner when the selfish attack is performed in the network, and when the network has the incentive-compatibility property. This metric is calculated as follows:

\begin{equation}
\label{application_metric_relative_revenue}
\frac{Number\;of\;Mined\;Block\;by\;i\textsuperscript{th}\;Miner}{Total\;Number\;of\;Mined\;Blocks}
\end{equation}  

\item \textbf{Lower Bound Threshold}: This metric evaluates the minimum computing power that a selfish miner should provide to start an attack.

\end{enumerate}

\subsection{Experiment}
This experiment aims to provide a comparison of the proposed defense using more sophisticated learning automaton with the well-known defense, namely tie-breaking. In the tie-breaking, when a miner is known of the fork, choose one of the branches of that fork in a uniform random manner.

We assume that there are two categories of miners in the network. All of the miners which obey Bitcoin protocols form a honest pool. By contradiction, all of the selfish miners together form a selfish pool. We prefer to study the performance of the defense from the selfish pool point of view.

To reach our goal, 10000 blocks were generated. $K$ can swing between $K_{min} = 1$ and $K_{max} = 5$. Also, five configurations for VASLA are considered respectively:1-$P$ model with $\lambda_1 = 0$ and $\lambda_2 = 0$, 2-$L_{R\_I}$ model with $\lambda_1 = 0.01$ and $\lambda_2 = 0$, 3-$L_{P\_I}$ model with $\lambda_1 = 0$ and $\lambda_2 = 0.01$, 4-$L_{R\_P}$ with with $\lambda_1 = 0.01$ and $\lambda_2 = 0.01$, 5-$L_{R\_{\epsilon P}}$ with with $\lambda_1 = 0.1$ and $\lambda_2 = 0.01$.

\begin{figure*}[!t]
\centering
\subfloat[Config 1]{\includegraphics[width=0.33\textwidth]{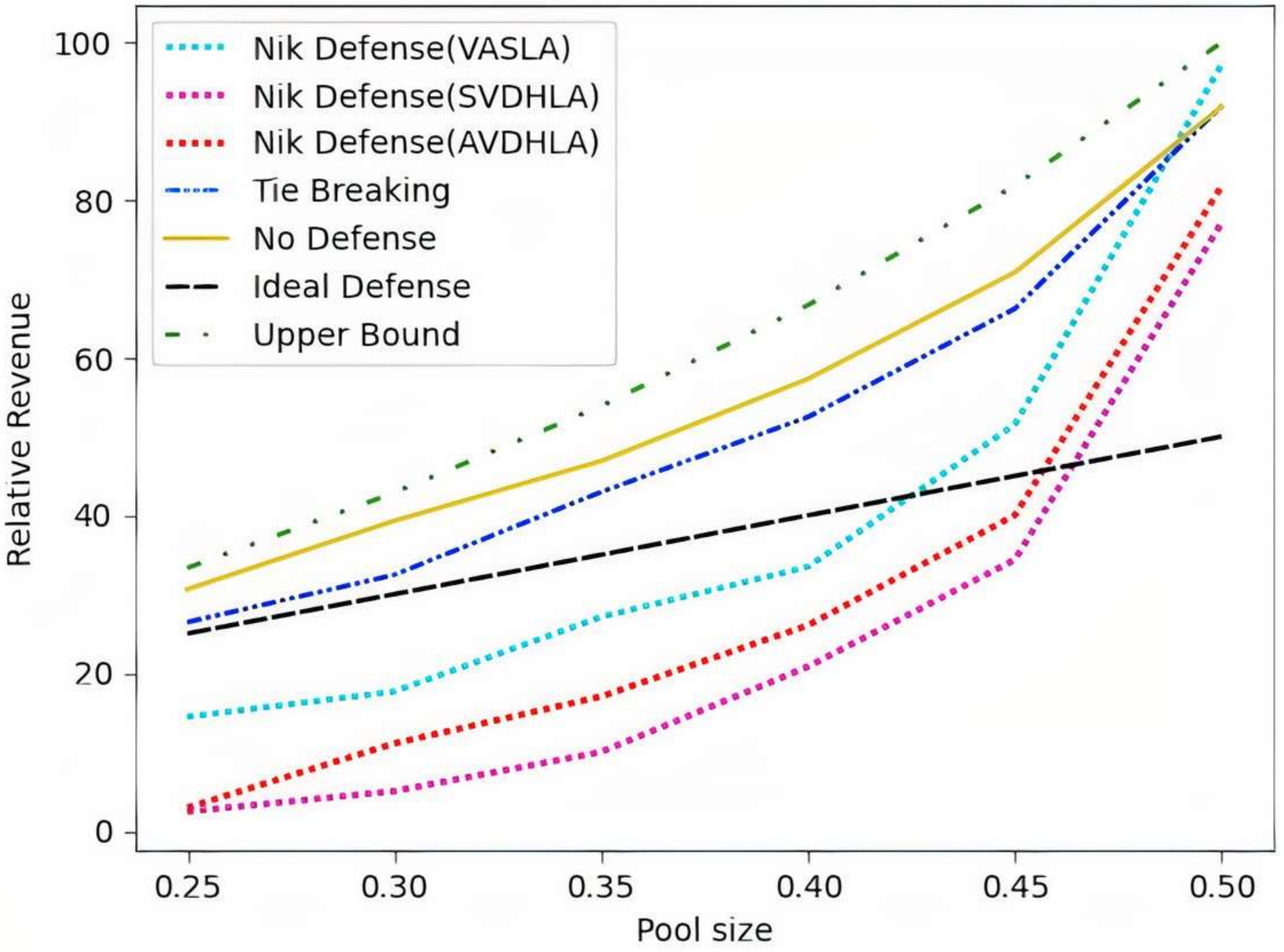}} 
\subfloat[Config 2]{\includegraphics[width=0.33\textwidth]{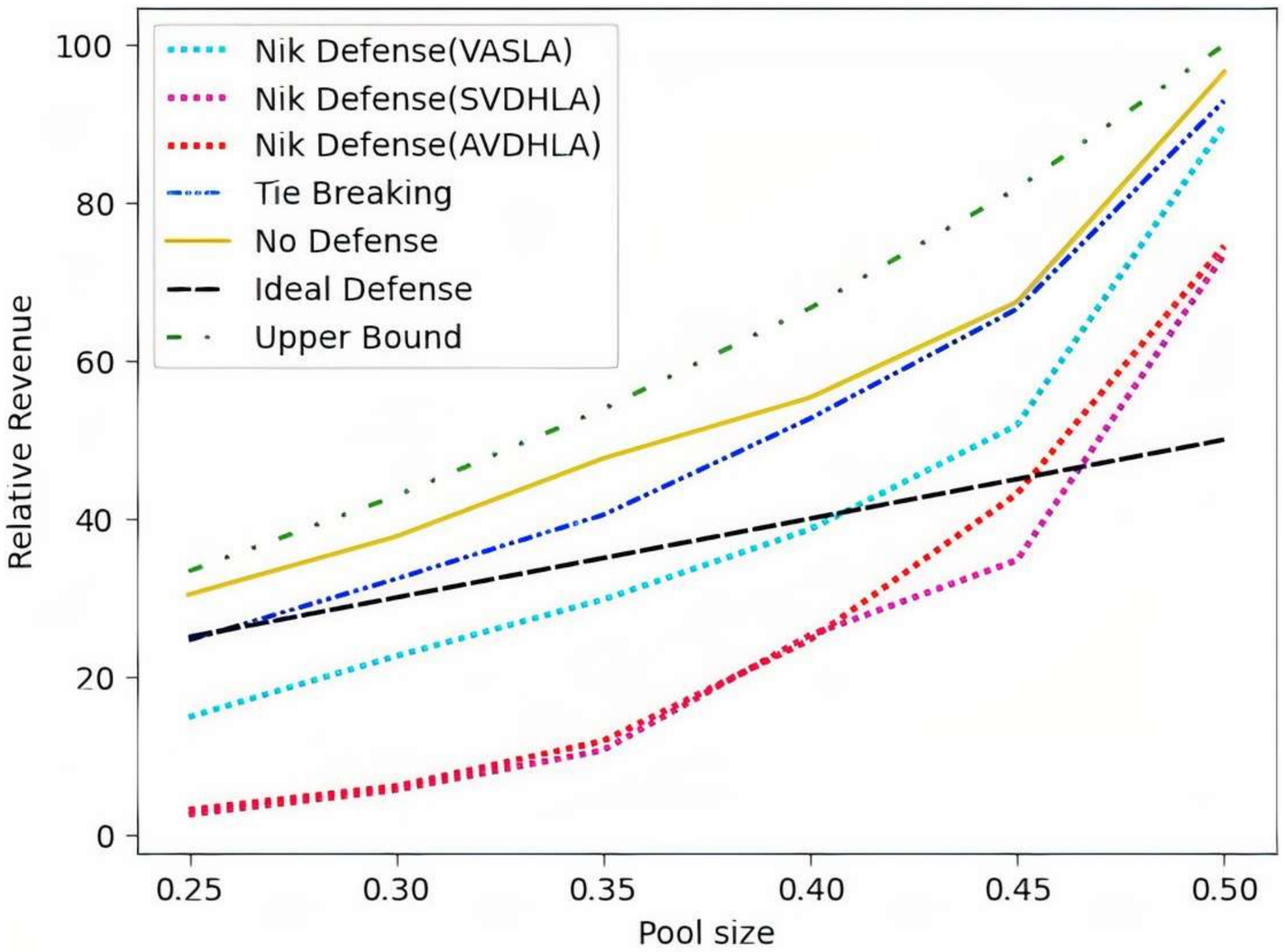}}
\subfloat[Config 3]{\includegraphics[width=0.33\textwidth]{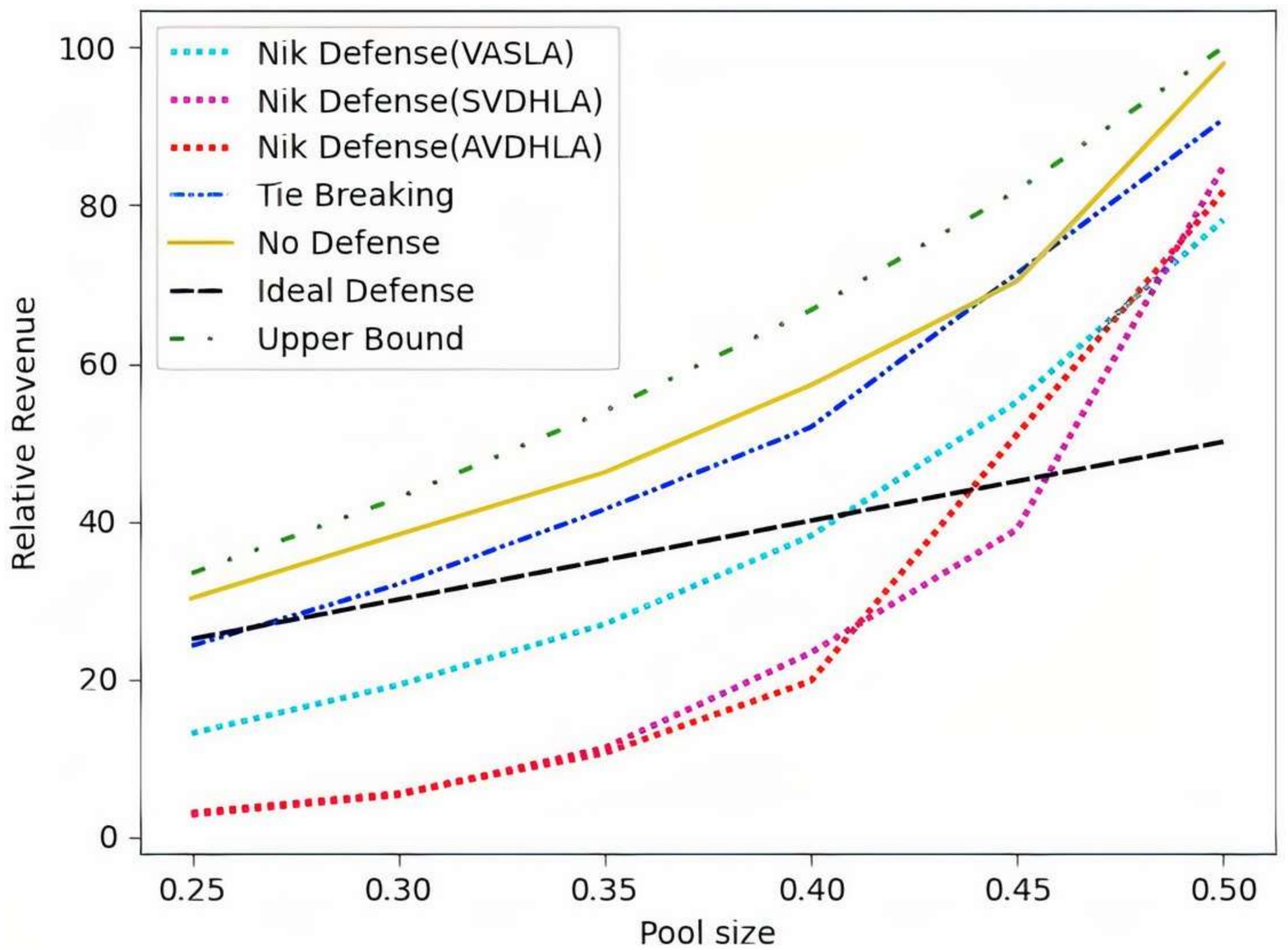}}\\
\subfloat[Config 4]{\includegraphics[width=0.33\textwidth]{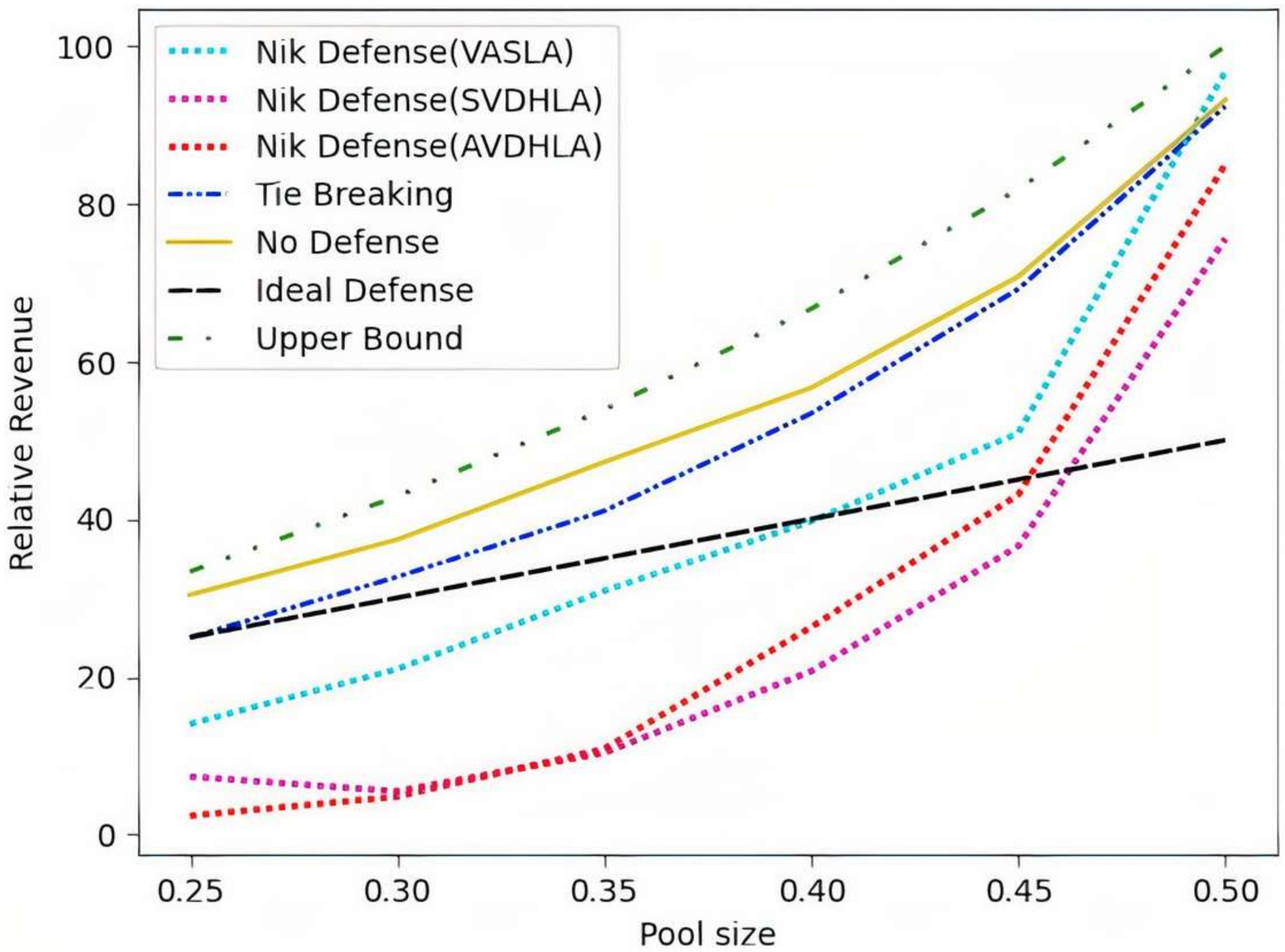}}
\subfloat[Config 5]{\includegraphics[width=0.33\textwidth]{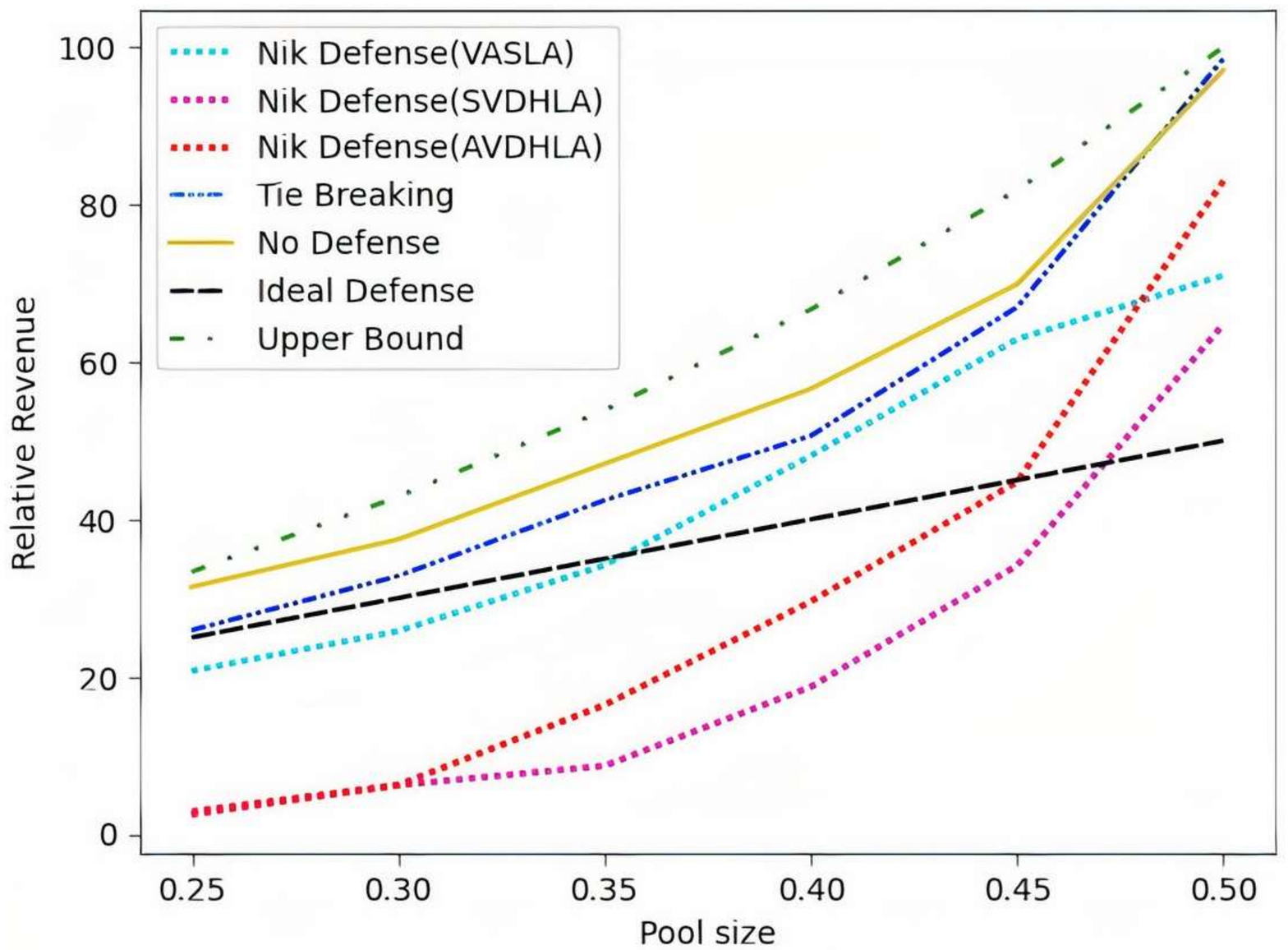}}\\
\caption{The performance of Nik defense in compare to tie-breaking with respect to relative revenue }
\label{fig_app_experiment}
\end{figure*}

The results in Fig.\ref{fig_app_experiment} show interesting points. At first, they show the superiority of the proposed defense over the tie-breaking with respect to the relative revenue of the selfish miners. Secondly, among various kind of automata, AVDHLA outperforms the others in conjunction with relative revenue. Both of evidences lead us to the outstanding performance of automaton in the complex environment like the blockchain. Also, it shows that AVDHLA adopt each depth of predefined actions correctly to get higher reward. 

As well as the relative revenue, the other metric, namely the lower bound threshold is studied. the results which is demonstrated in Table \ref{table_ex_nik_defense} reveal interesting points another time from the other side. The proposed defense using the power of VDHLA increases the lower bound threshold from 0.25 up to 0.4 approximately. VDHLA tries to get rewards based on the forks that created in the network. It can detect when a fork needs a decision by the weight of height parameter precisely. As a matter of fact, this decision in SVDHLA or AVDHLA is made from the actual understanding of the automaton from unknown environment like the blockchain.

\begin{table}[!h]
\renewcommand{\arraystretch}{1.3}
\caption{Experimental Results of the Nik Defense with Respect to the Lower Bound Threshold}
\label{table_ex_nik_defense}
\centering
\resizebox{\columnwidth}{!}{
\begin{tabular}{lccccc}
\hline
Defense & $P$ & $L_{R\_I}$ & $L_{P\_I}$ & $L_{R\_P}$ & $L_{R\_{\epsilon P}}$\\
\hline
\hline
Nik (AVDHLA) & \textbf{0.45} & \textbf{0.45} & \textbf{0.42} & \textbf{0.44} & \textbf{0.45} \\
Nik (SVDHLA) & 0.46 & 0.46 & 0.45 & 0.46 & 0.46 \\
Nik (FSLA) & \textbf{0.42} & \textbf{0.40} & \textbf{0.40} & \textbf{0.41} & \textbf{0.36} \\
Tie-breaking & 0.25 & 0.25 & 0.25 & 0.25 & 0.25\\
\hline
\end{tabular}}
\end{table}

Eventually, we can trust VDHLA as decision maker in the blockchain networks. Also, some extra experiments are designed to investigate the effect of other parameters on the quality of defense with respect to introduced metrics in Appendix \ref{section:appendix_exp_nik_defense}. 

\section{Conclusion and Future Research Directions}
The greatest weakness of $L_{KN,K}$ learning automaton as the most prominent family member of FSLA is the problem of choosing an appropriate depth. Such a problem has been remaining unsolved since the appearance of  $L_{KN,K}$. Lack of the effective solution has caused this type of learning automaton to fade in recent years. Therefore, we were motivated to solve this problem using an intelligent self-adaptive model, named VDHLA.

The proposed model was evaluated in various aspects. From getting more rewards and less penalties perspective, it outperformed FSLA in different environments including: Stationary, Non-Stationary such as Markovian Switching and State-Dependent.

In addition to the predefined environments, we tries to apply the new model in an uncertain environment like blockchain. We designed a new strategy against the selfish mining attack which threatened the incentive-compatibility of Bitcoin. In that strategy, VDHLA performed successfully compared to FSLA in the designed experiments.

Consequently, the proposed model can be a robust alternative to FSLA. Since FSLAs have been used in a wide range of application in different areas, VDHLA also can be used instead of them with just a bit of modification.

Moreover, it is the first time that these automata combined together in such a way. There will be many open problems in the area of combining simple but powerful learning automata. we are so hopeful such a combination will promise the bright future ahead of AI. 

\appendices
\section{Experiment 1.1 Extras \label{section:appendix_exp_1_1}}
\begin{figure}[!t]
\centering
\subfloat[TNR of SVDHLA]{\includegraphics[width=0.25\textwidth]{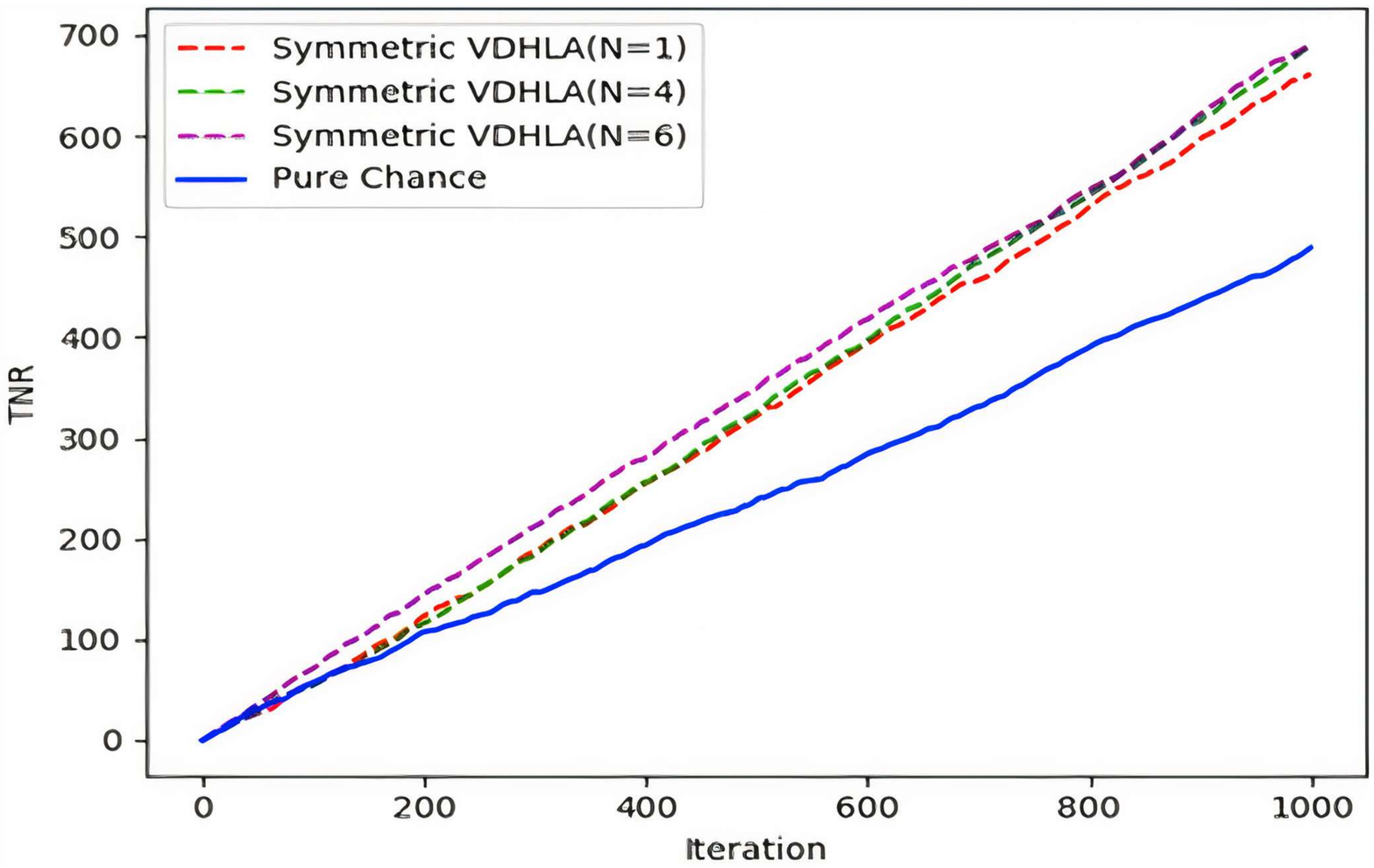}} 
\subfloat[TNAS of SVDHLA]{\includegraphics[width=0.25\textwidth]{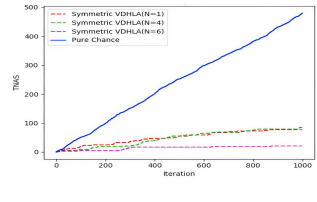}}\\ 
\subfloat[TNR of AVDHLA]{\includegraphics[width=0.25\textwidth]{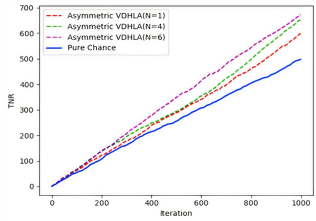}} 
\subfloat[TNAS of AVDHLA]{\includegraphics[width=0.25\textwidth]{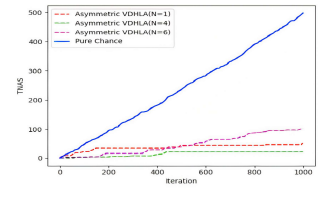}}\\ 
\caption{Experimental results of Ex 1.1 with reward probability vector of (0.3, 0.7)}
\label{fig_VDHLA_EX_1_1_2}
\end{figure}

\begin{figure}[!t]
\centering
\subfloat[TNR of SVDHLA]{\includegraphics[width=0.25\textwidth]{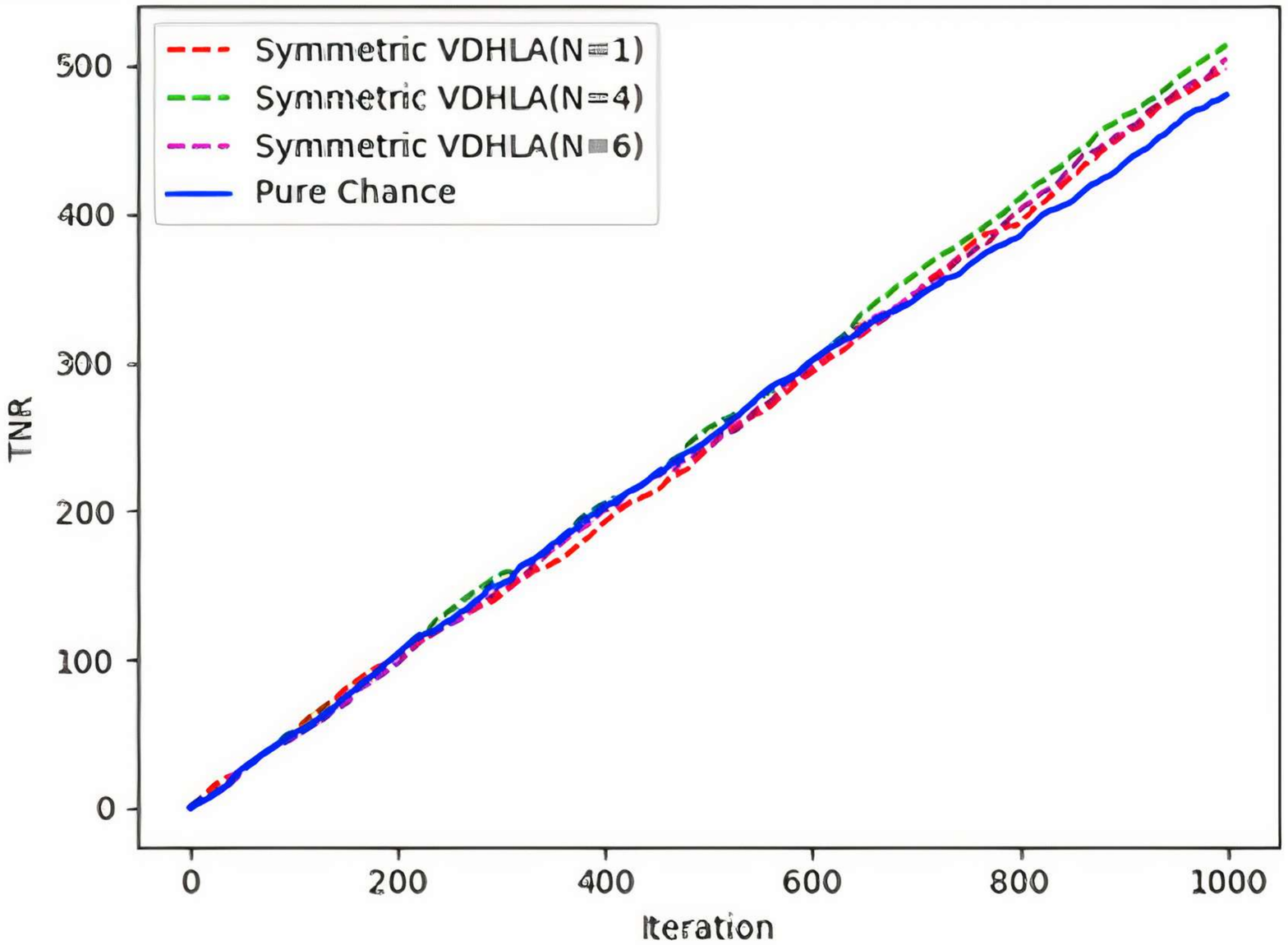}} 
\subfloat[TNAS of SVDHLA]{\includegraphics[width=0.25\textwidth]{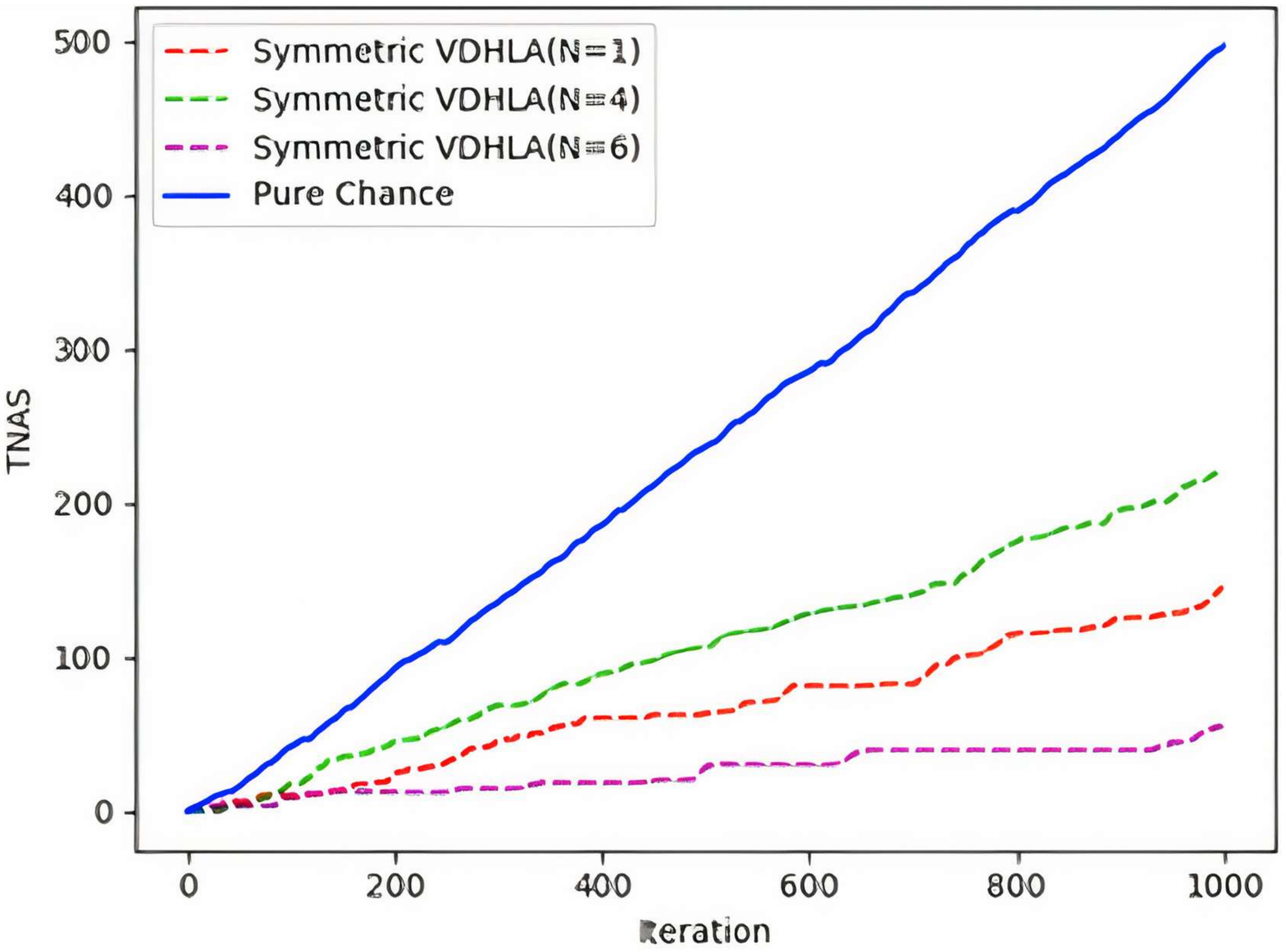}}\\ 
\subfloat[TNR of AVDHLA]{\includegraphics[width=0.25\textwidth]{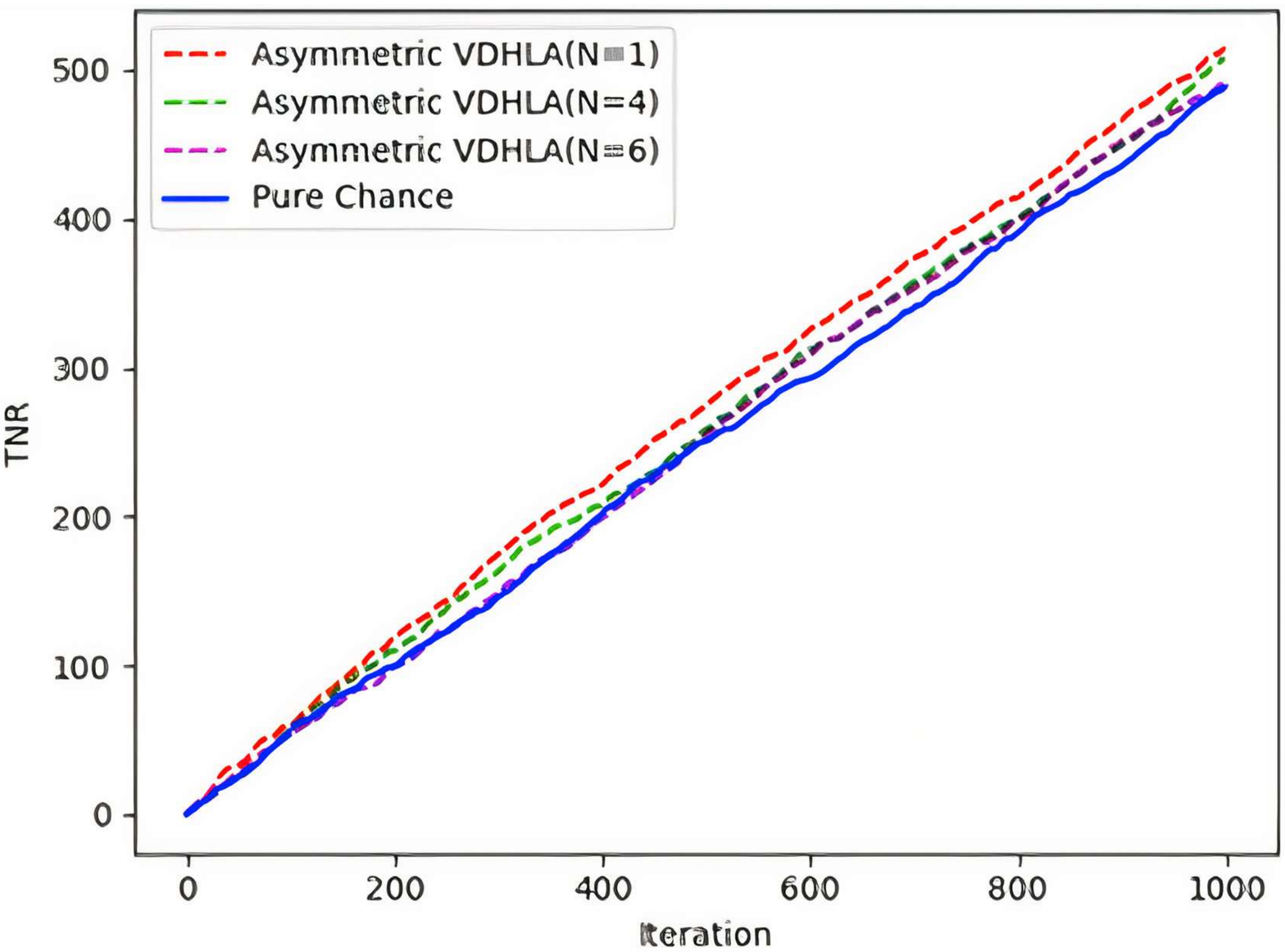}} 
\subfloat[TNAS of AVDHLA]{\includegraphics[width=0.25\textwidth]{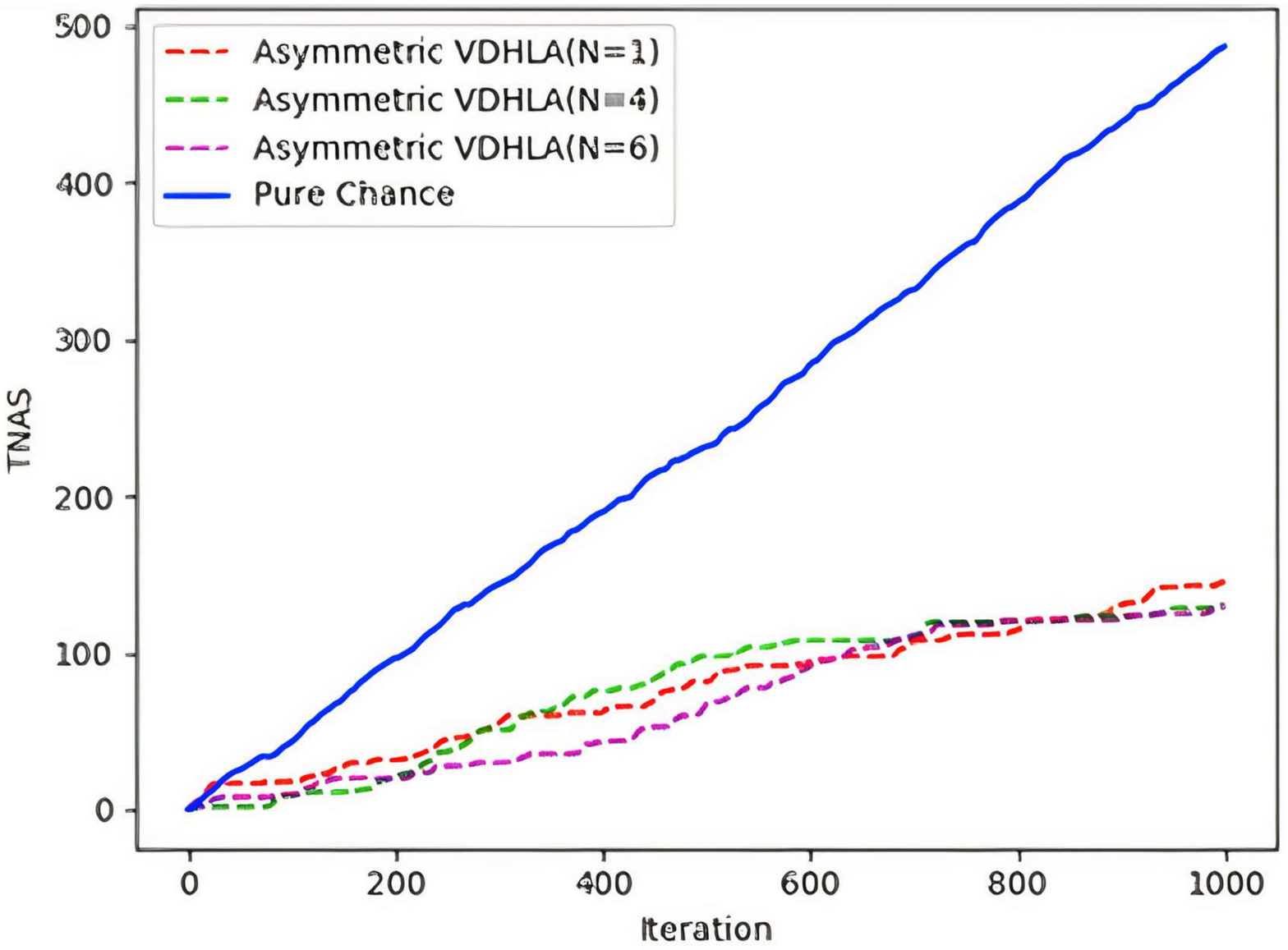}}\\ 
\caption{Experimental results of Ex 1.1 with reward probability vector of (0.5, 0.5)}
\label{fig_VDHLA_EX_1_1_3}
\end{figure}

Extra experiments are designed to learn more about the capability of new automata. This time the probability of getting reward of $\alpha_1$ increases up to 0.3, meanwhile $\alpha_2$ decreases to 0.7. For the the third time, a totally random environment is considered with the probability of getting reward for both actions are 0.5 equally. Results of these experiments are shown in Fig.\ref{fig_VDHLA_EX_1_1_2} and Fig.\ref{fig_VDHLA_EX_1_1_3} respectively.

Results shows that both of them get rewards more than pure chance learning automaton. On the other hand, they have a lot less number of action switching. This leads us to the capability of learning in VDHLA clearly.

Besides, we should mention that the greater difference between the probability of getting rewards between two actions (like (0.9, 0.1) probability vector in Experiment 1.1), the better VDHLA can adopt itself with the outer environment. Because VDHLA increases its depth decisively. As a result don't change the chosen action easily.

\section{Experiment 1.2 Extras \label{section:appendix_exp_1_2}}
Completing our comparison of VDHLA family with FSLA needs more experiments. To do this, experiment 1.2 is repeated for learning automata with 2 and 5 actions. Table \ref{table_ex_1_2_SVDHLA_results_appendix}, \ref{table_ex_1_2_1_AVDHLA_results_appendix} show the results.

\begin{table}[!h]
\renewcommand{\arraystretch}{1.3}
\caption{Experimental Results of Experiment 1.2 for SVDHLA}
\label{table_ex_1_2_SVDHLA_results_appendix}
\centering
\resizebox{\columnwidth}{!}{
\begin{tabular}{lcccccccc}
\hline
Model & \multicolumn{2}{l}{\;\;\;\;Config 1} & \multicolumn{2}{l}{\;\;\;\;Config 2} & \multicolumn{2}{l}{\;\;\;\;Config  3} & \multicolumn{2}{l}{\;\;\;\;Config 4} \\
& TNR & TNAS & TNR & TNAS & TNR & TNAS & TNR & TNAS\\
\hline
&&&&K = 2&&&&\\
\hline
SVDHLA & \textbf{800} & \textbf{46} & \textbf{799} & \textbf{12} & \textbf{839} & \textbf{0} & \textbf{794} & \textbf{2} \\
FSLA & 671 & 329 & 768 & 32 & 781 & 6 & 790 & 3 \\
\hline
&&&&K = 5&&&&\\
\hline
SVDHLA & \textbf{783} & \textbf{30} & \textbf{770} & \textbf{35} & \textbf{803} & \textbf{1} & \textbf{807} & \textbf{0} \\
FSLA & 459 & 541 & 763 & 42 & 785 & 17 & 798 & 1 \\
\hline
\end{tabular}}
\end{table}

\begin{table}[!h]
\renewcommand{\arraystretch}{1.3}
\caption{Experimental Results of Experiment 1.2 for AVDHLA}
\label{table_ex_1_2_1_AVDHLA_results_appendix}
\centering
\resizebox{\columnwidth}{!}{
\begin{tabular}{lcccccccc}
\hline
Model & \multicolumn{2}{l}{\;\;\;\;Config 1} & \multicolumn{2}{l}{\;\;\;\;Config 2} & \multicolumn{2}{l}{\;\;\;\;Config  3} & \multicolumn{2}{l}{\;\;\;\;Config 4} \\
& TNR & TNAS & TNR & TNAS & TNR & TNAS & TNR & TNAS\\
\hline
&&&&K = 2&&&&\\
\hline
AVDHLA & \textbf{8032} & \textbf{31} & \textbf{8075} & \textbf{25} & \textbf{8045} & \textbf{8} & \textbf{8000} & \textbf{1} \\
FSLA & 6736 & 3264 & 7986 & 142 & 7953 & 31 & 8000 & 4 \\
\hline
&&&&K = 5&&&&\\
\hline
AVDHLA & \textbf{7934} & \textbf{123} & \textbf{8018} & \textbf{37} & \textbf{7990} & \textbf{26} & \textbf{8008} & \textbf{7} \\
FSLA & 4600 & 5400 & 7667 & 475 & 7980 & 45 & 8002 & 4 \\
\hline
\end{tabular}}
\end{table}

These results show the superiority of VDHLA family over FSLA with respect to TNR and TNAS. Lower values of $K$ can help the automaton find the favorable action easier, therefore TNR is higher than an experiment with higher number of $K$. Also, this has effect on reducing total number of action switching because the automaton finds the favorable action among the lower number of actions simply.

\section{Experiment 1.4 Extras \label{section:appendix_exp_1_4}}
This appendix dedicates to the phase two of experiment 1.4. It is repeated for the second time with the inner VASLAs of kind $L_{R\_I}$. The configurations are all the same. As a matter of fact, we want to investigate the effect of inner VASLAs on the results.

The results in Table \ref{table_ex_1_4_results_appendix} reveal an interesting point. The AVDHLA is superior the SVDHLA with respect to TNR and TNAS. we can claim that SVDHLA is more sensitive to updating scheme of the inner VASLA. Because there is just one VASLA is conducted in the SVDHLA, hence depth of all actions depends on this VASLA. Consequently, updating scheme has an important role on the performance of SVDHLA.

\begin{table}[!h]
\renewcommand{\arraystretch}{1.3}
\caption{Experimental Results of Experiment 1.4 with respect to TNR and TNAS}
\label{table_ex_1_4_results_appendix}
\centering
\resizebox{\columnwidth}{!}{
\begin{tabular}{lcccccccc}
\hline
Model & \multicolumn{2}{l}{\;\;\;\;Config 1} & \multicolumn{2}{l}{\;\;\;\;Config 2} & \multicolumn{2}{l}{\;\;\;\;Config  3} & \multicolumn{2}{l}{\;\;\;\;Config 4} \\
& TNR & TNAS & TNR & TNAS & TNR & TNAS & TNR & TNAS\\
\hline
&&&&K = 2&&&&\\
\hline
AVDHLA & \textbf{8016} & \textbf{38} & \textbf{7884} & \textbf{26} & \textbf{7977} & \textbf{5} & \textbf{8019} & \textbf{9} \\
SVDHLA & 7927 & 296 & 7317 & 1820 & 7992 & 24 & 7993 & 8 \\
\hline
&&&&K = 5&&&&\\
\hline
AVDHLA & \textbf{7975} & \textbf{78} & \textbf{7970} & \textbf{28} & \textbf{7985} & \textbf{6} & \textbf{8055} & \textbf{3} \\
SVDHLA & 6645 & 3765 & 7970 & 70 & 7959 & 5 & 7993 & 3 \\
\hline
&&&&K = 9&&&&\\
\hline
AVDHLA & \textbf{7492} & \textbf{638} & \textbf{7965} & \textbf{72} & \textbf{8057} & \textbf{11} & \textbf{7986} & \textbf{8} \\
SVDHLA & 7948 & 81 & 7982 & 27 & 7885 & 144 & 7869 & 18 \\
\hline
\end{tabular}}
\end{table}

\section{Markovian Switching Extras \label{section:appendix_markovian_switching_exp}}
Several extra experiments are designed to assess the performance of the proposed automaton in Markovian Switching environment in order to compare it with FSLA with respect to TNR and TNAS metrics.

\begin{figure*}[!t]
\centering
\subfloat[Scenario 1]{\includegraphics[width=0.5\textwidth]{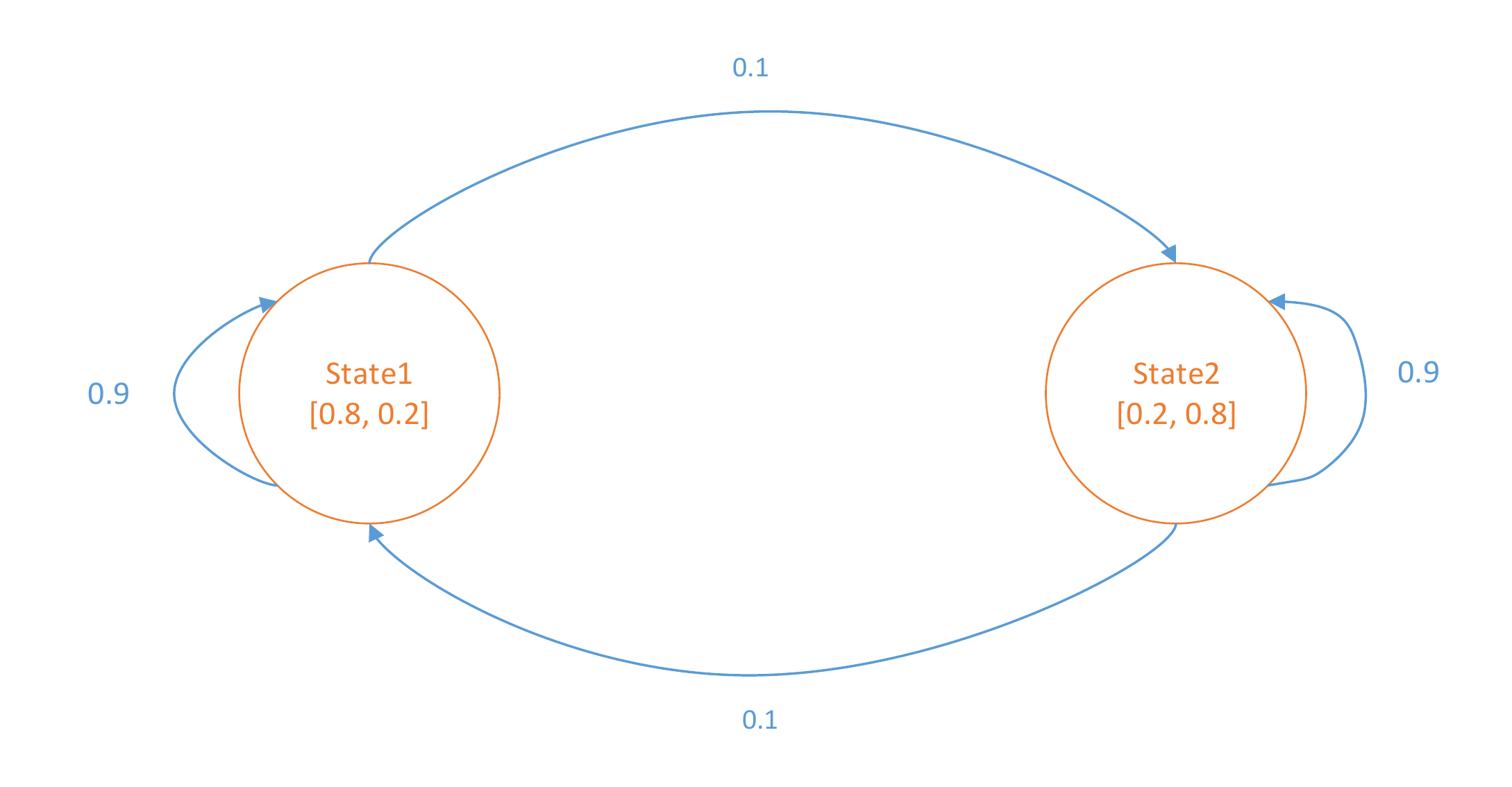}} 
\subfloat[Scenario 2]{\includegraphics[width=0.5\textwidth]{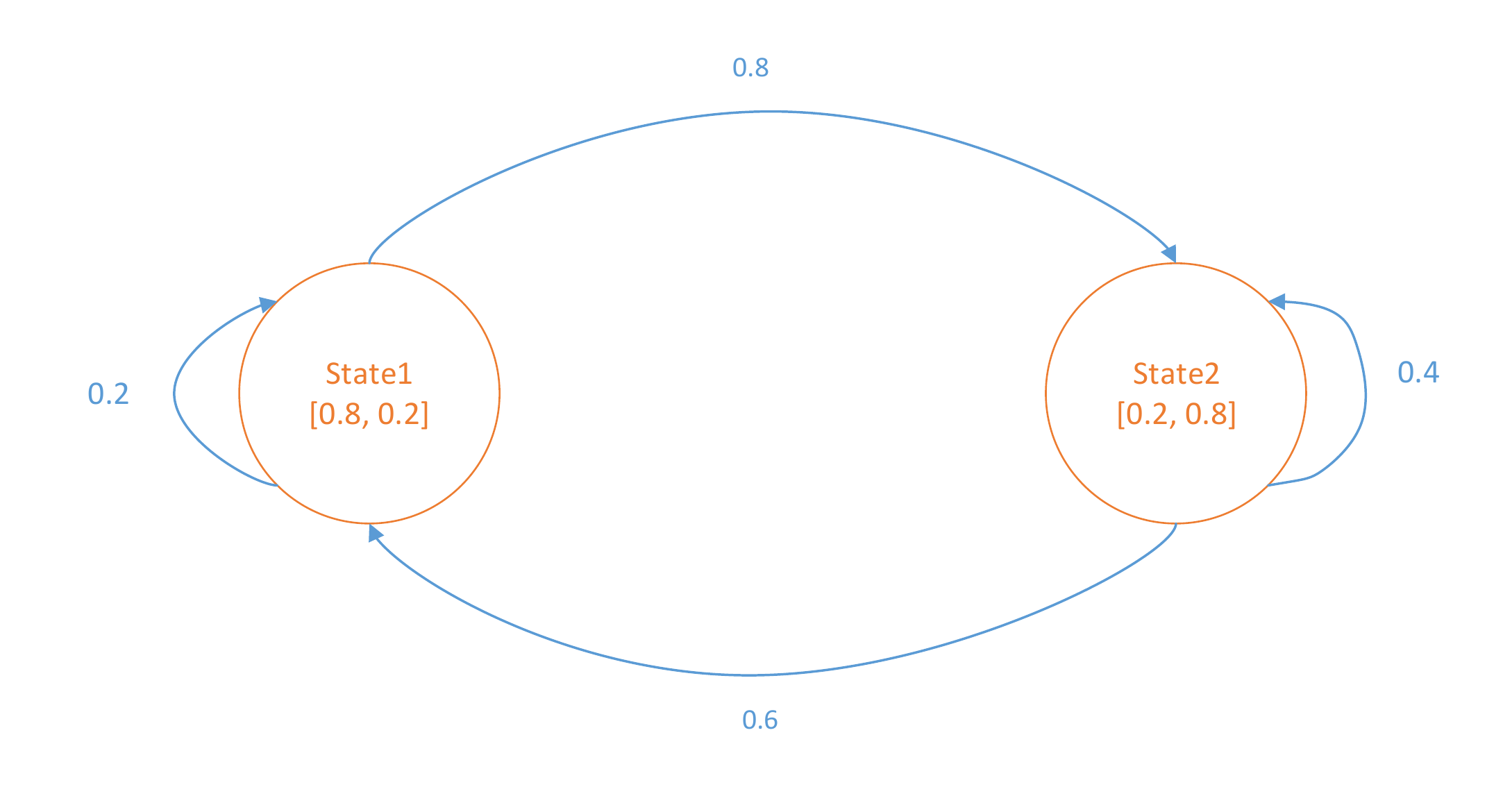}}\\ 
\subfloat[Scenario 3]{\includegraphics[width=0.5\textwidth]{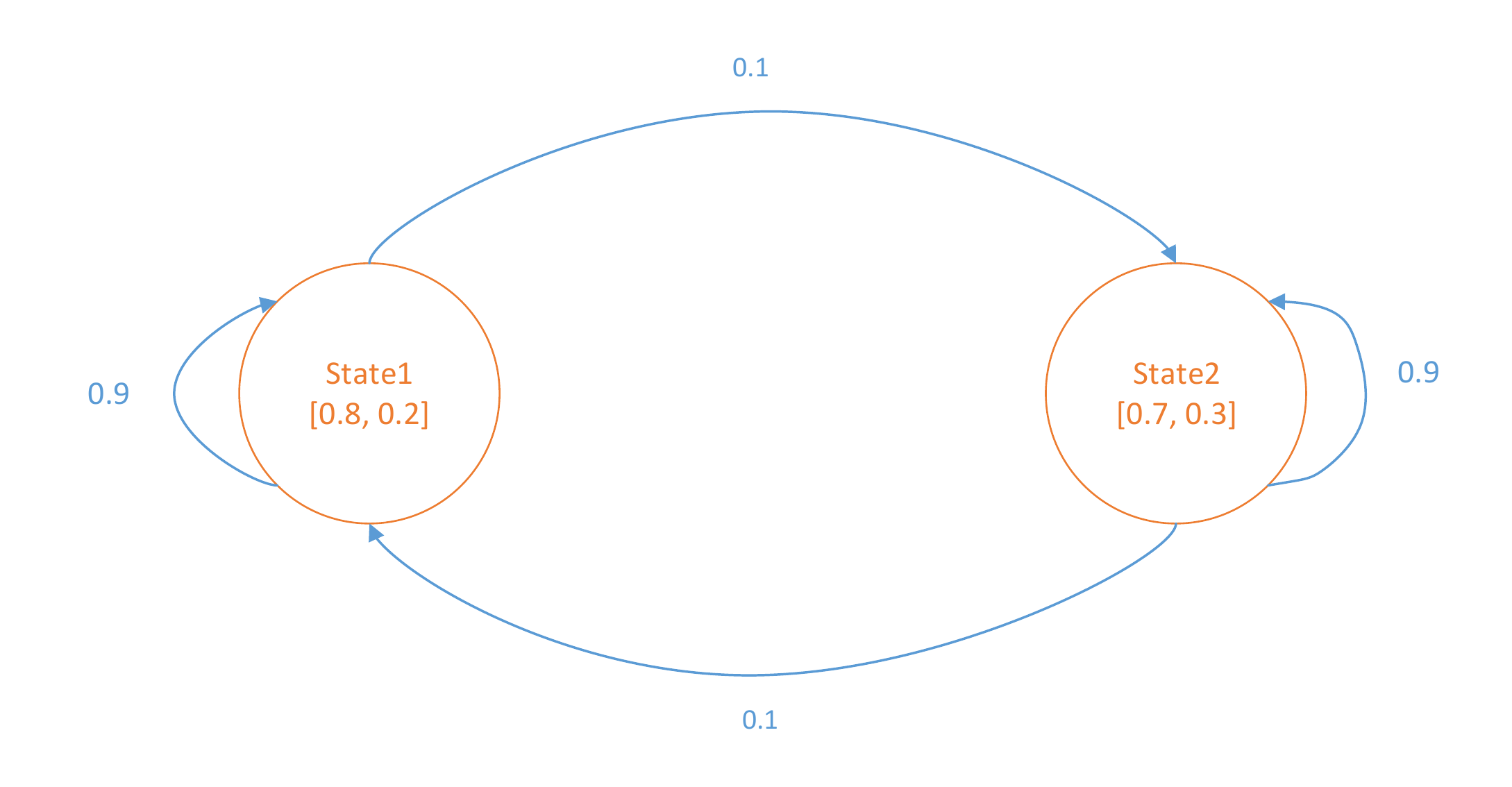}}
\subfloat[Scenario 4]{\includegraphics[width=0.5\textwidth]{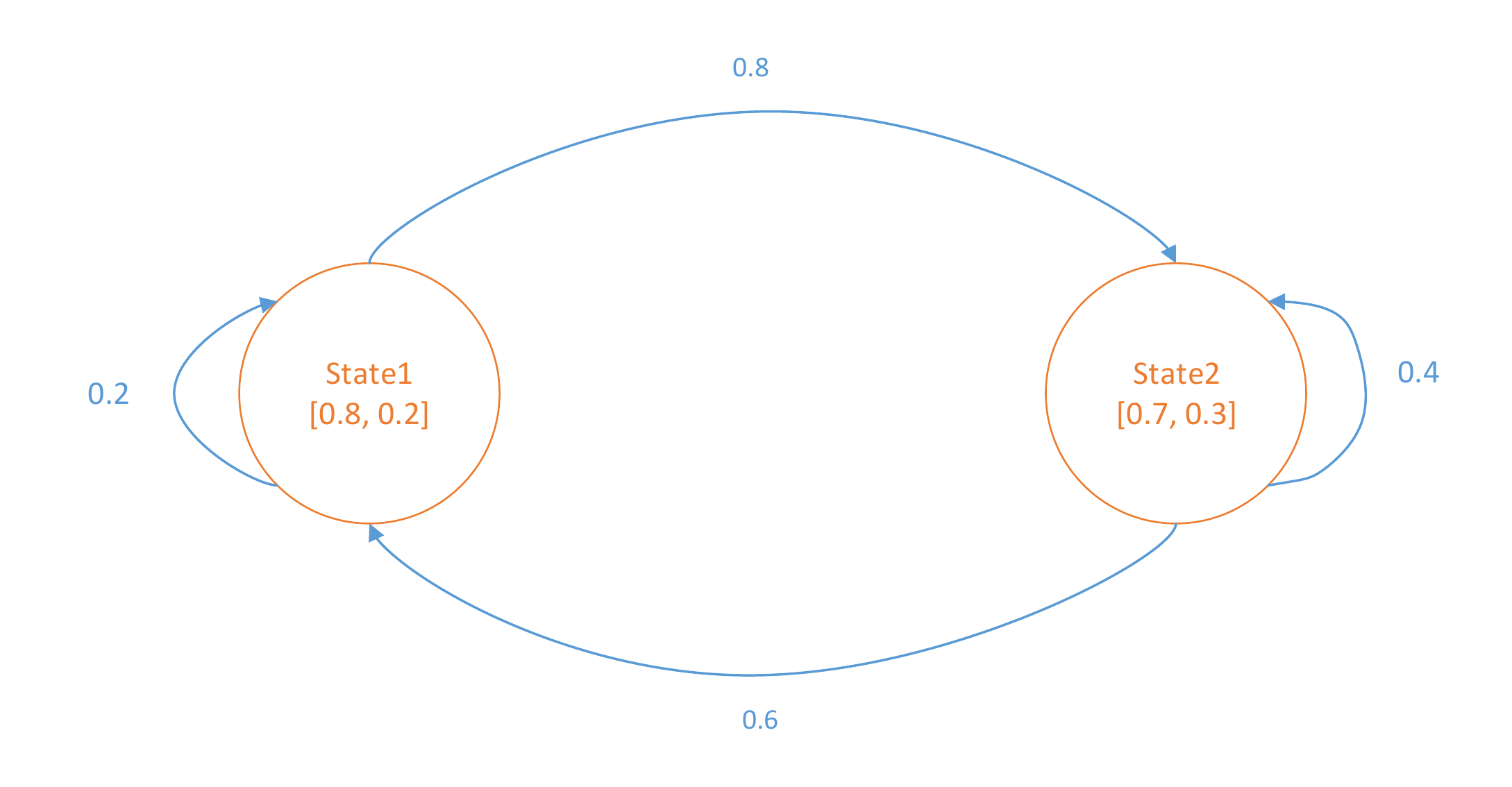}}\\ 
\caption{Four scenarios of Markovian Switching environment}
\label{fig_markovian_switching_scanarios_appendix}
\end{figure*}

For better understanding of the traits of VDHLA in Markovian Switching environment, four scenarios of Fig.\ref{fig_markovian_switching_scanarios_appendix} with a simple two state markov chain is considered as follows:

\begin{enumerate}

\item This scenario devotes to a Markovian Switching environment with two states which inclines to stay in the current state and switching from one state to the other causes the favorable action to change. More details are explained in \ref{sec:Markovian_Switching_Scenario_1}.

\item This scenario devotes to a Markovian Switching environment with two states which inclines to change the current state and switching from one state to the other causes the favorable action to change. More details are explained in \ref{sec:Markovian_Switching_Scenario_2} .

\item This scenario devotes to a Markovian Switching environment with two states which inclines to stay in the current state and switching from one state to the other just decreases the probability of the favorable action, but the favorable action is the same as before. More details are explained in \ref{sec:Markovian_Switching_Scenario_3}.

\item This scenario devotes to a Markovian Switching environment with two states which inclines to change the current state and switching from one state to the other just decreases the probability of the favorable action, but the favorable action is the same as before. More details are explained in \ref{sec:Markovian_Switching_Scenario_4} .

\end{enumerate}

The desired configurations in four scenarios have common property such as: 2, 5, 9 is considered for number of allowed actions ($K$), inner VASLAs are $L_{R\_I}$ with $\lambda_1 = 0.01$ and $\lambda_2 = 0$, and the initial depth is 1, 3, 5, 7 for configuration 1 till 4 respectively. All of the reported result are for 10000 iterations.

\subsection{Scenario 1 \label{sec:Markovian_Switching_Scenario_1}}
The aim of this scenario is evaluating the proposed automaton in Marovian Switching environment with respect to TNR and TNAS. The considered environment is a two-state markov chain in which switching from one state to the other changes the favorable action. In addition, the tendency to change from one state to the other is very low.

From mathematical standpoint, in transition matrix of the markov chain, the probability of $a_{ij} \quad i = j$ is more than $a_{ij} \quad i \neq j$. The following transition matrix is considered to reach our desired condition.

\begin{equation}
\label{equ_markovian_1_transition_appendix}
T = \left(\begin{IEEEeqnarraybox}[][c]{,c/c/c,}
0.9& &0.1\\
0.1& &0.9
\end{IEEEeqnarraybox}\right)
\end{equation}

Following reward matrix is regarded to satisfy the condition of this experiment. Also, Fig.\ref{fig_markovian_switching_scanarios_appendix} a visualizes the desired Markovian Switching environment.

\begin{equation}
\label{equ_markovian_1_reward_appendix}
R = \left(\begin{IEEEeqnarraybox}[][c]{,c/c/c,}
0.8& &0.2\\
0.2& &0.8
\end{IEEEeqnarraybox}\right)
\end{equation}

Before analyzing the results which is shown in Table \ref{table_ex_markovian_switching_results_1_appendix}, it is better to convert the Markovian Switching environment to the stationary environment with the steady state analysis of the markov chain as follows:

\begin{IEEEeqnarray}{c}
\label{equ_markovian_1_conversion_appendix}
v_{\infty}T = v_{\infty} \IEEEyesnumber\IEEEyessubnumber*\\
v_{\infty}
\left(\begin{IEEEeqnarraybox}[][c]{,c/c/c,}
0.9& &0.1\\
0.1& &0.9
\end{IEEEeqnarraybox}\right)
= v_{\infty}\\
\left(\begin{IEEEeqnarraybox}[][c]{,c/c/c,}
x& &y
\end{IEEEeqnarraybox}\right)
\left(\begin{IEEEeqnarraybox}[][c]{,c/c/c,}
0.9& &0.1\\
0.1& &0.9
\end{IEEEeqnarraybox}\right) 
= \left(\begin{IEEEeqnarraybox}[][c]{,c/c/c,}
x& &y
\end{IEEEeqnarraybox}\right)\\
\begin{cases}
	0.9 x + 0.1 y = x \\
    0.1 x + 0.9 y = y \\
    x + y = 1
\end{cases} \\
\begin{cases}
	x = 0.5\\
    y = 0.5
\end{cases}
\end{IEEEeqnarray}

Multiplication of the steady state matrix to the reward probability matrix of Markovian Switching environment yields to the reward probability matrix of the equivalent stationary environment.

\begin{IEEEeqnarray}{c}
\label{equ_markovian_1_transition_appendix}
v_{\infty}R_{Markovian \enspace Switching} = R_{Stationary} \IEEEyesnumber\IEEEyessubnumber*\\
\left(\begin{IEEEeqnarraybox}[][c]{,c/c/c,}
0.5& &0.5
\end{IEEEeqnarraybox}\right)
\left(\begin{IEEEeqnarraybox}[][c]{,c/c/c,}
0.8& &0.2\\
0.2& &0.8
\end{IEEEeqnarraybox}\right)
=
\left(\begin{IEEEeqnarraybox}[][c]{,c/c/c,}
0.5& &0.5
\end{IEEEeqnarraybox}\right)
\end{IEEEeqnarray}

Consequently, an automaton deals with a completely random environment in which both of action 1 and action 2 have the same probability of being reward equal to 0.5.

\begin{table}[!h]
\renewcommand{\arraystretch}{1.3}
\caption{Experimental Results of the First Scenario with Respect to TNR and TNAS}
\label{table_ex_markovian_switching_results_1_appendix}
\centering
\resizebox{\columnwidth}{!}{
\begin{tabular}{lcccccccc}
\hline
Model & \multicolumn{2}{l}{\;\;\;\;Config 1} & \multicolumn{2}{l}{\;\;\;\;Config 2} & \multicolumn{2}{l}{\;\;\;\;Config  3} & \multicolumn{2}{l}{\;\;\;\;Config 4} \\
& TNR & TNAS & TNR & TNAS & TNR & TNAS & TNR & TNAS\\
\hline
&&&&K = 2&&&&\\
\hline
AVDHLA & \textbf{6189} & \textbf{1054} & \textbf{5638} & \textbf{376} & \textbf{5580} & \textbf{376} & \textbf{5829} & \textbf{480} \\
SVDHLA & 6471 & 951 & 5977 & 755 & 5558 & 295 & 5744 & 428 \\
FSLA & \textbf{6350} & \textbf{3650} & \textbf{6408} & \textbf{982} & \textbf{6093} & \textbf{667} & \textbf{5730} & \textbf{573} \\
\hline
&&&&K = 5&&&&\\
\hline
AVDHLA & \textbf{4905} & \textbf{1682} & \textbf{5003} & \textbf{2973} & \textbf{4942} & \textbf{2190} & \textbf{4831} & \textbf{1951} \\
SVDHLA & 4746 & 1380 & 5082 & 2279 & 5058 & 1810 & 4746 & 1485 \\
FSLA & \textbf{4090} & \textbf{5910} & \textbf{5248} & \textbf{2654} & \textbf{5103} & \textbf{2073} & \textbf{5027} & \textbf{1729} \\
\hline
&&&&K = 9&&&&\\
\hline
AVDHLA & \textbf{4741} & \textbf{3695} & \textbf{4226} & \textbf{2932} & \textbf{4295} & \textbf{2488} & \textbf{4245} & \textbf{2355} \\
SVDHLA & 4269 & 2827 & 4104 & 3142 & 4162 & 4195 & 4292 & 2008 \\
FSLA & \textbf{2764} & \textbf{7236} & \textbf{4300} & \textbf{3960} & \textbf{4320} & \textbf{3277} & \textbf{4202} & \textbf{2984} \\
\hline
\end{tabular}}
\end{table}

By looking at the results and the mathematical analysis together, we can understand that the designed environment is totally random, so the learning process in such an environment is impossible. Most of the time, AVDHLA performs better than SVDHLA and FSLA with respect to the TNR and TNAS because of the flexibility of this automaton to choose the desired depth for each action separately. Overall, we can't expect from these automata to learn from this environment.

\subsection{Scenario 2 \label{sec:Markovian_Switching_Scenario_2}}
This scenario is conducted to study the proposed automaton in Markovian Switching environment in which switching from one state to the other will change the favorable action completely. Also, the tendency to change from one state to the other is relatively high.

Mathematically, in transition matrix of markov chain, the probability of $a_{ij} \quad i = j$ is less than $a_{ij} \quad i \neq j$. The following transition matrix is considered to reach our desired condition.

\begin{equation}
\label{equ_markovian_2_transition_appendix}
T = \left(\begin{IEEEeqnarraybox}[][c]{,c/c/c,}
0.2& &0.8\\
0.6& &0.4
\end{IEEEeqnarraybox}\right)
\end{equation}

The following reward matrix is used to satisfy the condition of this experiment. Also, Fig.\ref{fig_markovian_switching_scanarios_appendix} b visualizes the desired Markovian Switching environment.

\begin{equation}
\label{equ_markovian_2_reward_appendix}
R = \left(\begin{IEEEeqnarraybox}[][c]{,c/c/c,}
0.8& &0.2\\
0.2& &0.8
\end{IEEEeqnarraybox}\right)
\end{equation}

Before analyzing the results which is shown in Table \ref{table_ex_markovian_switching_results_2_appendix}, it is better to convert the Markovian Switching environment to the stationary environment with the steady state analysis of the markov chain as follows:

\begin{IEEEeqnarray}{c}
\label{equ_markovian_2_conversion_appendix}
v_{\infty}T = v_{\infty} \IEEEyesnumber\IEEEyessubnumber*\\
v_{\infty}
\left(\begin{IEEEeqnarraybox}[][c]{,c/c/c,}
0.2& &0.8\\
0.6& &0.4
\end{IEEEeqnarraybox}\right)
= v_{\infty}\\
\left(\begin{IEEEeqnarraybox}[][c]{,c/c/c,}
x& &y
\end{IEEEeqnarraybox}\right)
\left(\begin{IEEEeqnarraybox}[][c]{,c/c/c,}
0.2& &0.8\\
0.6& &0.4
\end{IEEEeqnarraybox}\right) 
= \left(\begin{IEEEeqnarraybox}[][c]{,c/c/c,}
x& &y
\end{IEEEeqnarraybox}\right)\\
\begin{cases}
	0.2 x + 0.6y = x \\
    0.8 x + 0.4y = y \\
    x + y = 1
\end{cases} \\
\begin{cases}
	x = 0.42\\
    y = 0.58
\end{cases} 
\end{IEEEeqnarray}

Multiplication of the steady state matrix to the reward probability matrix of Markovian Switching environment yields to the reward probability matrix of the equivalent stationary environment.

\begin{IEEEeqnarray}{c}
\label{equ_markovian_2_transition_appendix}
v_{\infty}R_{Markovian \enspace Switching} = R_{Stationary} \IEEEyesnumber\IEEEyessubnumber*\\
\left(\begin{IEEEeqnarraybox}[][c]{,c/c/c,}
0.42& &0.58
\end{IEEEeqnarraybox}\right)
\left(\begin{IEEEeqnarraybox}[][c]{,c/c/c,}
0.8& &0.2\\
0.2& &0.8
\end{IEEEeqnarraybox}\right)
=
\left(\begin{IEEEeqnarraybox}[][c]{,c/c/c,}
0.45& &0.54
\end{IEEEeqnarraybox}\right)
\end{IEEEeqnarray}

Therefore, the second action is more likely to being rewarded by the outer environment. We should denote that the difference between these automata are not a lot enough to say that one action is favorable, but we expect from the automaton to learn to choose the second action more than the first one. 

\begin{table}[!h]
\renewcommand{\arraystretch}{1.3}
\caption{Experimental Results of the Second Scenario with Respect to TNR and TNAS}
\label{table_ex_markovian_switching_results_2_appendix}
\centering
\resizebox{\columnwidth}{!}{
\begin{tabular}{lcccccccc}
\hline
Model & \multicolumn{2}{l}{\;\;\;\;Config 1} & \multicolumn{2}{l}{\;\;\;\;Config 2} & \multicolumn{2}{l}{\;\;\;\;Config  3} & \multicolumn{2}{l}{\;\;\;\;Config 4} \\
& TNR & TNAS & TNR & TNAS & TNR & TNAS & TNR & TNAS\\
\hline
&&&&K = 2&&&&\\
\hline
AVDHLA & \textbf{5177} & \textbf{565} & \textbf{5183} & \textbf{946} & \textbf{5105} & \textbf{765} & \textbf{5362} & \textbf{93} \\
SVDHLA & 5248 & 591 & 5278 & 597 & 5390 & 170 & 5290 & 289 \\
FSLA & \textbf{4291} & \textbf{5709} & \textbf{4971} & \textbf{1570} & \textbf{5101} & \textbf{811} & \textbf{5218} & \textbf{494} \\
\hline
&&&&K = 5&&&&\\
\hline
AVDHLA & \textbf{3834} & \textbf{2490} & \textbf{3635} & \textbf{3013} & \textbf{3761} & \textbf{2639} & \textbf{3772} & \textbf{2554} \\
SVDHLA & 3266 & 3915 & 3939 & 2261 & 4094 & 1881 & 3749 & 2546 \\
FSLA & \textbf{2014} & \textbf{7986} & \textbf{3134} & \textbf{4285} & \textbf{3436} & \textbf{3448} & \textbf{3663} & \textbf{2872} \\
\hline
&&&&K = 9&&&&\\
\hline
AVDHLA & \textbf{2267} & \textbf{5636} & \textbf{2415} & \textbf{5292} & \textbf{2816} & \textbf{4493} & \textbf{2884} & \textbf{4331} \\
SVDHLA & 2720 & 4564 & 2434 & 5268 & 2554 & 5005 & 2930 & 4168 \\
FSLA & \textbf{1150} & \textbf{8850} & \textbf{2170} & \textbf{5973} & \textbf{2298} & \textbf{5526} & \textbf{2685} & \textbf{4722} \\
\hline
\end{tabular}}
\end{table}

The results and the mathematical analysis show that both SVDHLA and AVDHLA outperform FSLA with respect to TNR and TNAS. This performance can be seen in lower initial depth. The underlying reason is the confusion of automaton with changing the favorable action in each state and tendency of the environment to change the current state. As a matter of fact, higher values for depth, shows bias towards the favorable action and as a result making the right decision for the automaton with a lot of costs. Overall, we can infer from such environments in which the favorable action changes periodically, VDHLA might be a good choice. 

\subsection{Scenario 3 \label{sec:Markovian_Switching_Scenario_3}}
The purpose of this scenario is assessing the proposed automaton in Markovian Switching environment in conjunction with TNR and TNAS. The considered environment is a two-state markov chain in which switching from one state to the other will decrease or increase the probability of being reward in terms of the favorable action, but that action is still the favorable one. As well, the environment intends to stay in the current state.

Analytically, in the transition matrix of the markov chain, the probability of element $a_{ij} \quad i = j$ is more than $a_{ij} \quad i \neq j$. Transition matrix of such an environment is as follows:

\begin{equation}
\label{equ_markovian_3_transition_appendix}
T = \left(\begin{IEEEeqnarraybox}[][c]{,c/c/c,}
0.9& &0.1\\
0.1& &0.9
\end{IEEEeqnarraybox}\right)
\end{equation}

To reach the desired environment, the following reward matrix for the markov chain is considered. Also Fig.\ref{fig_markovian_switching_scanarios_appendix} c shows the desired Markovian Switching environment.

\begin{equation}
\label{equ_markovian_3_reward_appendix}
R = \left(\begin{IEEEeqnarraybox}[][c]{,c/c/c,}
0.8& &0.2\\
0.7& &0.3
\end{IEEEeqnarraybox}\right)
\end{equation}

Before analysis of the result in Table \ref{table_ex_markovian_switching_results_3_appendix} , conversion of the Markovian Switching environment to the stationary environment should be done using the steady state analysis of the markov chain as follows:

\begin{IEEEeqnarray}{c}
\label{equ_markovian_3_conversion_appendix}
v_{\infty}T = v_{\infty} \IEEEyesnumber\IEEEyessubnumber*\\
v_{\infty}
\left(\begin{IEEEeqnarraybox}[][c]{,c/c/c,}
0.9& &0.1\\
0.1& &0.9
\end{IEEEeqnarraybox}\right)
= v_{\infty}\\
\left(\begin{IEEEeqnarraybox}[][c]{,c/c/c,}
x& &y
\end{IEEEeqnarraybox}\right)
\left(\begin{IEEEeqnarraybox}[][c]{,c/c/c,}
0.9& &0.1\\
0.1& &0.9
\end{IEEEeqnarraybox}\right) 
= \left(\begin{IEEEeqnarraybox}[][c]{,c/c/c,}
x& &y
\end{IEEEeqnarraybox}\right)\\
\begin{cases}
	0.9 x + 0.1 y = x \\
    0.1 x + 0.9 y = 1 \\
    x + y = 1
\end{cases} \\
\begin{cases}
	x = 0.5\\
    y = 0.5
\end{cases}
\end{IEEEeqnarray}

Getting the reward matrix of stationary environment depends on multiplication of the steady state matrix to reward probability matrix of Markovian Switching environment.

\begin{IEEEeqnarray}{c}
\label{equ_markovian_3_transition_appendix}
v_{\infty}R_{Markovian \enspace Switching} = R_{Stationary} \IEEEyesnumber\IEEEyessubnumber*\\
\left(\begin{IEEEeqnarraybox}[][c]{,c/c/c,}
0.5& &0.5
\end{IEEEeqnarraybox}\right)
\left(\begin{IEEEeqnarraybox}[][c]{,c/c/c,}
0.8& &0.2\\
0.7& &0.3
\end{IEEEeqnarraybox}\right)
=
\left(\begin{IEEEeqnarraybox}[][c]{,c/c/c,}
0.75& &0.25
\end{IEEEeqnarraybox}\right)
\end{IEEEeqnarray}

Accordingly, Markovian Switching environment turns into the stationary environment with the favorable action 1 with 75\% and action 2 with 25\% of time being rewarded.

\begin{table}[!h]
\renewcommand{\arraystretch}{1.3}
\caption{Experimental Results of the Third Scenario with Respect to TNR and TNAS}
\label{table_ex_markovian_switching_results_3_appendix}
\centering
\resizebox{\columnwidth}{!}{
\begin{tabular}{lcccccccc}
\hline
Model & \multicolumn{2}{l}{\;\;\;\;Config 1} & \multicolumn{2}{l}{\;\;\;\;Config 2} & \multicolumn{2}{l}{\;\;\;\;Config  3} & \multicolumn{2}{l}{\;\;\;\;Config 4} \\
& TNR & TNAS & TNR & TNAS & TNR & TNAS & TNR & TNAS\\
\hline
&&&&K = 2&&&&\\
\hline
AVDHLA & \textbf{7551} & \textbf{33} & \textbf{7421} & \textbf{216} & \textbf{7422} & \textbf{90} & \textbf{7469} & \textbf{73} \\
SVDHLA & 7452 & 132 & 7465 & 320 & 7420 & 80 & 7496 & 70 \\
FSLA & \textbf{6319} & \textbf{3681} & \textbf{7346} & \textbf{361} & \textbf{7433} & \textbf{60} & \textbf{7424} & \textbf{7} \\
\hline
&&&&K = 5&&&&\\
\hline
AVDHLA & \textbf{4566} & \textbf{2384} & \textbf{4690} & \textbf{2316} & \textbf{4650} & \textbf{2043} & \textbf{4696} & \textbf{2107} \\
SVDHLA & 4655 & 1876 & 4374 & 2790 & 4539 & 2866 & 4510 & 2956 \\
FSLA & \textbf{3604} & \textbf{6396} & \textbf{4677} & \textbf{2888} & \textbf{4465} & \textbf{2397} & \textbf{4609} & \textbf{2117} \\
\hline
&&&&K = 9&&&&\\
\hline
AVDHLA & \textbf{4224} & \textbf{3630} & \textbf{4162} & \textbf{4071} & \textbf{4168} & \textbf{3413} & \textbf{4385} & \textbf{2553} \\
SVDHLA & 4103 & 3689 & 4103 & 2934 & 4032 & 2538 & 4233 & 3447 \\
FSLA & \textbf{2865} & \textbf{7135} & \textbf{4149} & \textbf{4149} & \textbf{4110} & \textbf{3585} & \textbf{4066} & \textbf{2900} \\
\hline
\end{tabular}}
\end{table}

The results show the superiority of AVDHLA over both SVDHLA and FSLA with respect to TNR and TNAS. In lower values of $K$ and $N$, this superiority is more obvious. The main reason of the first one is that the lower value of $K$ can help AVDHLA to tune the depth of each action faster. Consequently it performs better by getting more rewards meanwhile with lower number of action switching. The second superiority is in low depth, since AVDHLA can asymmetrically chosen its depth relatives to SVDHLA and FSLA. As we expected, this automaton works better in such an environment in conjunction with the experiment metrics. 

\subsection{Scenario 4 \label{sec:Markovian_Switching_Scenario_4}}
This scenario is conducted to evaluate the proposed automaton in Markovian Switching environment in conjunction to TNR and TNAS. Two-state markov chain in which the favorable action in all state is the same as before but changing the state can decrease or increase the probability of being rewarded of the favorable action. On the other hand, the environment tries to change the current state.

Mathematically, in transition matrix of the markov chain, the probability of $a_{ij} \quad i = j$ is less than $a_{ij} \quad i \neq j$. The following transition matrix is considered for this scenario.

\begin{equation}
\label{equ_markovian_4_transition_appendix}
T = \left(\begin{IEEEeqnarraybox}[][c]{,c/c/c,}
0.2& &0.8\\
0.6& &0.4
\end{IEEEeqnarraybox}\right)
\end{equation}

The regarded reward matrix to satisfy the specified conditions is as follows. Also, in Fig.\ref{fig_markovian_switching_scanarios_appendix} d the desired Markovian Switching environment is shown.

\begin{equation}
\label{equ_markovian_4_reward_appendix}
R = \left(\begin{IEEEeqnarraybox}[][c]{,c/c/c,}
0.8& &0.2\\
0.7& &0.3
\end{IEEEeqnarraybox}\right)
\end{equation}

To have a better overview on the specified environment, we shall convert Markovian Switching environment into the stationary environment using steady state analysis of the markov chain as follows:

\begin{IEEEeqnarray}{c}
\label{equ_markovian_4_conversion_appendix}
v_{\infty}T = v_{\infty} \IEEEyesnumber\IEEEyessubnumber*\\
v_{\infty}
\left(\begin{IEEEeqnarraybox}[][c]{,c/c/c,}
0.2& &0.8\\
0.6& &0.4
\end{IEEEeqnarraybox}\right)
= v_{\infty}\\
\left(\begin{IEEEeqnarraybox}[][c]{,c/c/c,}
x& &y
\end{IEEEeqnarraybox}\right)
\left(\begin{IEEEeqnarraybox}[][c]{,c/c/c,}
0.2& &0.8\\
0.6& &0.4
\end{IEEEeqnarraybox}\right) 
= \left(\begin{IEEEeqnarraybox}[][c]{,c/c/c,}
x& &y
\end{IEEEeqnarraybox}\right)\\
\begin{cases}
	0.2 x + 0.6 y = x \\
    0.8 x + 0.4 y = 1 \\
    x + y = 1
\end{cases} \\
\begin{cases}
	x = 0.42\\
    y = 0.58
\end{cases}
\end{IEEEeqnarray}

The final result is gotten from the multiplication of steady state matrix into the reward probability matrix of Markovian Switching environment.

\begin{IEEEeqnarray}{c}
\label{equ_markovian_3_transition_appendix}
v_{\infty}R_{Markovian \enspace Switching} = R_{Stationary} \IEEEyesnumber\IEEEyessubnumber*\\
\left(\begin{IEEEeqnarraybox}[][c]{,c/c/c,}
0.42 & &0.58
\end{IEEEeqnarraybox}\right)
\left(\begin{IEEEeqnarraybox}[][c]{,c/c/c,}
0.8& &0.2\\
0.7& &0.3
\end{IEEEeqnarraybox}\right)
=
\left(\begin{IEEEeqnarraybox}[][c]{,c/c/c,}
0.74& &0.26
\end{IEEEeqnarraybox}\right)
\end{IEEEeqnarray}

Consequently, Markovian Switching environment turns into the stationary environment with the probability of being reward for the first action is 0.74 and for the second one is 0.26.

\begin{table}[!h]
\renewcommand{\arraystretch}{1.3}
\caption{Experimental Results of the Last Scenario with Respect to TNR and TNAS}
\label{table_ex_markovian_switching_results_3_appendix}
\centering
\resizebox{\columnwidth}{!}{
\begin{tabular}{lcccccccc}
\hline
Model & \multicolumn{2}{l}{\;\;\;\;Config 1} & \multicolumn{2}{l}{\;\;\;\;Config 2} & \multicolumn{2}{l}{\;\;\;\;Config  3} & \multicolumn{2}{l}{\;\;\;\;Config 4} \\
& TNR & TNAS & TNR & TNAS & TNR & TNAS & TNR & TNAS\\
\hline
&&&&K = 2&&&&\\
\hline
AVDHLA & \textbf{7323} & \textbf{129} & \textbf{7449} & \textbf{36} & \textbf{7409} & \textbf{23} & \textbf{7469} & \textbf{1} \\
SVDHLA & 7364 & 44 & 7301 & 168 & 7430 & 24 & 7417 & 24 \\
FSLA & \textbf{6148} & \textbf{3852} & \textbf{7227} & \textbf{398} & \textbf{7365} & \textbf{63} & \textbf{7380} & \textbf{6} \\
\hline
&&&&K = 5&&&&\\
\hline
AVDHLA & \textbf{2807} & \textbf{4509} & \textbf{2997} & \textbf{4064} & \textbf{2784} & \textbf{4775} & \textbf{3028} & \textbf{4050} \\
SVDHLA & 2733 & 4940 & 2959 & 4280 & 2868 & 4458 & 2899 & 4321 \\
FSLA & \textbf{1997} & \textbf{8003} & \textbf{2674} & \textbf{4975} & \textbf{2946} & \textbf{4251} & \textbf{3001} & \textbf{3999} \\
\hline
&&&&K = 9&&&&\\
\hline
AVDHLA & \textbf{2275} & \textbf{5740} & \textbf{2549} & \textbf{5135} & \textbf{2934} & \textbf{4225} & \textbf{2742} & \textbf{4538} \\
SVDHLA & 2770 & 4564 & 2439 & 5270 & 2536 & 5049 & 2408 & 5265 \\
FSLA & \textbf{1231} & \textbf{8769} & \textbf{2219} & \textbf{5850} & \textbf{2522} & \textbf{5012} & \textbf{2613} & \textbf{4855} \\
\hline
\end{tabular}}
\end{table}

By looking at the results, VDHLA outperforms FSLA with respect to TNR and TNAS. In such an environment with existence of the favorable action, VDHLA can increase its depth through the favorable action. As a result, it will collect more rewards with less number of action switching. Eventually, VDHLA can be robust in this environment in conjunction with the considered metrics.

\section{Nik Defense Experiment Extras \label{section:appendix_exp_nik_defense}}
This appendix is dedicated to perform extra experiments on the proposed defense in the existence of VDHLA as a decision maker. Three scenarios of  are considered as follows:

\begin{enumerate}

\item This scenario examines the effect of fail-safe parameter on the quality of the proposed defense.

\item To understand how $\tau$ parameter affects the quality of the proposed defense, this scenario is conducted.

\item As well as other parameters, number of $\tau$ in one $\theta$ will be studied to see the quality of the proposed defense with the various values of $\theta$.

\end{enumerate}

We should remark that 10000 blocks are generated for each scenario. Inner VASLAs are $L_{R\_{\epsilon P}}$ with $\lambda_1=0.1$ and $\lambda_2=0.01$.

\subsection{Scenario 1}
This scenario is conducted to study the proposed defense with various values of K to see the effect of the fail-safe parameter on the relative revenue and the lower bound threshold. For this purpose, $[1, 3]$, $[2, 4]$, and $[1, 5]$ intervals of $K$ are considered to perform the experiment. K values can swing between the minimum value and the max discreetly. The value of $\tau$ is about the time of mining five blocks, and one $\theta$ has ten time intervals.

\begin{figure}[!h]
\centering
\subfloat[SVDHLA]{\includegraphics[width=0.25\textwidth]{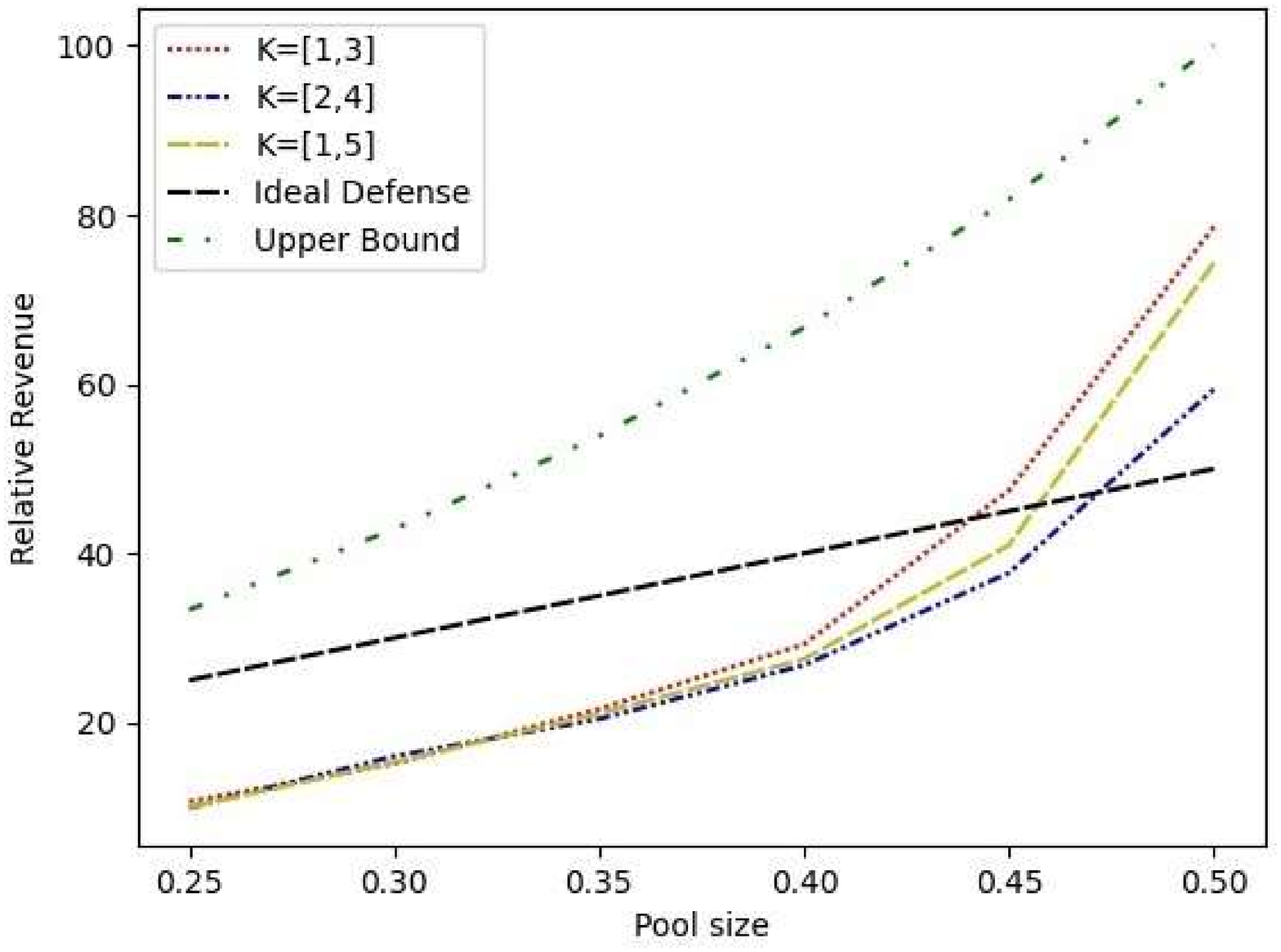}} 
\subfloat[AVDHLA]{\includegraphics[width=0.25\textwidth]{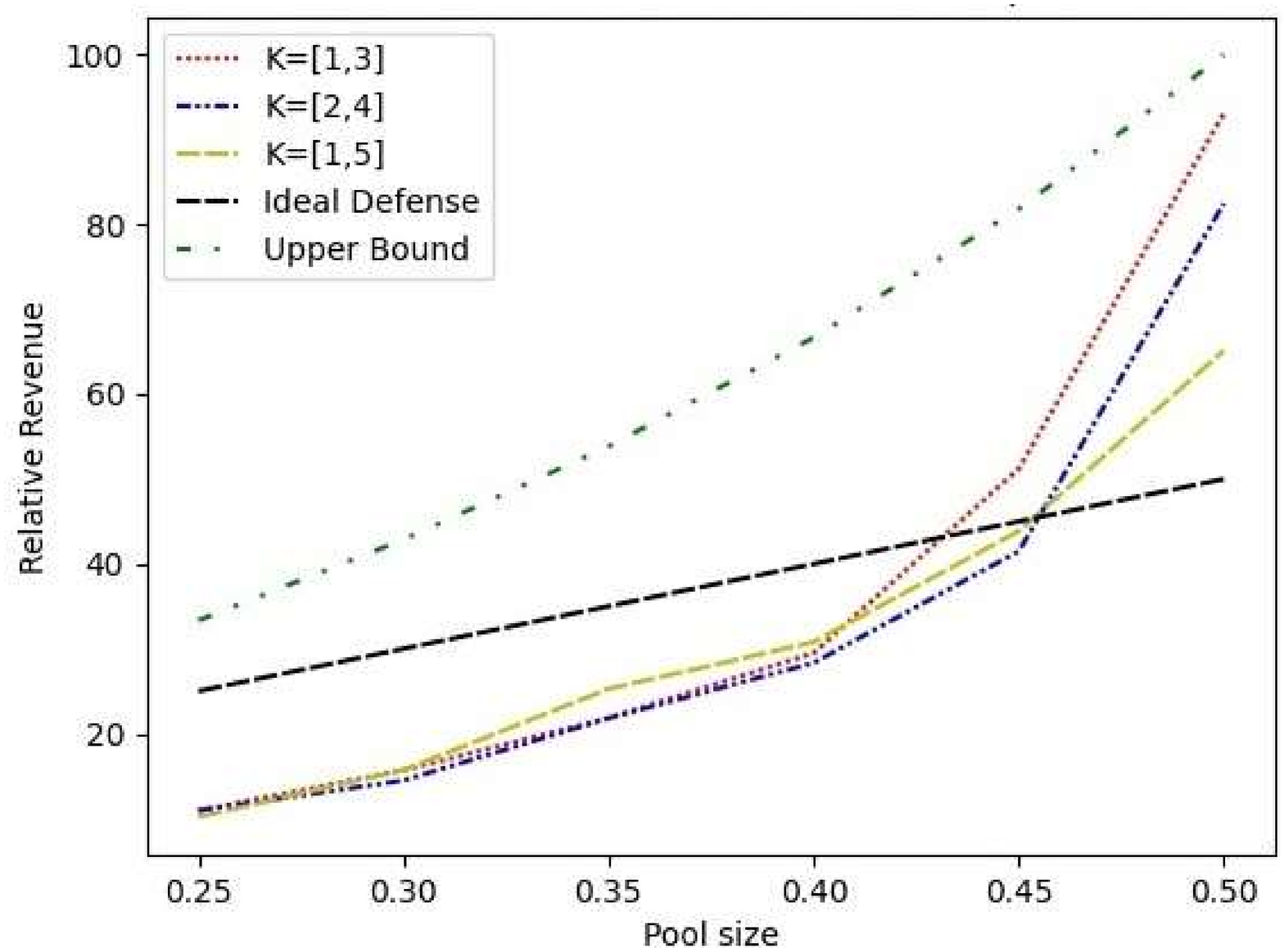}}\\ 
\caption{Experimental results of Scenario 1 with respect to the relative revenue}
\label{fig_app_scenario_1_appendix}
\end{figure} 

The results in Fig.\ref{fig_app_scenario_1_appendix} demonstrate the superiority of $[1, 5]$ for AVDHLA, and $[2, 4]$ for SVDHLA with respect to the relative revenue. It shows the tendency of the automaton toward the higher values of $K$. In AVDHLA, the automaton asymmetrically chose the depth for its action, so longer intervals can help it to choose the accurate value for the depth. Meanwhile, since SVDHLA can learn faster, the interval with neither high value nor low value for $K$ will be fit.
On the other hand observations show the same result for the lower bound threshold. Overall we can claim that longer values of fail-safe interval is better for AVDHLA because of the freedom to choose asymmetrically and narrow intervals are better for SVDHLA in relation to the experiment metrics.

\subsection{Scenario 2 \label{sec:Nik_Scenario_2}}
The ultimate objective of this scenario is to study the impact of $\tau$ time interval on the quality of the proposed defense with respect to the relative revenue. To aim this, three different values of $\tau$ are considered: $\tau=5,9,18$. The $K$ parameter swings between $K_{min} = 1$ and $K_{max} = 3$. For sake of example, $\tau = 5$ means that one $\tau$ time interval is equivalent to the mining time of five blocks. Also, one $\theta$ has ten $\tau$.

\begin{figure}[!h]
\centering
\subfloat[SVDHLA]{\includegraphics[width=0.25\textwidth]{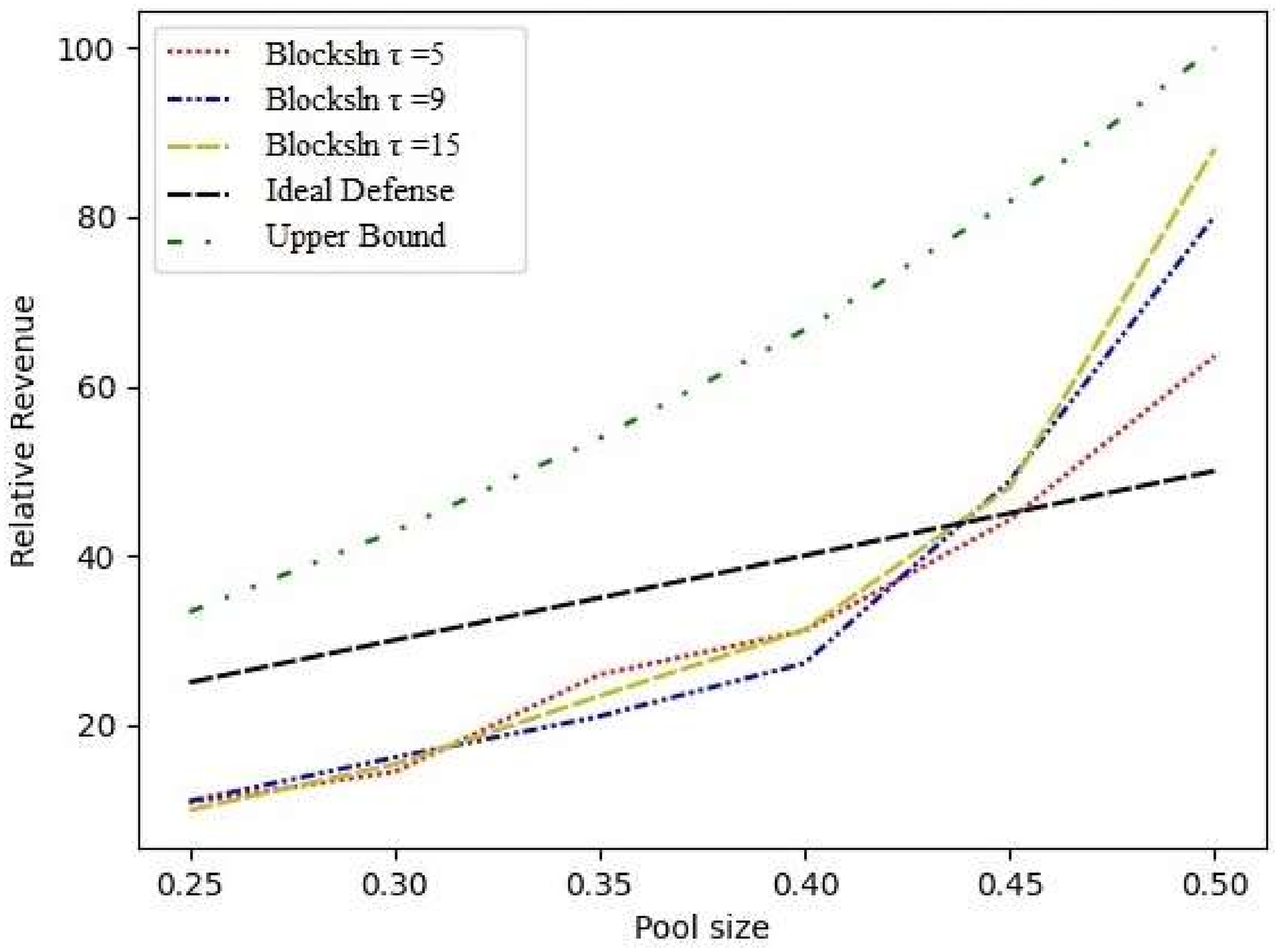}} 
\subfloat[AVDHLA]{\includegraphics[width=0.25\textwidth]{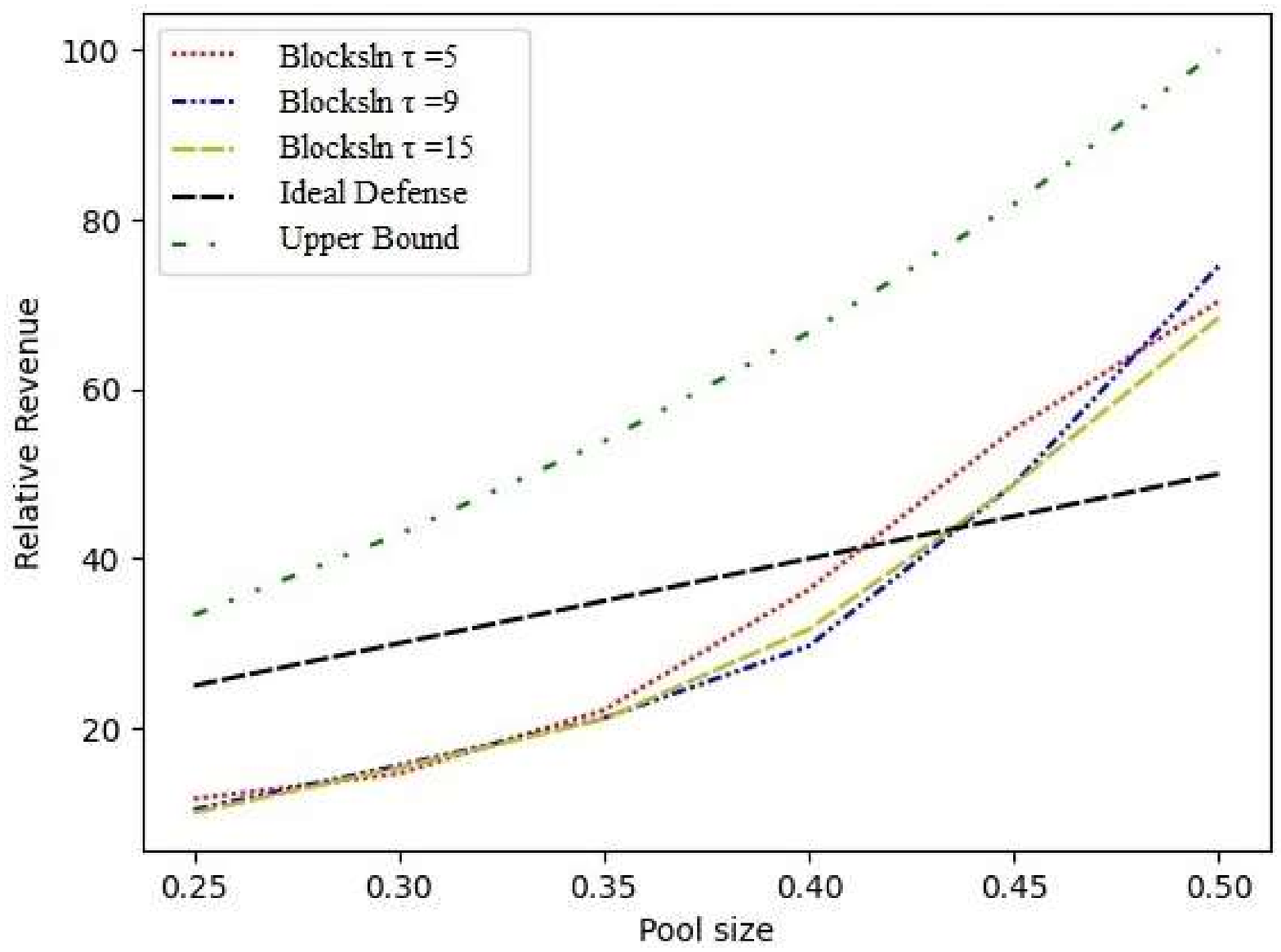}}\\ 
\caption{Experimental results of Scenario 2 with respect to the relative revenue}
\label{fig_app_scenario_2_appendix}
\end{figure} 

The obtained results are displayed in Fig. \ref{fig_app_scenario_2_appendix}. From these results, one can observe that the lower values of $\tau$ lead to the less relative revenue for the selfish miners. This is near to our expectations, since the number of $\tau$ in one $\theta$ is constant. On the hand, lower values of $\tau$ lead to increase the number of $\theta$ in one simulation. It means that the automaton get more feedback from the environment. As a result, it will tune $\beta$ parameter effectively.

\subsection{Scenario 3}
The last scenario is about investigating the impact of number of $\tau$ in one $\theta$ on the performance of the proposed defense in relation to the relative revenue. For this purpose, $\tau$ is considered about the time of mining five blocks. As well, three values of 6, 12, and 18 are regarded number of $\tau$ in one $\theta$.

\begin{figure}[!h]
\centering
\subfloat[SVDHLA]{\includegraphics[width=0.25\textwidth]{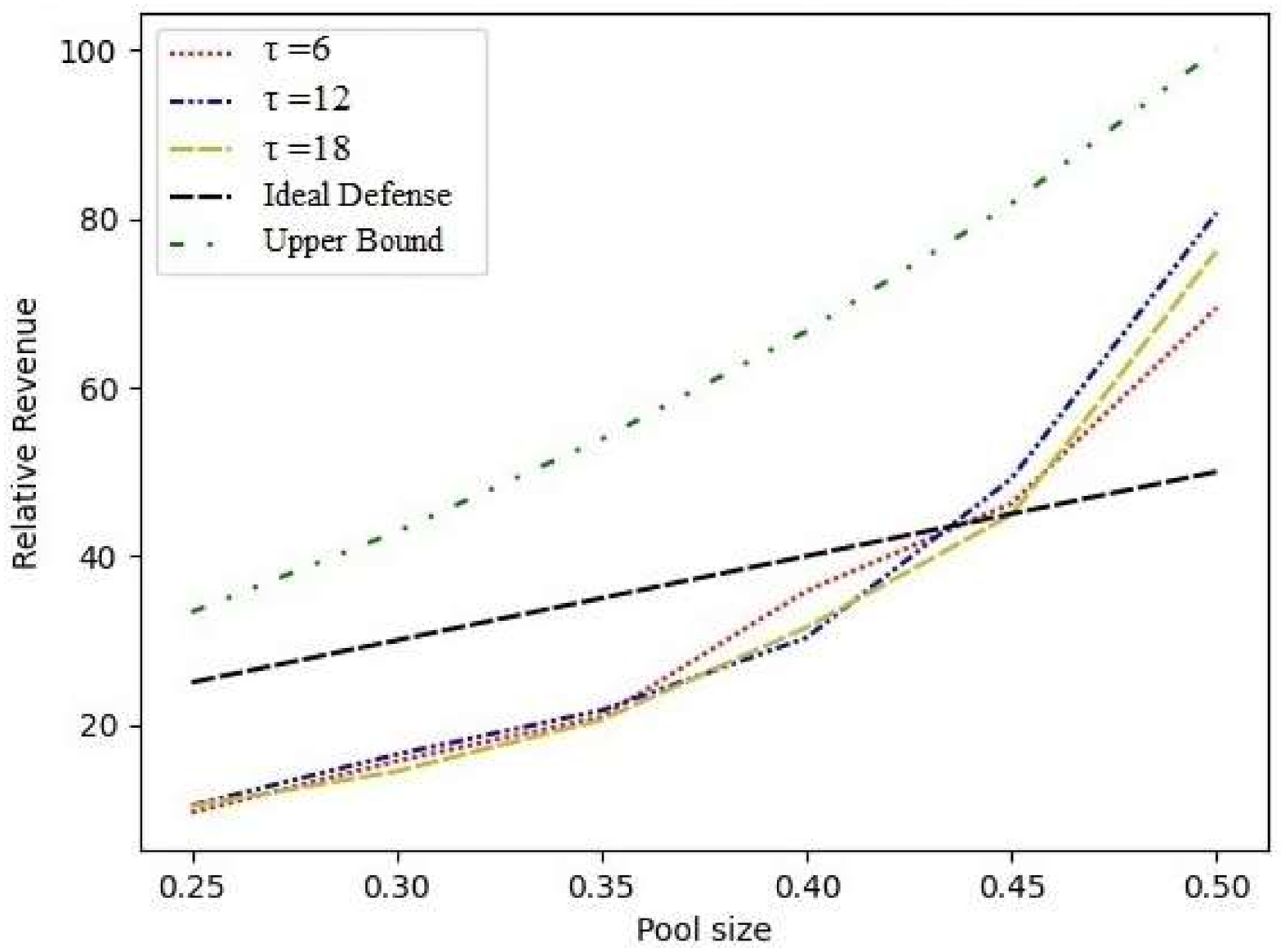}} 
\subfloat[AVDHLA]{\includegraphics[width=0.25\textwidth]{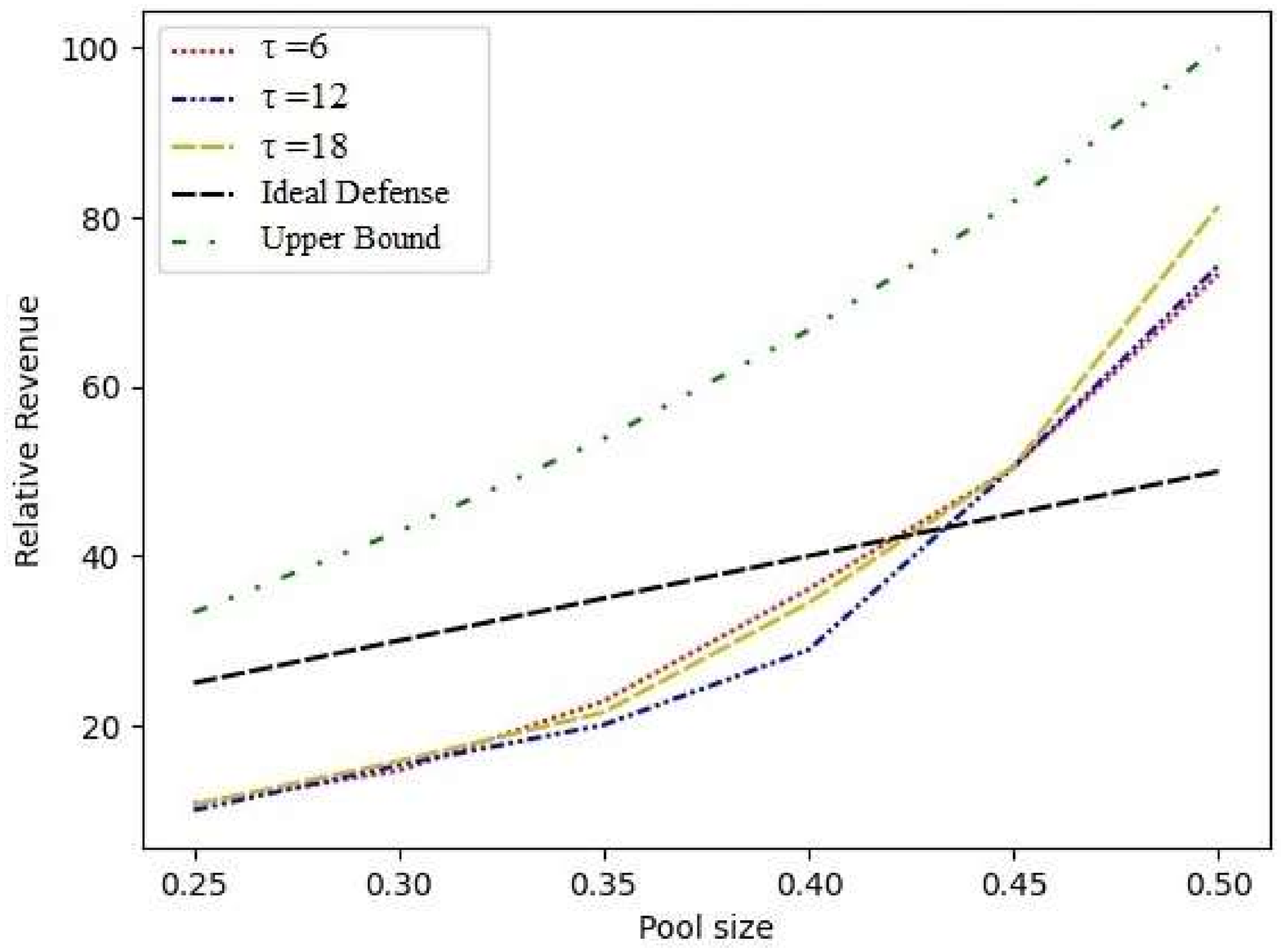}}\\ 
\caption{Experimental results of Scenario 3 with respect to the relative revenue}
\label{fig_app_scenario_3_appendix}
\end{figure} 

By looking at the results in Fig.\ref{fig_app_scenario_3_appendix} for both SVDDHLA and AVDHLA, the superiority of the lower number of $\tau$ in one $\theta$ with respect to the relative revenue of the selfish miners is obvious.

As predicted from the behavior of the proposed automaton, the less number of $\tau$ in one $\theta$ leads to more number of $\theta$ in one simulation. This means that the automaton can adopt itself with the complex environment such as blockchain by setting the $\beta$ parameter.

Eventually, we can claim from this scenario of the experiment and the previous one (\ref{sec:Nik_Scenario_2}) that tunning the reinforcement signal accurately leads to the more expensive selfish mining attack. As a result, we have a better defense in conjunction with the relative revenue. 

\ifCLASSOPTIONcaptionsoff
  \newpage
\fi



%

%

\begin{IEEEbiography}[{\includegraphics[width=1in,height=1.25in,clip,keepaspectratio]{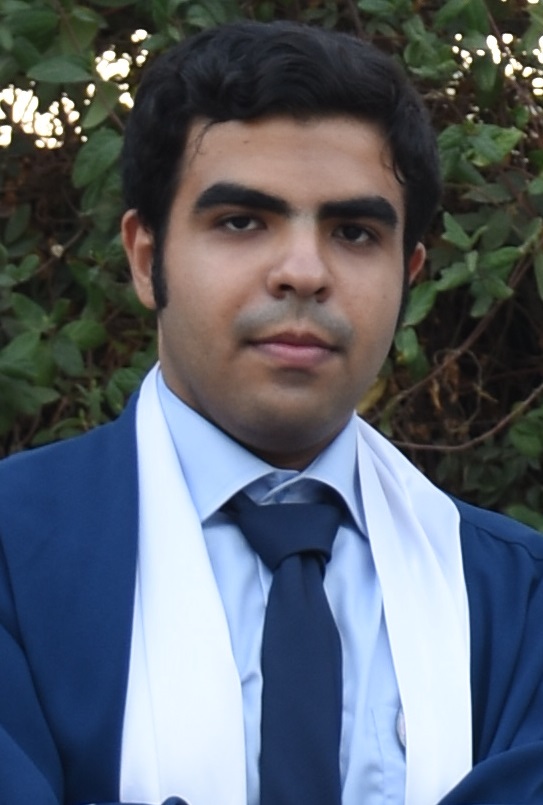}}]{Ali Nikhalat-Jahromi}
He received the B.Sc degree in Electrical Engineering in 2018 and then the M.Sc degree in Computer Engineering in 2021, both from Amirkabir University of Technology, Tehran, Iran. His research interests include Software Systems, Parallel and Distributed Systems, Machine Learning and Programming Languages.
\end{IEEEbiography}

\begin{IEEEbiography}[{\includegraphics[width=1in,height=1.25in,clip,keepaspectratio]{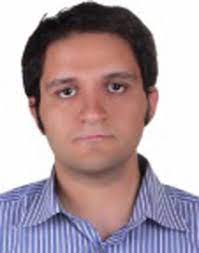}}]{Ali Mohammad Saghiri}
He received the B.Sc. degree from University of Science and Culture, in 2007 and the M.Sc. and Ph.D. degrees from AmirKabir University of Technology, Tehran, Iran, in 2010 and 2017, respectively, all in computer engineering. He published more than 40 scientific papers on international conferences and journals among which Journal of Network and Computer Applications(JNCA), Applied Intelligence, and international journal of communication systems. His research interests include the Internet of Things, Blockchain, and Artificial Intelligence.
\end{IEEEbiography}

\begin{IEEEbiography}[{\includegraphics[width=1in,height=1.25in,clip,keepaspectratio]{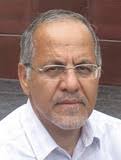}}]{Mohammad Reza Meybodi}
He received the B.Sc. and M.Sc. degrees in economics from Shahid Beheshti University, Tehran, Iran, in 1973 and 1977, respectively, the M.Sc. and Ph.D. degrees in computer science from Oklahoma University, Norman, OK, USA, in 1980 and 1983, respectively.,He was an Assistant Professor with Western Michigan University, Kalamazoo, MI, USA, from 1983 to 1985, and an Associate Professor with Ohio University, Athens, OH, USA, from 1985 to 1991. He is currently a Full Professor with the Computer Engineering Department, Amirkabir University of Technology, Tehran. His research interests include wireless networks, fault tolerant systems, learning systems, parallel algorithms, soft computing, and software development.
\end{IEEEbiography}



\end{document}